\documentclass{article} \usepackage[preprint]{tmlr}
\usepackage{booktabs}
\usepackage{multirow}
\usepackage{makecell}
\usepackage{graphicx}
\usepackage{float}
\usepackage{amsmath}
\usepackage{amssymb}
\usepackage{xcolor}
\definecolor{navyblue}{RGB}{31,79,216}
\newcommand{\ToyTau}{5.0}       \newcommand{\ToyEnpLow}{3.2}    \newcommand{\ToyEnpHigh}{6.8}   

\usepackage{hyperref}
\hypersetup{
  colorlinks=true,
  linkcolor=navyblue,   citecolor=navyblue,   urlcolor=navyblue,    }

\title{Ceiling-Clipped Acceptance Histograms Indicate Stranded Speed-up in Block-Diffusion Speculative Decoding}

\author{\centering \normalsize\bf Ephrem Wu \\
        \small\rm Advanced Micro Devices, Inc., USA}

\def\month{07}
\def\year{2026}
\def\openreview{\url{https://openreview.net/forum?id=XXXXXXXXXX}}

\providecommand{\CycleTimeRatioCount}{56}
\providecommand{\CycleTimeRatioMedian}{1.003}
\providecommand{\CycleTimeRatioSd}{2.5}

\providecommand{\SpeedupChangeArmAMedian}{+5.7\%}
\providecommand{\SpeedupChangeArmBHighCeilingMedian}{+11.9\%}
\providecommand{\SpeedupChangeArmBMedian}{+10.2\%}
\providecommand{\SpeedupChangeGapCount}{56}
\providecommand{\SpeedupChangeGapMedian}{0.4}

\providecommand{\FamilySize}{24}

\providecommand{\FamilyPositiveAfterHolm}{24}

\providecommand{\FamilyNegativeAfterHolmWord}{zero}
\providecommand{\FamilyMedianDtau}{+1.0}

\providecommand{\FamilyMinDtau}{+0.5}
\providecommand{\FamilyMaxDtau}{+1.7}
\providecommand{\SummaryDtauMin}{0.23}
\providecommand{\SummaryDtauMax}{1.37}

\providecommand{\FamilyAllSize}{56}

\providecommand{\FamilyAllPositiveAfterHolm}{53}

\providecommand{\FamilyAllMedianDtau}{+0.6}
\providecommand{\FamilyAllMinDtau}{+0.07}
\providecommand{\FamilyAllMaxDtau}{+1.7}

\providecommand{\FamilyIncrMedianDtau}{+0.8}
\providecommand{\FamilyIncrMinDtau}{+0.38}
\providecommand{\FamilyIncrMaxDtau}{+1.1}

\providecommand{\JetTopBudget}{256}

\providecommand{\JetBenchCountWord}{seven}

\providecommand{\JetArmABeats}{128}

\providecommand{\JetTopWinsWord}{five}

\providecommand{\JetTopTiesWord}{two}
\providecommand{\JetPointBeatsAll}{64}

\providecommand{\JetClearBeatsAllArmB}{32}

\providecommand{\MassShiftNetDflareEightBAime}{-0.01}

\providecommand{\MassShiftNetDflareEightBLcb}{-0.08}

\providecommand{\MassShiftNetDflareEightBMath}{-0.41}

\providecommand{\MassShiftNetDflashEightBAime}{-0.07}

\providecommand{\NaiveFourBMathNativeSp}{6.5}
\providecommand{\NaiveFourBMathNaiveSp}{3.0}
\providecommand{\NaiveFourBMathDropPct}{54}
\providecommand{\NaiveFourBMathNativeEnp}{8.8}
\providecommand{\NaiveFourBMathNaiveEnp}{4.1}
\providecommand{\NaiveEightBMathNativeSp}{6.6}
\providecommand{\NaiveEightBMathNaiveSp}{6.4}

\providecommand{\NaiveEightBMathNativeEnp}{9.0}
\providecommand{\NaiveEightBMathNaiveEnp}{8.7}

\providecommand{\GateArmARho}{+0.88}

\providecommand{\GateArmAN}{35}
\providecommand{\GateArmARhoEn}{+0.89}

\providecommand{\GateArmAWithinGemmaTwelveB}{+0.86}

\providecommand{\GemmaTarget}{Gemma-4-12B-IT}

\providecommand{\GemmaBenchCountWord}{seven}

\providecommand{\GemmaMedianDtau}{+0.41}
\providecommand{\GemmaMinDtau}{+0.09}
\providecommand{\GemmaMaxDtau}{+0.73}

\providecommand{\LossAblationConditions}{6}
\providecommand{\LossAblationBenches}{5}
\providecommand{\LossAblationMeanMin}{6.75}
\providecommand{\LossAblationMeanMax}{6.79}

\providecommand{\ExpFracDflare}{1.2}
\providecommand{\ExpFracDflash}{3.8}
\providecommand{\ExpFracDflareCont}{1.0}
\providecommand{\ExpFracDflashCont}{2.0}

\providecommand{\TauEnpDflashFourBMathTau}{9.45}
\providecommand{\TauEnpDflashFourBMathEnp}{9.07}
\providecommand{\TauEnpDflashFourBMathDiffMag}{0.38}
\providecommand{\TauEnpDflareEightBAimeTau}{9.60}
\providecommand{\TauEnpDflareEightBAimeEnp}{10.02}
\providecommand{\TauEnpDflareEightBAimeDiffMag}{0.41}

\providecommand{\LowCeilNullMbppDtau}{+0.07 [-0.01, +0.14] (ns)}
\providecommand{\LowCeilSplitMbppCont}{+0.80}
\providecommand{\LowCeilSplitMbppExp}{+0.21}

\providecommand{\RecipeTrainWindow}{3{,}072}
\providecommand{\RecipeDistillMaxNew}{2{,}048}

\providecommand{\MetricMaxDivergenceWord}{a token}

\providecommand{\NBenchWithinWord}{seven}
\providecommand{\SpearmanCritWithin}{0.786}

\providecommand{\EosNTotalMathFiveHundred}{500}
\providecommand{\EosNMinMathFiveHundred}{476}
\providecommand{\EosNTotalAime}{179}
\providecommand{\EosNMinAime}{111}

\providecommand{\EosNTotalLcb}{1055}
\providecommand{\EosNMinLcb}{962}

\providecommand{\EosInflDflareEightBAimeTauEos}{8.68}
\providecommand{\EosInflDflareEightBAimeTauAll}{10.27}
\providecommand{\EosInflDflareEightBAimeTauDelta}{+1.60}
\providecommand{\EosInflDflareEightBAimeEnEos}{8.94}
\providecommand{\EosInflDflareEightBAimeEnAll}{11.74}
\providecommand{\EosInflDflareEightBAimeEnDelta}{+2.80}

\providecommand{\ControlDataOnlyMin}{+0.05}
\providecommand{\ControlDataOnlyMax}{+0.21}
\providecommand{\ControlExpansionMin}{+0.26}
\providecommand{\ControlExpansionMax}{+1.02}

\providecommand{\CorpusSweepTenKMedianDtau}{+0.52}
\providecommand{\CorpusSweepThirtyKMedianDtau}{+0.57}
\providecommand{\CorpusSweepHundredKMedianDtau}{+0.57}
\providecommand{\CorpusSweepTenKLongMedianDtau}{+0.42}
\providecommand{\CorpusSweepThirtyKLongMedianDtau}{+0.65}
\providecommand{\CorpusSweepHundredKLongMedianDtau}{+0.69}

\providecommand{\SeedCount}{three}

\providecommand{\SeedMaxDev}{0.029}

\providecommand{\SpeedupCorrMinR}{0.97}

\providecommand{\GemmaArmBOverBSixteenMin}{+0.29}
\providecommand{\GemmaArmBOverBSixteenMax}{+0.98}
\providecommand{\GemmaArmBOverDflashMin}{+1.8}
\providecommand{\GemmaArmBOverDflashMax}{+5.1}

\providecommand{\GptOssTarget}{GPT-OSS-20B}

\providecommand{\GateArmBContGemmaN}{35}

\providecommand{\GateArmBContGemmaWithinDflareEightB}{+1.00}
\providecommand{\GateArmBContGemmaWithinDflashEightB}{+1.00}
\providecommand{\GateArmBContGemmaWithinDflareFourB}{+0.96}
\providecommand{\GateArmBContGemmaWithinDflashFourB}{+1.00}
\providecommand{\GateArmBContGemmaWithinGemmaTwelveB}{+0.64}

\providecommand{\QwenLadderDFlareSpeedGainHigh}{7.4}
\providecommand{\QwenLadderDFlareSpeedGainLow}{4.8}

\providecommand{\QwenLadderDFlareCeilingLast}{0.3}

\providecommand{\QwenLadderDFlareBenchesStepOne}{7}
\providecommand{\QwenLadderDFlareResolvedStepTwo}{7}
\providecommand{\QwenLadderDFlareBenchesStepTwo}{7}

\providecommand{\QwenLadderDFlashSpeedGainHigh}{8.5}
\providecommand{\QwenLadderDFlashSpeedGainLow}{3.7}

\providecommand{\QwenLadderDFlashCeilingLast}{0.1}

\providecommand{\QwenLadderDFlashResolvedStepTwo}{5}
\providecommand{\QwenLadderDFlashBenchesStepTwo}{7}

\providecommand{\QwenLadderLastStepCells}{14}
\providecommand{\QwenLadderLastStepInconclusive}{13}

\providecommand{\QwenLadderNative}{16}
\providecommand{\QwenLadderBenches}{7}
 \providecommand{\GateLooMeanAbs}{0.134}
\providecommand{\GateLooPredictions}{28}
\providecommand{\GateLooFolds}{4}
\providecommand{\GateLooGainMin}{+0.07}
\providecommand{\GateLooGainMax}{+1.02}
 \providecommand{\GateHeldoutQwenArmAN}{28}
\providecommand{\GateHeldoutQwenArmARho}{+0.87}
\providecommand{\GateHeldoutQwenArmARhoEn}{+0.92}

\providecommand{\GateHeldoutGemmaArmARhoEn}{+0.86}

\providecommand{\GateHeldoutGemmaArmABiasEn}{+0.09}
\providecommand{\GateHeldoutGptOssArmAN}{14}

\providecommand{\GateHeldoutGptOssArmARhoEn}{+0.46}

\providecommand{\GateHeldoutGptOssArmABiasEn}{+0.20}

\providecommand{\GateHeldoutQwenArmBRho}{+0.97}
\providecommand{\GateHeldoutQwenArmBRhoEn}{+0.95}

\providecommand{\GateHeldoutGemmaArmBRhoEn}{+0.68}

\providecommand{\GateHeldoutGemmaArmBBiasEn}{+0.19}

 \providecommand{\TOnePairs}{8}
\providecommand{\TOneArmBClear}{8}
\providecommand{\TOneArmAClear}{5}
\providecommand{\TOneArmAHalf}{3}
\providecommand{\TOneShareMin}{27}
\providecommand{\TOneShareMax}{69}

\providecommand{\TOneCycleArmBPositive}{8}
\providecommand{\TOneCycleArmAPositive}{4}
 \providecommand{\LossRerunConditions}{7}
\providecommand{\LossRerunWidestSpread}{0.054}
\providecommand{\LossRerunWidestBench}{MATH-500}
\providecommand{\LossRerunUniformBenches}{7}

\providecommand{\LossRerunUniformMaxDiff}{0.020}

\providecommand{\LossRerunPairs}{42}
\providecommand{\LossRerunPairsExcluding}{6}
\providecommand{\LossRerunPairsLargest}{0.052}
 \newcommand{\spell}[1]{\ifcase#1 zero\or one\or two\or three\or four\or five\or six\or seven\or eight\or nine\else#1\fi}
\newcommand{\Spell}[1]{\ifcase#1 Zero\or One\or Two\or Three\or Four\or Five\or Six\or Seven\or Eight\or Nine\else#1\fi}

\begin{document}

\maketitle

\begin{abstract}
Speculative decoding speeds up generation using an efficient draft model (drafter) to propose tokens that
a target model verifies in one pass, preserving the target's output distribution. High-acceptance
block-diffusion drafters, such as DFlash and DFlare, fill an entire block in one parallel pass. In many
cycles, the target model accepts all draft tokens in the block, so the drafter exhausts its trained
horizon before verification ever fails. We call this unrealized acceptance \emph{stranded speed-up}.
Evaluating a drafter by its per-prompt or per-cycle mean committed length alone hides this phenomenon,
whereas the acceptance histogram exposes this bottleneck as a spike in the \emph{ceiling bin}, the
fraction of cycles that accept the entire block.
We recommend reviewing the acceptance histogram before spending more training compute.
Naively widening the block at inference could reduce acceptance for drafters with bidirectional attention,
since, once the block outgrows its training size, the drafter's bidirectional
attention shifts its distribution even at early positions, which empirically degrades acceptance there.
We therefore post-train the drafter to a longer block with a short curriculum that
emphasizes the newly exposed positions, a method we call DBloom.
We continue training the pretrained DFlash and DFlare drafters at the original block size of 16,
before expanding them to 24 (B16 to B24) on \mbox{Qwen3-8B} and \mbox{Qwen3-4B}. Compared to the original B16 drafters,
this technique raises per-prompt committed length by up to $\FamilyMaxDtau$~tokens and decode speed-up by a median of
$\SpeedupChangeArmBMedian$ versus $\SpeedupChangeArmAMedian$ without the continuation.
On high-ceiling benchmarks, the speed-up rises to $\SpeedupChangeArmBHighCeilingMedian$.
On \GemmaTarget{} and \GptOssTarget{}, expansion also raises committed length on every
benchmark. Against JetSpec, a
tree drafter, DBloom has higher committed length by point estimate on every benchmark at tree budgets up to
\JetPointBeatsAll~nodes.
\end{abstract}

\section{Introduction}

\begin{figure}[htp]
\centering
\includegraphics[width=0.75\linewidth]{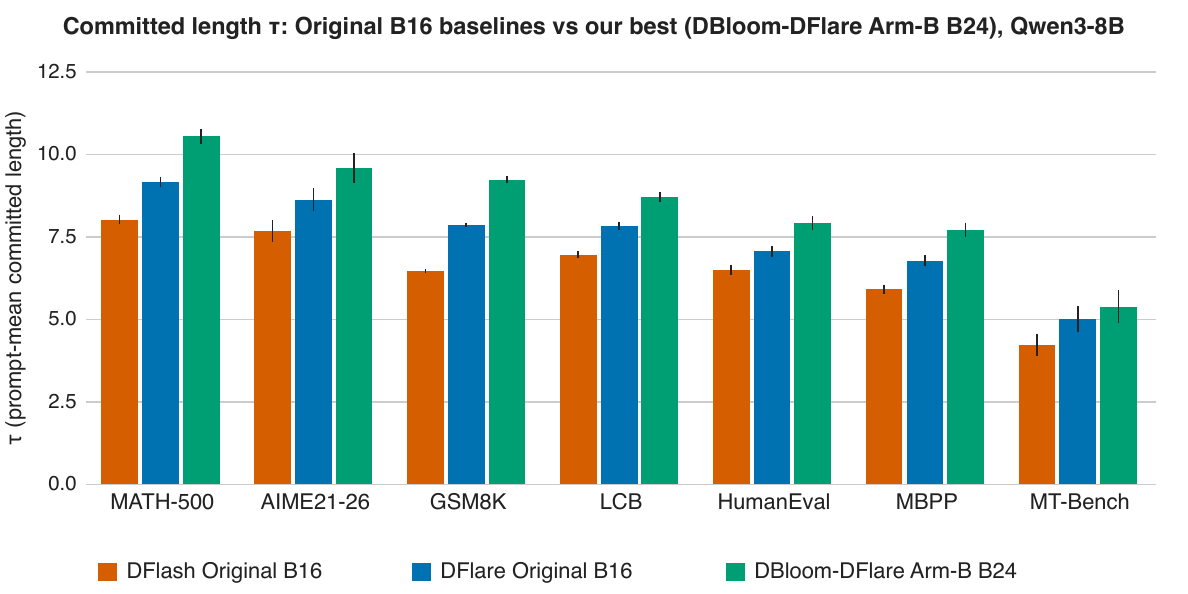}\\[1pt]
\includegraphics[width=0.75\linewidth]{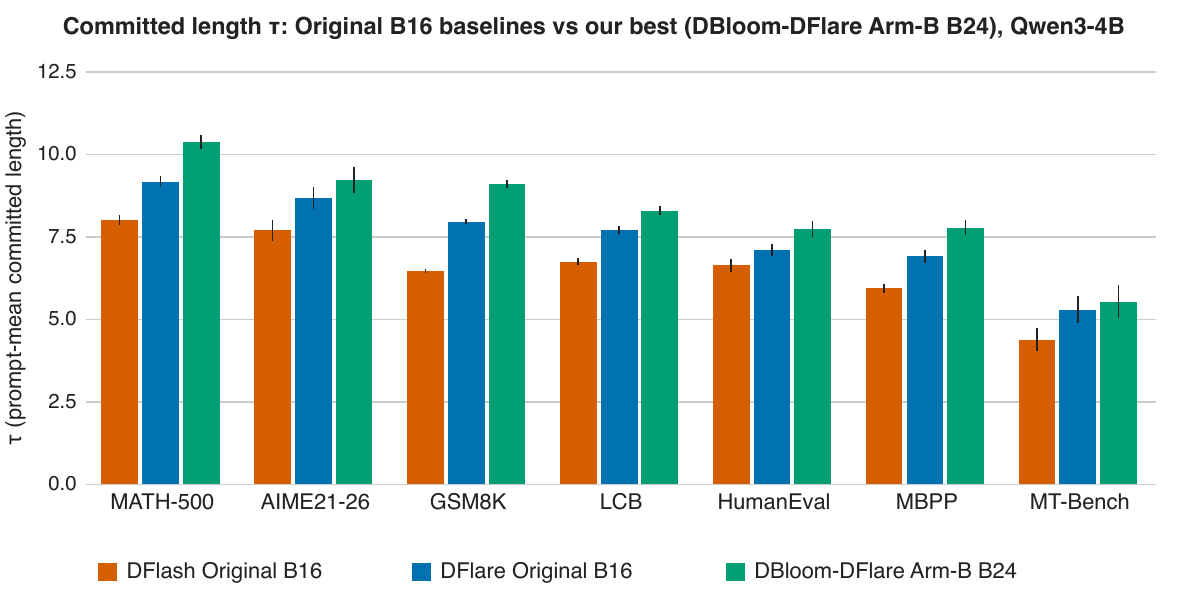}\\[1pt]
\includegraphics[width=0.75\linewidth]{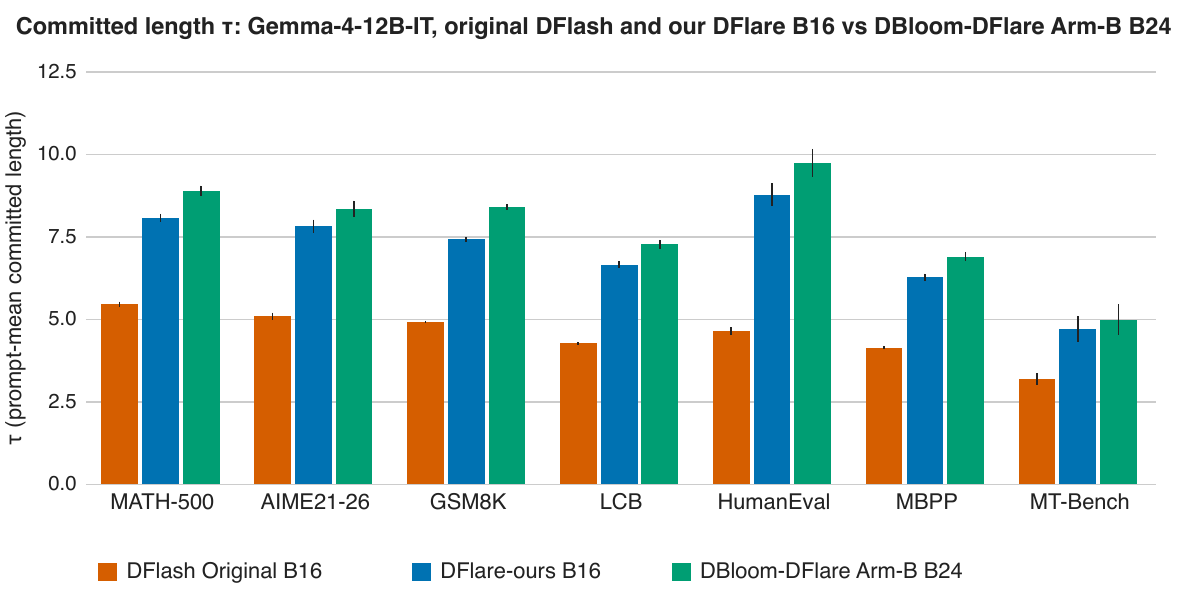}
\caption{\textbf{Block-horizon expansion raises committed length across benchmarks.} Per-benchmark
prompt-mean committed length $\tau$. Top and middle: the original DFlash and DFlare B16 drafters and our
DBloom-DFlare Arm-B B24 drafter, on Qwen3-8B and Qwen3-4B. Bottom: the model-family check on
\GemmaTarget{}, with the original DFlash and our DFlare B16 drafters against our DBloom-DFlare Arm-B B24.
For \GemmaTarget{}, the DBloom effect is the DFlare B16 to Arm-B B24 change.
The expanded drafter clears the baselines on the high-ceiling benchmarks and remains at or above baseline
on the low-ceiling ones. Whiskers are 95\% CIs.}
\label{fig:hero}
\end{figure}

Speculative decoding drafts several tokens and verifies them in one target pass, yielding a provably
identical distribution to the target's \citep{leviathan2023,chen2023}.
Open-weight \emph{autoregressive block-diffusion}\footnote{We hereafter write \emph{block-diffusion} for \emph{autoregressive block-diffusion}.}~\citep{bd3lm2025} drafters, such as DFlash
\citep{dflash2026} and DFlare \citep{dflare2026}, report a higher mean \emph{committed length}, the
tokens produced per target-and-drafter forward pass, than earlier autoregressive drafters such as
EAGLE and EAGLE-3 \citep{eagle2024,eagle3_2025}.
We extend DFlash and DFlare to recover additional speed-up by analyzing their acceptance histograms and
expanding the block only where the histogram is clipped at its ceiling, a method we call DBloom. Figure~\ref{fig:hero} shows the result.

In each decode cycle, a block-diffusion drafter fills a small block of tokens in a single pass, of
which the target accepts a verified prefix. Most evaluations in the literature
summarize acceptance by its mean, which cannot distinguish a drafter that often fills the block
from one repeatedly cut off at the block boundary, even though it could offer more acceptable tokens.
The per-cycle acceptance histogram can separate these cases because a spike in the \emph{ceiling bin},
the fraction of cycles in which the target accepts the entire block, indicates that
the target keeps accepting draft tokens until the drafter runs out of them.
DBloom expands the block to uncap this stranded acceptance.

Unfortunately, naively widening the diffusion block at inference does not recover stranded acceptance
(Section~\ref{sec:motivation})\footnote{Note that naive block expansion at inference for \emph{autoregressive drafters} does \emph{not} reduce acceptance.}. Expansion improves speed-up only after retraining with a larger block,
so a spike in the ceiling bin indicates that this retraining is worthwhile.

We make the following contributions.
\begin{enumerate}
\item \textbf{The ceiling bin flags stranded speed-up.} We show
that the mean committed length obscures block-limited acceptance, which the acceptance histogram's ceiling bin reveals.
Writing B$k$ for a block size of $k$ tokens, the ceiling-bin mass correlates with the
acceptance improvement from expanding the block from B16 to B24.
Specifically, the Spearman correlation coefficient is $\rho = \GateHeldoutQwenArmARho$ to
$\GateHeldoutQwenArmBRho$ on the four Qwen3~\citep{DBLP:journals/corr/abs-2505-09388,dflash2026,dflare2026}
drafters with which we develop our training recipe: Qwen3-8B and 4B using DFlash and DFlare drafters.
On two other families, \GemmaTarget{}~\citep{gemmateam2026gemma4technicalreport} and
\GptOssTarget{}~\citep{gpt-oss}, the association is positive but weaker; these small checks do not
establish reliable ranking across families.
\item \textbf{A short post-training step expands the block.}
Once a B16 drafter is trained, widening the block to B24 with a horizon-weighted loss over only about
30K prompts raises the per-prompt committed length by a median of $\FamilyIncrMedianDtau$ tokens
(up to $\FamilyIncrMaxDtau$) on the high-ceiling benchmarks, measured against the drafter from which the expansion starts.
That corpus is $\ExpFracDflareCont$\% of the DFlare training budget and $\ExpFracDflashCont$\% of the DFlash budget,
counting the drafter's own pretraining and continuation. The full recipe (Arm B) continues training at B16
before block expansion. Against the original B16 drafter, it improves decode speed-up by a median of
$\SpeedupChangeArmBMedian$. Expansion alone increases it by a median of $\SpeedupChangeArmAMedian$.
\item \textbf{A linear block of 24 exceeds trees of up to \JetPointBeatsAll{} nodes in committed-length point estimates.} As an out-of-design check,
we compare our DBloom-expanded drafters against JetSpec, a concurrent, high-acceptance tree-based
decoder we did not use to design DBloom. On Qwen3-8B, in a prompt-matched paired comparison,
both arms have higher committed-length point estimates on every benchmark at budgets up to \JetPointBeatsAll{}~nodes, and
the lead of our best drafter (Arm B) is statistically significant on every benchmark up to
\JetClearBeatsAllArmB{}~nodes. At \JetTopBudget{} nodes, Arm B wins significantly on \JetTopWinsWord{} benchmarks.
The other \JetTopTiesWord{} comparisons are inconclusive. Arm A loses significantly on one benchmark at that budget
(Section~\ref{sec:res-jetspec}).
\end{enumerate}

\section{Background and related work}
\label{sec:background}

\paragraph{Draft-model families and our baselines.} Speculative-decoding drafters split into
autoregressive and block-diffusion families.
On the autoregressive side, EAGLE autoregressively
predicts the target's second-to-top-layer features, whereas EAGLE-3 predicts tokens directly from
fused multi-layer features~\citep{eagle2024,eagle3_2025}. Both methods raise acceptance well above that of the
original speculative decoders~\citep{chen2023,leviathan2023}, but retain sequential dependence across
draft depth.
Domino combines a parallel drafter backbone with a lightweight sequential causal-correction head,
retaining causal dependence while greatly reducing the cost of autoregressive draft
execution~\citep{domino2026}.
JetSpec~\citep{jetspec2026} trains a causal parallel draft head on the
target's fused hidden states and produces a path-conditioned candidate
\emph{tree} in a single forward pass. It uses a tree-causal attention mask so
each node depends on its ancestor tokens along the branch. Since the tree is
drafted in one parallel pass, increasing the node budget mostly increases
tree-selection and target-verification work instead of adding autoregressive
draft depth. We compare against this method as a strong parallel
tree-drafting baseline, and Section~\ref{sec:res-jetspec} shows that our
linear B24 drafter achieves comparable or higher committed length in our
evaluation.
Block-diffusion drafters
instead build on the block-diffusion language model framework~\citep{bd3lm2025}, which denoises a
sequence in blocks with parallel sampling. Applied to speculative decoding,
DFlash~\citep{dflash2026} and DFlare~\citep{dflare2026} reserve position 0 of each decode block as an
\emph{anchor}, the bonus token the target committed last cycle, denoise the other $B{-}1$ positions in
one pass, and let the target accept a prefix of length $n$, $0 \le n \le B{-}1$, before committing the
next anchor.

We test DBloom on two high-acceptance block-diffusion drafters, the original DFlash and DFlare drafters, and compare it against JetSpec,
the strongest tree-based decoder in our setting. DFlash is weaker than DFlare,
at least in the evaluated Qwen3 settings, but is the better-known, peer-reviewed
drafter, so we expand it as well. The DFlash authors have already examined the signal we build on.
Their Table 8 reports that a block-8 drafter fully accepts the block $35.7\%$ of the time on MATH-500,
which they interpret as evidence that the block horizon is too short in many cycles. They train a separate drafter for each block size,
however, and find that a drafter generalizes to a \emph{smaller} inference block but not a larger one,
leaving adaptive block sizing to future work. We instead turn the full-block fraction into a
cross-benchmark indicator of whether the block size limits acceptance (Section~\ref{sec:gate}),
and recover the lost acceptance by post-training an original drafter to the larger block with little added data
(Section~\ref{sec:method}).

\paragraph{Scaling the draft budget.} A separate line increases acceptance by spending more draft
budget per step. JetSpec grows its candidate tree to as many as 256 nodes to increase committed
length, leaving open whether the original drafter is already throttled by its block size.
DDTree is the block-diffusion counterpart of
this idea, constructing a draft tree from a block-diffusion drafter's per-position distributions and
selecting, under a fixed node budget, the continuations most likely to match the target
\citep{ddtree2026}. The two methods pull different levers.
DDTree spends more budget as a tree at a \emph{fixed} block size, whereas we expand the block \emph{horizon} by retraining and keep a linear budget of 24. The methods remain complementary since a horizon-expanded drafter could itself be drafted as a tree.
Two concurrent works refine this budget lever in ways that complement our horizon expansion. CaDDTree
argues that acceptance length alone cannot justify a node budget, and instead selects tree
structure and budget to optimize throughput directly, modeling draft and verification latency per round
\citep{caddtree2026}. Our indicator addresses a complementary upstream decision, whether retraining to enlarge the block horizon is worthwhile. After such expansion, a CaDDTree-style method could still optimize tree structure and node budget at inference. BASTION builds query-dependent trees over a block-diffusion drafter under a latency
budget \citep{bastion2026}. Like CaDDTree, it keeps the trained block fixed and varies the tree, so it
applies on top of the wider block we train.

\section{Why naive block expansion rarely increases speed-up}
\label{sec:motivation}

If a large ceiling bin of a block-diffusion drafter's histogram indicates the block is the bottleneck, the obvious next step is to increase the block size at inference without retraining. In every cell we measured, doing so reduced speed-up.

Table~\ref{tab:naive-ordering} measures this across benchmarks and target sizes. As naive B24 loses
speed-up, the ceiling-bin mass drops to near zero
everywhere (the native and naive ceiling columns of Table~\ref{tab:naive-massshift}). On the 8B targets, accepted
length barely changes, and speed-up drops by only a few percent because the larger drafter still predicts
most of the block. On the 4B targets, the front rarely survives to the old boundary, so almost no mass
reaches the new positions, and speed-up falls sharply (DFlare Qwen3-4B on MATH-500 from
$\NaiveFourBMathNativeSp\times$ to $\NaiveFourBMathNaiveSp\times$, a $\NaiveFourBMathDropPct\%$ drop in
speed-up).

\begin{table}[tbp]
\centering
\small
\caption{Naive block expansion loses speed-up, one row per drafter cell. Mean committed length $\tau$ over the seven benchmarks at B16 and at B24 without training, and the mean change in speed-up.}
\label{tab:naive-ordering}
\begin{tabular}{lrrr}
\toprule
drafter & \makecell{native \\ B16 \\ $\tau$} & \makecell{naive \\ B24 \\ $\tau$} & \makecell{naive \\ B24 \\ speed-up \\ $\Delta$} \\
\midrule
DFlare-8B & 7.48 & 7.28 & -3\% \\
DFlare-4B & 7.54 & 3.45 & -51\% \\
DFlash-8B & 6.54 & 6.27 & -4\% \\
DFlash-4B & 6.55 & 4.43 & -30\% \\
\bottomrule
\end{tabular}
\end{table}
 
Naive expansion raises committed length on no cell. On the 8B target, a modest ceiling survives past the
old boundary, so committed length stays about flat (DFlare Qwen3-8B on MATH-500, $E[n]{+}1$ dips only
from $\NaiveEightBMathNativeEnp$ to $\NaiveEightBMathNaiveEnp$), whereas on the 4B target the ceiling
vanishes and committed length drops ($\NaiveFourBMathNativeEnp$ to $\NaiveFourBMathNaiveEnp$ on
MATH-500). Even where the 8B committed length barely moves, speed-up still falls (MATH-500
$\NaiveEightBMathNativeSp\times$ to $\NaiveEightBMathNaiveSp\times$) because each cycle costs more and
the front erodes.

A block-diffusion drafter fills a block in one pass, so a B16 drafter has learned only to unmask a
B16 horizon. Run it at B24, and the proposal distribution shifts even at early positions.
Empirically, front-of-block verification then erodes.
 The front of the block therefore sets the committed length, and if an early
position fails, the tail never matters. Reliable gains therefore require retraining at the larger block.

\section{Method}
\label{sec:method}

The original DFlash and DFlare drafters are B16. Our method, DBloom, post-trains them to B24, guided by a
quick diagnostic that indicates where a larger block will boost acceptance. It inspects the
acceptance histogram the original drafter already produces, so it costs one evaluation pass and no
training. We start there, then turn to the curriculum that acts on the positions the diagnostic flags.

\subsection{The acceptance histogram and its ceiling bin}
\label{sec:gate}

Most evaluations of block-diffusion drafters report the mean committed length and little about its
distribution. We instead record the full per-cycle acceptance histogram, the empirical distribution of
$n \in \{0, 1, \ldots, B{-}1\}$ across decode cycles. The bin at $n = B{-}1$ is the \emph{ceiling
bin}, the cycles where the target accepts all $B{-}1$ draft tokens, and its mass is the
\emph{block-ceiling fraction}.

The ceiling bin is the rightmost bin, and a spike there indicates acceptance is \emph{ceiling-clipped}.
A cycle ends for one of two reasons. When acceptance stops at some $n < B{-}1$, the target has rejected a
proposed token and discarded the remaining draft tokens. When it reaches $n = B{-}1$, no rejection has occurred within the offered horizon, and the block boundary ends the cycle and censors the count at
$B{-}1$. A large ceiling bin therefore shows that the original drafter is block-limited and that
enlarging the block can recover the clipped acceptance. Appendix~\ref{app:blocksize} makes this precise,
expressing accepted length as a censored sum whose ceiling mass is the clipped part.

We measure this ceiling bin on the original drafter before spending training compute, and a high ceiling
indicates where expansion is likely to boost acceptance. Where the ceiling is high, the block is the binding constraint, so
expanding the drafter to a larger block converts ceiling-bin mass into longer accepted prefixes. Where it
is low, the block rarely limits acceptance, so expansion has little to recover, and we leave those
low-ceiling domains out of the expansion corpus (Section~\ref{sec:setup}). Expansion still
produces one B24 drafter that we evaluate on every benchmark, so the ceiling bin indicates where a drafter
gains from expansion. Section~\ref{sec:results} quantifies
how well the baseline block-ceiling fraction tracks the realized gain.

\subsection{Curriculum post-training and the two arms}
\label{sec:curriculum}

Section~\ref{sec:motivation} showed that running the original B16 drafter at B24 without retraining loses speed-up, so post-training is a prerequisite for expansion. When the indicator points to expansion, DBloom post-trains the original drafter. We initialize from the original B16 weights and fine-tune at B24, so the positions the B16 drafter already learned (block offsets $1$ through $15$) start from a competent drafter, and the eight new tail positions (offsets $16$ through $23$) must be learned. The drafter conditions on the target's fused hidden states, which are independent of the block size, so the same target responses can label a drafter at any block size. Both steps train on self-distilled prompts and target responses drawn from the separate corpora of Table~\ref{tab:corpora}.

We use a two-level (flat-step) horizon-weighted loss with unit weight on the old positions and a constant $3\times$ boost on the new tail positions. An ablation over \spell{\LossRerunConditions} per-position weighting schemes, including a uniform weight without any boost, finds that these options change the committed length by at most $\LossRerunWidestSpread$~tokens on any benchmark (Appendix~\ref{app:lossablation30k}). Because acceptance stops at the first mismatch, the front of the block determines the result before the up-weighted tail is reached. We keep the flat step because every drafter in this paper uses it.

We reach the expanded B24 drafter by one of two arms that share this short curriculum, corpus, and knobs and
differ only in the starting point.
\textbf{Arm A} applies the expansion above to the original B16
drafter in a single post-training step.
\textbf{Arm B} first runs a B16 continuation fine-tune to strengthen the domains
the original corpus under-represented, then expands that continued drafter with
the same step as Arm A.
DBloom is a horizon-expansion operator. It takes whatever B16 drafter it is given, reads that
drafter's acceptance histogram, and applies the same 30K-prompt expansion. The two arms differ only in
the drafter handed to that operator. Arm A expands the original checkpoint directly, which tests
data-efficient expansion of a public model. Arm B first strengthens the drafter with a same-block
continuation, then expands the result, which tests whether the horizon still limits acceptance once the
native-block drafter is stronger. We therefore judge each expansion against the drafter it actually
expands. The incremental effect of DBloom is the gain over its input B16 (the original B16 for Arm A,
the continuation-trained B16 for Arm B), while the improvement over the original B16 is the
end-to-end system result.

\section{Experimental setup}
\label{sec:setup}

\subsection{Evaluation design}

We evaluate a controlled two-by-two grid of two original block-diffusion drafter architectures, DFlash
\citep{dflash2026} and DFlare \citep{dflare2026}, each paired with two targets, Qwen3-8B and
Qwen3-4B. Crossing drafter architecture with target scale means that a finding that stays valid across all four
cells is less likely to be an artifact of one drafter or one target size.
We fix every design choice: the training corpus and the domains it includes, the per-position loss
weighting, the position where the tail boost starts, the learning rate and the epoch count.
Only the block size varies, following each family's original value.
We expand two other model families, \GemmaTarget{} (Section~\ref{sec:res-gains}) and \GptOssTarget{},
with the same expansion settings to test the association across families. This holds the recipe fixed,
not the target, tokenizer or drafter, so it does not isolate the ceiling bin causally.

Alongside this grid, we compare against JetSpec~\citep{jetspec2026}, a concurrent tree-based drafter. We designed the expanded drafters without referencing JetSpec, so the comparison is out of design.
JetSpec releases a
single Qwen3-8B draft head, so this comparison is on Qwen3-8B, and we run it at each of
its tree-node budgets (Section~\ref{sec:res-jetspec}).

Rather than transcribing reported values, we evaluate every drafter ourselves under one consistent protocol.
All decoding is greedy at temperature $0$ (except the temperature-$1$ check of Appendix~\ref{app:t1}) with a maximum
generation length of 16{,}384 tokens. We use the full test set of every benchmark, then keep the EOS-terminated responses (Section~\ref{sec:metrics}).
The benchmarks and their full sizes are GSM8K (1{,}319), MATH-500 (500), the AIME21-26
competition-math set (179), HumanEval (164), MBPP (257), LiveCodeBench (1{,}055), and MT-Bench (80).
MT-Bench is two-turn, so its 80 questions yield 160 scored responses.
We evaluate both turns, generating the second from the target's
own first-turn reply, so a result from first turns alone is not comparable to ours.

Because we evaluate the original drafters ourselves, we are not tied to the
30-problem AIME25 subset the source papers quote, and the larger AIME21-26 set yields tighter
acceptance estimates. Accuracy contamination is not the quantity we measure as committed length depends
only on how well the drafter tracks the target on a given prompt.

\subsection{Metric conventions}
\label{sec:metrics}

We report two committed-length summaries per decode cycle and keep them distinct throughout. Let a prompt
$p$ run $c_p$ decode cycles with accepted counts $n_{p,1}, \ldots, n_{p,c_p}$, and let
$\bar{n}_p = \frac{1}{c_p}\sum_i n_{p,i}$ be its per-prompt mean accepted count. Over $P$
prompts,
\begin{equation}
\tau = 1 + \frac{1}{P}\sum_{p} \bar{n}_p,
\qquad
E[n]{+}1 = 1 + \frac{\sum_p \sum_i n_{p,i}}{\sum_p c_p}.
\end{equation}
Each counts the tokens the target commits per cycle, namely the $n$ accepted draft tokens plus the one bonus
token, so we call $\tau$ the \emph{prompt-mean committed length} and $E[n]{+}1$ the \emph{cycle-mean
committed length}. This per-cycle count is the standard speculative-decoding throughput measure
\citep{leviathan2023,chen2023}. The two metrics differ only in weighting. $\tau$ takes the prompt as the
statistical unit, while $E[n]{+}1$ averages over all cycles equally. They can diverge by more than
\MetricMaxDivergenceWord{}, most often with $E[n]{+}1$ below $\tau$ when long generations draft poorly,
so we report both statistics and keep them separate. Appendix~\ref{app:tau} provides the exact identity and
real examples.
We bootstrap prompts for absolute committed lengths (per-prompt means for $\tau$, cluster/ratio
estimates for $E[n]{+}1$) and use paired prompt bootstraps for $\Delta\tau$, with 95\% intervals.
Speed-up and paired cycle-mean changes are point estimates. Speed-up is speculative decoding throughput
times the run's stored target-only time per token.

\paragraph{Committed length and speed-up.}
With $L_B$ and $D_B$ the total output tokens and decode time on the selected responses, we report
$S_B=(L_B/D_B)\,t_{\text{target},B}$, where $t_{\text{target},B}$ is the run's stored target-only time per token.
Ignoring terminal-token boundary effects, $L_B/D_B\approx(E_B[n]{+}1)/t_{\text{cycle},B}$, where $t_{\text{cycle},B}$ is the mean decode-cycle time, including drafting and verification, so
$S_B/S_{B'}\approx\frac{E_B[n]{+}1}{E_{B'}[n]{+}1}\frac{t_{\text{cycle},B'}}{t_{\text{cycle},B}}\frac{t_{\text{target},B}}{t_{\text{target},B'}}$,
whose last factor cancels only for a common target-only reference. On Qwen3, over \CycleTimeRatioCount{} cells
(four target-drafter combinations, two arms and seven benchmarks), the B24-to-original-B16 cycle-time ratio
has median \CycleTimeRatioMedian{} and standard deviation \CycleTimeRatioSd\%, so cycle cost is near-constant
here, and the reported speed-up changes also include variation in the target-only reference
(Table~\ref{tab:speedup-change}). All timings are single-stream decode measurements on our unoptimized
research code. We have not evaluated batched-serving performance.

\paragraph{Acceptance is measured over EOS-terminated responses.} A response that never emits an
end-of-sequence (EOS) token runs to the generation-length cap, usually by repeating a phrase. A repeated
phrase is easy to predict, so the target accepts near the block ceiling, which inflates the measured acceptance on the
benchmarks where these runaways occur. We therefore compute $\tau$,
$E[n]{+}1$, and the block-ceiling fraction over only the EOS-terminated responses. How many responses a
run keeps depends on its target, drafter, and decoding temperature, and in our runs the drop
concentrated on competition math. Among the greedy DFlash and DFlare runs underlying the per-cell expansion tables, the minimum retained count, computed separately for each benchmark, was $\EosNMinAime$ of
$\EosNTotalAime$ responses on AIME21-26, $\EosNMinMathFiveHundred$ of $\EosNTotalMathFiveHundred$ on
MATH-500, and $\EosNMinLcb$ of $\EosNTotalLcb$ on LiveCodeBench, with nearly all responses kept on the
four short-generation benchmarks. We recompute speculative throughput on these same responses.
The target-only normalization is the stored whole-run scalar, not a reference remeasured on the EOS subset.

\subsection{Training data by arm and step}

Both arms start with the original B16 drafter and then post-train it. All corpora are
self-distilled, so the target regenerates its own responses, and the DFlash and DFlare drafters for a
given target share a single corpus per step. Each corpus includes one response for each unique English-only prompt,
and is decontaminated against the evaluation sets using 32-gram matching.

The Arm-A expansion and the identical step-2 expansion in Arm B use a 30K-prompt
corpus drawn exclusively from high-ceiling domains, since block expansion adds acceptance only where the B16 drafter
already saturates the block. We exclude chat and short-form code because their benchmarks have little
ceiling-bin mass to convert, so expansion adds little there. At the shipped 1{,}876-step budget,
30K and 100K prompts yield nearly identical median gains. At 6{,}252 steps, 100K improves modestly on 30K.
This result supports the 30K operating point, which we do not claim as a universal saturation threshold (Appendix~\ref{app:corpussweep}).

\begin{table}[t]
\centering
\small
\caption{The two arms (left) and their training corpora (right). Both arms reach a B24 drafter from the same
original B16 drafter and share the \emph{identical} B24 expansion, so the only difference is Arm B's
same-block B16 continuation.}
\label{tab:corpora}
\begin{minipage}[c]{0.44\linewidth}
\centering
\includegraphics[width=\linewidth]{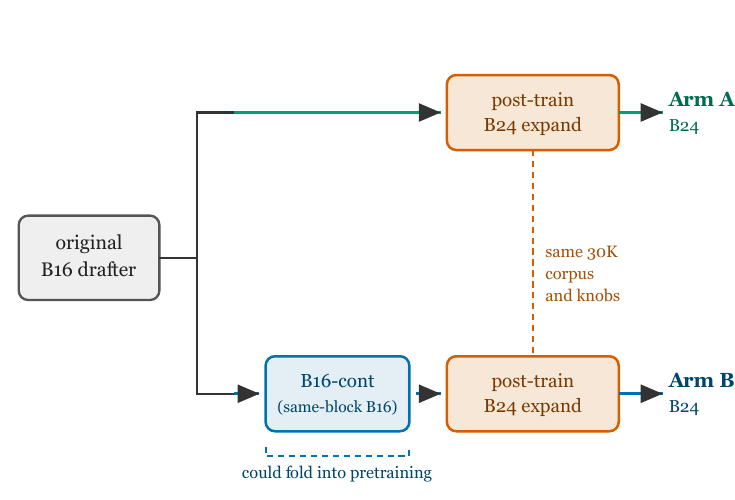}
\end{minipage}\hfill
\begin{minipage}[c]{0.54\linewidth}
\centering
\begin{tabular}{llr}
\toprule
Step & Source domains & Prompts \\
\midrule
\multicolumn{3}{l}{\emph{Arm A, and Arm-B step 2: from B16 to B24}} \\
\quad math & OpenMathInstruct-2 & 20{,}000 \\
\quad code & rStar-Coder (long competitive) & 10{,}000 \\
\quad \textbf{total} & & \textbf{$\sim$30{,}000} \\
\midrule
\multicolumn{3}{l}{\emph{Arm-B step 1: B16 continuation}} \\
\quad math & OpenMathInstruct-2 & 250{,}000 \\
\quad code & \makecell[l]{opc-sft-stage2, rStar-Coder,\\ OpenCodeInstruct} & 300{,}000 \\
\quad STEM & Nemotron-v2 STEM (replay) & 60{,}000 \\
\quad chat & Nemotron-v2 chat (replay) & 60{,}000 \\
\quad \textbf{total} & & \textbf{$\sim$670{,}000} \\
\midrule
\multicolumn{3}{l}{\emph{Loss-weighting ablation only (Appendix~\ref{app:lossablation})}} \\
\quad math & OpenMathInstruct-2 & 25{,}000 \\
\quad code & CodeAlpaca & 15{,}000 \\
\quad chat & WildChat-1M (first user turn) & 20{,}000 \\
\quad \textbf{total} & & \textbf{$\sim$60{,}000} \\
\bottomrule
\end{tabular}
\end{minipage}
\end{table}

The Arm-A expansion uses a two-level (flat-step)
horizon-weighted loss with boundary 16, a $3\times$ boost on the new tail positions, no decay, a learning rate
of $1\times10^{-5}$, four epochs, and a global batch size of 64. The Arm-B continuation is a same-block (B16)
fine-tune at a learning rate of $5\times10^{-5}$ for two epochs, with the STEM and chat slices acting as
anti-forgetting replay so that strengthening weak domains does not erode domains the original drafter
already handled well. Arm-B step 2 then reuses the Arm-A expansion corpus and knobs, changing only
the initialization. Because this continuation stays at the native B16 horizon and only rebalances
under-represented domains, the original DFlash and DFlare authors could have folded an equivalent step
into their own pretraining before releasing a B16 drafter.

Block-horizon expansion is cheaper than building the B16 drafter it starts from. The
$\sim$30K expansion corpus in Table~\ref{tab:corpora} is $\ExpFracDflare$\% of the $\sim$2.4M prompts
that trained DFlare and $\ExpFracDflash$\% of the $\sim$800K that trained DFlash
\citep{dflare2026,dflash2026}. Treating the continuation-trained B16 as the model we actually expand,
and counting its full training budget (original pretraining plus our $\sim$670K same-block
continuation), the same 30K expansion is $\ExpFracDflareCont$\% of the DFlare budget and
$\ExpFracDflashCont$\% of the DFlash budget. These fractions count distinct corpus prompts: the expansion adds
$\ExpFracDflare$--$\ExpFracDflash$\% relative to original pretraining, or about 1\% to 2\% relative to the combined pretraining and continuation corpus size.
The expansion starts from a drafter trained at its native block size and fine-tunes it at the larger block size.
A control shows that the extra 30K prompts alone do not account for the improvement. Post-training the original
B16 drafter on the same corpus at block size 16 adds only $\ControlDataOnlyMin$ to $\ControlDataOnlyMax$
tokens, versus $\ControlExpansionMin$ to $\ControlExpansionMax$ for the full B24 expansion
(Appendix~\ref{app:control}, Qwen3-8B DFlare).
For Qwen3 that B16 drafter is the original checkpoint. For Gemma-4-12B-IT,
which does not have an original DFlare drafter, we first trained the B16 DFlare drafter ourselves and then applied
the same expansion.
We compare against DFlare as the stronger architecture
(the DFlare authors ablate the two at a matched data budget and DFlare still wins~\citep{dflare2026})
and report DFlash as a second architecture to show the expansion also lifts a weaker base drafter.

\subsection{Reproduction gate}

As an external reference, we compare our B16 evaluation with the source papers' (DFlash and
DFlare) published acceptance for the original drafters. For each cell with a published value, we record our
$\tau$ with its 95\% bootstrap confidence interval and whether the published value falls
within that interval (Appendix~\ref{app:per-cell-results}). The source papers evaluate under a shorter
generation-length cap, whose effect Appendix~\ref{app:repro-cap} isolates. A published value outside
our interval reflects that protocol difference. For competition math,
the AIME21-26 set we report elsewhere has no published counterpart, so the reference row uses the
source papers' AIME25, which we also evaluate.

\subsection{Losslessness}
Greedy speculative decoding commits exactly the token the target would select at each step, so
block-horizon expansion changes only the drafter and therefore only decoding speed. The greedy
$\operatorname*{argmax}$ is not bit-identical across implementations, though, since batching and
floating-point reduction order can flip a near-tie and shift where a run reaches end-of-sequence, so
EOS-terminated counts vary slightly across runs. We therefore compute $\tau$ and $E[n]{+}1$ over only the
EOS-terminated responses, which the runaways would otherwise distort
(Section~\ref{sec:metrics}, Appendix~\ref{app:eos}).

\section{Results}
\label{sec:results}

We present three results corresponding to our three contributions.
The ceiling bin indicates where expansion can recover acceptance
(Section~\ref{sec:res-gate}). Data-efficient block-horizon expansion increases acceptance and speed-up on the
high-ceiling benchmarks across the grid (Section~\ref{sec:res-gains}). Against JetSpec at \JetTopBudget{} nodes,
the paired committed-length comparison favors Arm B significantly on \JetTopWinsWord{} benchmarks and is
inconclusive on \JetTopTiesWord{} (Section~\ref{sec:res-jetspec}). It does not establish equivalence.
The full per-benchmark
tables for all four cells and both arms are in Appendix~\ref{app:per-cell-results}. Unless noted otherwise, the
summary here compares our best configuration, the DBloom-DFlare Arm-B B24 drafter, against the
original B16 drafter.

Our evaluation reproduces the source papers' published B16 acceptance closely on the short-generation
benchmarks, where the evaluation protocols coincide. The larger gaps are confined to long-generation
math and follow from our longer maximum generation length. Every conclusion below is
a comparison at a fixed decoding budget, so it is valid regardless of how our absolute numbers
compare with the published ones. Instead of comparing averages, we test whether each published value falls
within our 95\% bootstrap interval. Appendix~\ref{app:repro-cap} presents the per-cell reproduction tables,
the percentage gaps, and a matched-protocol control that isolates the maximum generation length.

\subsection{The ceiling bin ranks where expansion gains most acceptance}
\label{sec:res-gate}
The ceiling-bin fraction tracks how much committed length improves after expansion.
Expanding to a larger block size yields higher committed length for benchmarks with a higher pre-expansion ceiling-bin fraction (Figure~\ref{fig:gate}).

For each arm, we fit the indicator on the four Qwen3 target-drafter combinations and evaluate it on
other-family data excluded from fitting. Gemma Arm A already appeared in an earlier version of this paper, so its
exclusion from the fit is retrospective. We fit committed-length increase against the pre-expansion
block-ceiling fraction on the $\GateHeldoutQwenArmAN$ Qwen3 points (four drafters, seven benchmarks). The Spearman correlation of the cycle-mean
($E[n]{+}1$) improvement is $\GateHeldoutQwenArmARhoEn$ for Arm A and $\GateHeldoutQwenArmBRhoEn$ for Arm B, and that of the
prompt-mean ($\tau$) improvement, which the DFlash, DFlare and JetSpec papers report, is $\GateHeldoutQwenArmARho$ and
$\GateHeldoutQwenArmBRho$. These points come from four related drafters and are not independent, so we do not compute confidence intervals for these exploratory correlations.

On \GemmaTarget{} and \GptOssTarget{} (Figure~\ref{fig:gate}, Table~\ref{tab:gate-heldout}), the ceiling-bin mass correlates less strongly with the cycle-mean ($E[n]{+}1$) improvement than on Qwen3.
The Spearman correlations are $\GateHeldoutGemmaArmARhoEn$ and
$\GateHeldoutGemmaArmBRhoEn$ on \GemmaTarget{} (Arms A and B) and $\GateHeldoutGptOssArmARhoEn$ on \GptOssTarget{} (Arm A, the two B12
drafters, $\GateHeldoutGptOssArmAN$ points). Only the Gemma Arm-A correlation exceeds its 5\% critical value. On all three
sets the line is above the measured increase, by
$\GateHeldoutGemmaArmABiasEn$, $\GateHeldoutGemmaArmBBiasEn$ and $\GateHeldoutGptOssArmABiasEn$ tokens on average.
These small checks on the same seven benchmarks suggest transfer, most clearly for Gemma Arm A,
but establish neither a calibrated gain estimate nor reliable cross-family ranking.

We use the ceiling fraction as an exploratory indicator of expansion benefit.
Appendix~\ref{app:block-ceiling-gate} presents the combined scatters, and Appendix~\ref{app:gateloo} adds a check that holds out one Qwen3 drafter at a time.

\subsection{Expansion raises acceptance and speed-up across the grid}
\label{sec:res-gains}

We evaluate two claims separately. Arm A tests the 30K-prompt data-efficiency claim, while Arm B shows
the best attainable result after an additional same-horizon continuation.
Block-horizon expansion boosts committed length and speed-up on the
high-ceiling benchmarks across all four cells, and the short 30K-prompt recipe captures those gains. Figure~\ref{fig:hero} (top and middle panels)
shows per-benchmark $\tau$ for the original DFlash
and DFlare B16 drafters alongside our best DBloom-DFlare Arm-B B24 drafter, on Qwen3-8B and Qwen3-4B. The expanded
drafter clears both baselines by a wide margin on the high-ceiling benchmarks and remains at or
just above baseline on the low-ceiling ones, consistent with the pattern the
ceiling fraction anticipates. At the smaller 4B scale, the same pattern applies, though the ordering
among the high-ceiling benchmarks is not strict (GSM8K gains more than AIME21-26).

\begin{table}[h]
\centering
\small
\caption{Expansion gain across all seven benchmarks (per-prompt committed length $\tau$, EOS-terminated prompts). Cont B16 is our continuation-trained B16 drafter. The first $\Delta\tau$ is the change over Cont B16, and the second is the change over the original B16.}
\label{tab:summary}
\resizebox{\ifdim\width>\linewidth\linewidth\else\width\fi}{!}{\begin{tabular}{lrrrrrrrrrr}
\toprule
\multirow{2}{*}{Benchmark} & \multicolumn{5}{c}{Qwen3-8B} & \multicolumn{5}{c}{Qwen3-4B} \\
\cmidrule(lr){2-6} \cmidrule(lr){7-11}
 & \makecell{Original \\ B16 \\ $\tau$} & \makecell{Cont \\ B16 \\ $\tau$} & \makecell{Arm-B \\ B24 \\ $\tau$} & \makecell{$\Delta\tau$ \\ (over \\ Cont \\ B16)} & \makecell{$\Delta\tau$ \\ (over \\ Original \\ B16)} & \makecell{Original \\ B16 \\ $\tau$} & \makecell{Cont \\ B16 \\ $\tau$} & \makecell{Arm-B \\ B24 \\ $\tau$} & \makecell{$\Delta\tau$ \\ (over \\ Cont \\ B16)} & \makecell{$\Delta\tau$ \\ (over \\ Original \\ B16)} \\
\midrule
GSM8K & 7.86 & 8.41 & 9.23 & +0.82 & +1.37 & 7.96 & 8.45 & 9.10 & +0.65 & +1.14 \\
MATH-500 & 9.17 & 9.42 & 10.54 & +1.12 & +1.37 & 9.18 & 9.30 & 10.39 & +1.08 & +1.21 \\
AIME21-26 & 8.63 & 8.53 & 9.58 & +1.05 & +0.95 & 8.68 & 8.50 & 9.22 & +0.72 & +0.55 \\
HumanEval & 7.07 & 7.33 & 7.93 & +0.59 & +0.86 & 7.10 & 7.29 & 7.74 & +0.45 & +0.63 \\
MBPP & 6.77 & 7.41 & 7.72 & +0.31 & +0.94 & 6.91 & 7.49 & 7.78 & +0.29 & +0.87 \\
LCB & 7.84 & 8.05 & 8.71 & +0.65 & +0.87 & 7.71 & 7.89 & 8.29 & +0.40 & +0.58 \\
MT-Bench & 5.00 & 5.11 & 5.38 & +0.27 & +0.38 & 5.30 & 5.37 & 5.53 & +0.16 & +0.23 \\
\bottomrule
\end{tabular}
}
\end{table}
 
Table~\ref{tab:summary} presents the gains numerically for all seven benchmarks per
cell, separating DBloom's incremental effect (the expansion over its input B16) from the end-to-end
system improvement (over the original B16). Across the grid, the full DBloom-DFlare Arm-B pipeline
raises $\tau$ by $\SummaryDtauMin$ to $\SummaryDtauMax$ tokens over the original B16 drafter end to end,
of which the 30K-prompt B24 expansion contributes the ``over Cont B16'' column. Arm A shows the same
expansion applied directly to the original checkpoint.
Within each fixed target-drafter pair, cycle-mean committed length tracks
measured decode-only speed-up with high fidelity
(Appendix~\ref{app:tau}, Table~\ref{tab:speedup-corr}).
The cycle-mean $E[n]{+}1$ rises with $\tau$ on those high-ceiling cells. The two statistics can
still diverge on a low-ceiling benchmark. On MT-Bench, a few 4B cells raise $\tau$ while $E[n]{+}1$ dips
slightly because long generations that draft poorly pull the cycle-mean below the per-prompt one (the
divergence of Appendix~\ref{app:tau}). We therefore report both statistics per cell without claiming that
they track each other.

The improvement goes beyond the DBloom-DFlare Arm-B B24 drafter. Aggregating all four cells and both arms over the three
high-ceiling benchmarks yields $4\times2\times3=\FamilySize{}$ comparisons. Judged against the drafter each
B24 actually expands (Arm A over the original B16, Arm B over the continuation-trained B16), the median
$\Delta\tau$ is $\FamilyIncrMedianDtau$ (up to $\FamilyIncrMaxDtau$). Judged end to end against
the original B16, the median is $\FamilyMedianDtau$ (up to $\FamilyMaxDtau$). A Holm correction over either
family (Appendix~\ref{app:holm}) leaves all \FamilyPositiveAfterHolm{} significantly positive with
\FamilyNegativeAfterHolmWord{} significantly negative.

The low-ceiling benchmarks, where the ceiling fraction indicates little room
for expansion, show less improvement. There, the Arm-A expansion yields only small gains, and the smallest is within noise.
On DFlash Qwen3-4B, the MBPP $\Delta\tau$ is \LowCeilNullMbppDtau{}, an interval that includes zero, so we
report it as no measurable change. Arm B lifts these benchmarks more,
but that gain comes almost entirely from the B16 continuation. On MBPP for DFlash Qwen3-8B, the continuation supplies $\LowCeilSplitMbppCont$ of $\tau$ while the
B24 expansion adds only $\LowCeilSplitMbppExp$. This matches the indicator, which tracks the incremental
value of a wider block instead of the domain-rebalancing continuation.
Appendix~\ref{app:per-cell-results} shows the per-cell and per-arm numbers, with the paired interval for
every comparison.

To test whether the expansion result is
specific to Qwen3, we run the identical Arm-A recipe on \GemmaTarget{}, a different model family and a
larger target. The same short expansion lifts committed length on all \GemmaBenchCountWord{} benchmarks
(median $\Delta\tau = \GemmaMedianDtau$, from $\GemmaMinDtau$ to $\GemmaMaxDtau$, every interval above
zero), and the ceiling indicator applies within the new cell (within-benchmark Spearman
$\rho = \GateArmAWithinGemmaTwelveB$, above the two-tailed $5\%$ critical value for
\NBenchWithinWord{} benchmarks). Appendix~\ref{app:per-cell-results} shows the per-cell numbers.

The Arm-B recipe also applies to \GemmaTarget{}, increasing the committed length over our from-scratch DFlare-Gemma
B16 drafter by $\GemmaArmBOverBSixteenMin$ to $\GemmaArmBOverBSixteenMax$ tokens. The within-benchmark Spearman correlation is
$\rho = \GateArmBContGemmaWithinGemmaTwelveB$. Our Arm-B B24 drafter commits
$\GemmaArmBOverDflashMin$ to $\GemmaArmBOverDflashMax$ more tokens than Z Lab's Gemma-DFlash drafter, but we also trained on more data, so the two effects are not separated. Table~\ref{tab:gemma-b24} reports both arms.
A single Arm-A expansion on \GptOssTarget{}, B8 to B12 at the same $1.5\times$ ratio, increases the committed
length on every benchmark for both original drafters (Tables~\ref{tab:gptoss-dflash-b24}
and~\ref{tab:gptoss-b24}).

The ceiling-bin mass indicates whether block expansion is worthwhile, and it also indicates when to stop. The per-cell tables in Appendix~\ref{app:per-cell-results} show the
block-ceiling fraction falling to a few percent at B24, so the B24 drafter rarely fills the wider block. With little
ceiling-bin mass left to convert, we stop at B24. The block-size study of Appendix~\ref{app:qwenladder}
reaches B32 (twice the original block size) on Qwen3-8B. We did not
train a B64 drafter. The nearly empty B32 ceiling suggests limited further benefit under this recipe,
without establishing that B64 cannot help.

\subsection{Comparing committed length with JetSpec}
\label{sec:res-jetspec}
\begin{figure}[h]
\centering
\includegraphics[width=0.82\linewidth]{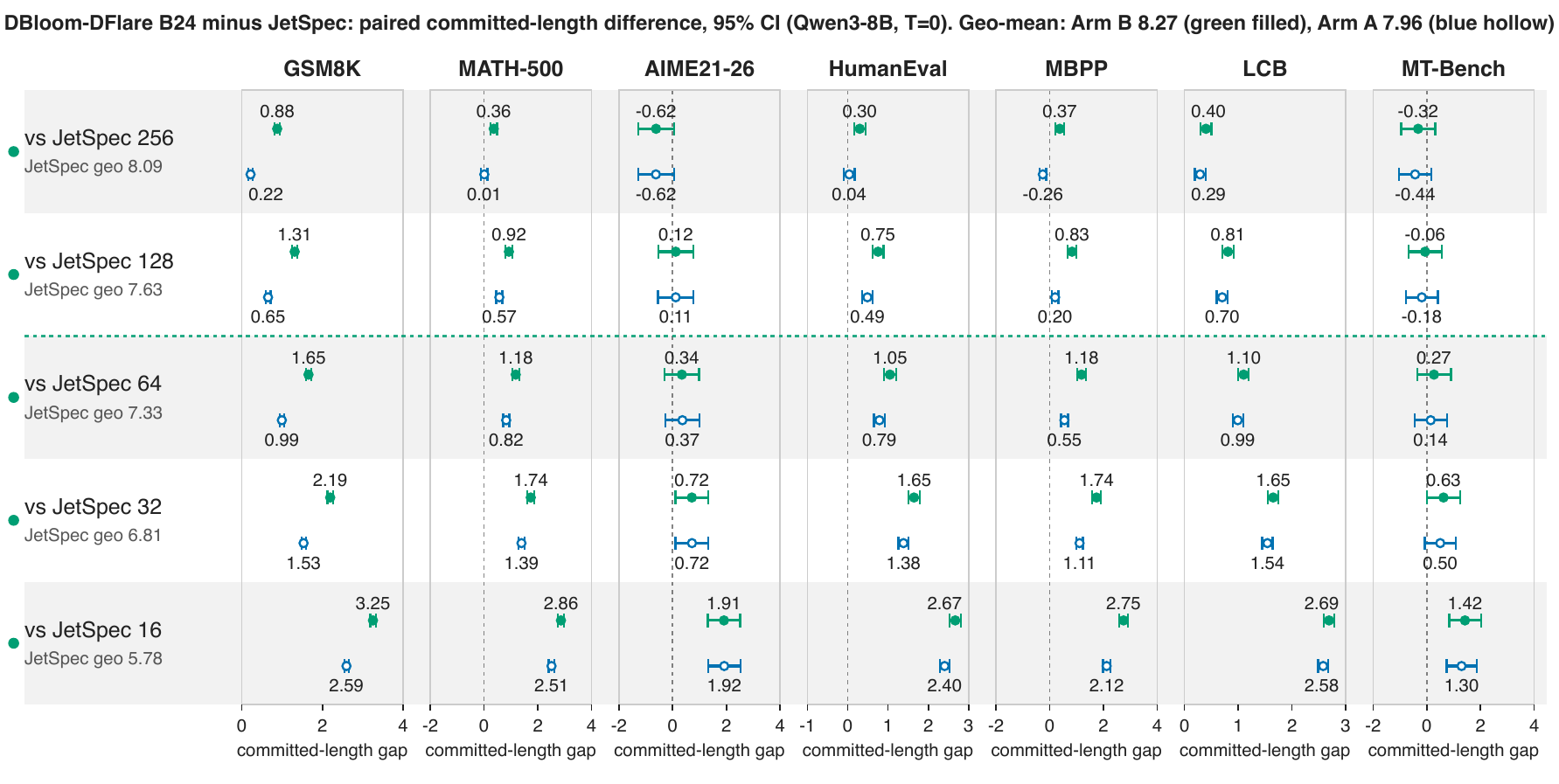}
\caption{Arm B shows no significant committed-length loss to JetSpec's tree. Arm A loses significantly on MBPP at \JetTopBudget{} nodes.
Paired difference $\Delta\tau$ (DBloom-DFlare B24 minus JetSpec) on Qwen3-8B at each tree-node budget, with
Arm B (green, filled) above Arm A (blue, hollow). Whiskers are 95\% paired-bootstrap intervals, and the dashed line
marks zero. Below the faint horizontal line, at budgets up to \JetPointBeatsAll{}~nodes, both arms exceed JetSpec's
committed length on every benchmark by point estimate. Some intervals include zero.}
\label{fig:jetspec}
\end{figure}

Against JetSpec's largest \JetTopBudget-node tree, DBloom-DFlare Arm B is ahead on
\JetTopWinsWord\ of \JetBenchCountWord\ benchmarks (positive confidence-interval lower bound)
and does not show significant loss on any. JetSpec's point
estimate leads on AIME21-26, by $0.62$ tokens ($p = 0.07$), and on MT-Bench, by $0.32$ tokens. Neither
difference is significant at the $5\%$ level. These comparisons are inconclusive, not evidence of equivalence.
Arm A has a significant committed-length loss against JetSpec on MBPP at this budget
(Appendix~\ref{app:jetspec-paired}).
We report the comparison
from the paired committed-length difference $\Delta\tau$ (ours minus JetSpec), computed per prompt on the set that
terminated under both methods, so the comparison is matched prompt for prompt (Figure~\ref{fig:jetspec}).
Because we never used JetSpec to design
DBloom, and we measure it ourselves at each tree budget under a common protocol, this is an independent
out-of-design check.

We compare committed length under one evaluation protocol. We have not completed a matched latency
or throughput comparison on a common implementation stack, so we make no claim of runtime or memory
advantage over JetSpec. That comparison remains future work.

A block size and a tree-node budget both count candidates, but they are not matched compute budgets.
The block drafter fills its block in one bidirectional pass. The near-constant cycle time measured for
B16 to B24 does not determine JetSpec's latency, which also depends on implementation and tree structure.
Appendix~\ref{app:jetspec-paired} reports the full per-budget paired $\Delta\tau$ table.

\section{Analysis and limitations}
\label{sec:analysis}

\subsection{Loss weighting has little effect in this setting}
\label{sec:lossablation}
Section~\ref{sec:curriculum} retains the two-level loss. The ablation does not establish its
superiority. On the main 30K expansion corpus, we retrain Qwen3-8B at B24 under
\spell{\LossRerunConditions} per-position weightings and evaluate all seven benchmarks, holding the drafter,
the corpus, the optimizer and the schedule fixed. The \spell{\LossRerunConditions} weightings are our recipe's
flat $3\times$ boost on the new tail positions, a uniform weight with no boost at all, three
exponential-decay profiles and two frontier profiles
(Appendix~\ref{app:lossablation30k}, Table~\ref{tab:lossrerun}). The widest spread between any two
weightings on any benchmark is $\LossRerunWidestSpread$~tokens of committed length, on
\LossRerunWidestBench{}. Of \LossRerunPairs{} paired comparisons against the main weighting,
\LossRerunPairsExcluding{} exclude zero, in both directions. The effects are small but not identically zero.

An earlier \spell{\LossAblationConditions}-profile ablation agrees. It ran on Qwen3-8B at B24 on the balanced 60K corpus of
Table~\ref{tab:corpora}, and its \spell{\LossAblationConditions} profiles fall in the same narrow band: mean $E[n]{+}1$ between \LossAblationMeanMin{} and \LossAblationMeanMax{}
over \spell{\LossAblationBenches} benchmarks (Appendix~\ref{app:lossablation}, Table~\ref{tab:lossablation}).
That ablation did not include the uniform weighting.

Acceptance stops at the first mismatch, so an early mismatch means the up-weighted tail is never
reached. We run the ablation only on Qwen3-8B.

Concurrent work reweights
along other axes: D-PACE weights draft positions dynamically by their contribution to acceptance length
\citep{dpace2026}, and \citet{whalen2026speculate} reshape the \emph{front} with a positional loss. Both
are compatible with our result, which reweights the \emph{new} tail positions.

\subsection{Scope}
\label{sec:scope}
Our controlled study is narrow by design. Its complete grid covers the Qwen3 family at 8B and 4B with
two block-diffusion drafter architectures, plus model-family checks on \GemmaTarget{} and \GptOssTarget{}.
Block expansion increases committed length on both families. The evidence for the indicator's transfer
is weaker than the Qwen3 evidence (Section~\ref{sec:res-gate}). Two extra targets do not establish universal transfer. To
check that the gains are not a lucky-seed artifact, we train the Qwen3-8B DFlare Arm-A drafter under
\SeedCount\ training seeds. The committed length agrees to within $\SeedMaxDev$ tokens across seeds, so
seed choice does not change the conclusion (Appendix~\ref{app:seed}).

Our controlled grid expands one step, from B16 to B24. Appendix~\ref{app:qwenladder} reaches twice the
original on Qwen3-8B, where B24-to-B32 differences are inconclusive in \QwenLadderLastStepInconclusive{} of
\QwenLadderLastStepCells{} comparisons. At temperature 1, Arm B improves $\tau$ significantly in all
\TOnePairs{} pairs and Arm A in \TOneArmAClear{}, with the rest inconclusive (Appendix~\ref{app:t1}).

\section{Conclusion}
\label{sec:conclusion}

A mean committed length cannot tell a draft-limited cycle from a block-limited one, whereas an acceptance
histogram can. A spike in its ceiling bin indicates that block expansion can recover acceptance.
Retraining on about 30K prompts raises the high-ceiling committed length by a median $\Delta\tau$ of
$\FamilyIncrMedianDtau$ (up to $\FamilyIncrMaxDtau$) tokens over the drafter each expansion starts from.
Across both arms, the end-to-end gain over the original B16 has median $\FamilyMedianDtau$ (up to
$\FamilyMaxDtau$) tokens, where Arm B also carries the same-block continuation before expansion. The ceiling bin flags
where a wider block pays off before we spend training compute.

\bibliographystyle{tmlr}

\appendix

\section{Training recipe}
\label{app:recipe}

This appendix provides enough detail to reproduce our expanded drafters. The block-horizon expansion is
always a fine-tune, initialized from a B16 checkpoint in contrast to a B24 trained from scratch. The expansion
recipe is identical across both architectures (DFlash and DFlare) and every target, with the only
per-cell difference being the initialization checkpoint.

\paragraph{Objective.} The drafter denoises the $B{-}1$ non-anchor positions of a block in a
single parallel pass ($K = 1$, one denoise pass per block), conditioned on the target's fused
hidden states. Training is supervised with hard-label cross-entropy against the target's
generated token IDs at each position. Labeling with the target's generations makes the
corpus self-distilled. The target produces those responses at temperature $0.6$ for the B16
continuation and greedily for the B24 expansion.

\paragraph{Two-level (flat-step) horizon-weighted loss.} When expanding from B16 to B24, the eight newly
added tail positions (block offsets $16$ through $23$) have never been trained. We weight the
per-position cross-entropy with a flat step, assigning weight $1.0$ to the old offsets and a constant $3\times$ boost to the new tail
positions, with no decay across the block. Appendix~\ref{app:lossablation} shows that this per-position
loss weighting has only small effects in the tested setting (also Appendix~\ref{app:lossablation30k}),
so we do not add further complexity.

\paragraph{Training anchors.} We train on sequences of up to $\RecipeTrainWindow$ tokens
(the target's self-distilled prompt and response, generated with a $\RecipeDistillMaxNew$-token cap).
Each training sequence contributes many blocks to the loss
via strided block-start positions, from which we sample \texttt{num\_anchors} $= 512$ per
sequence. These training anchors are only a sampling mechanism to cover many block starts within a long
sequence. They are unrelated to the inference-time block anchor of Section~\ref{sec:background}
(position 0 of a decode block, the target's committed token), which shares the word but not the
mechanism.

\paragraph{Per-step optimization.} The Arm-A expansion (and the identical Arm-B step 2) uses
a learning rate of $1\times10^{-5}$, four epochs, and a global batch size of 64. The Arm-B step 1 continuation is
a same-block (B16) fine-tune at a learning rate of $5\times10^{-5}$ for two epochs, with the STEM
and chat replay slices of Table~\ref{tab:corpora} serving as anti-forgetting regularization. All
optimization uses AdamW with gradient clipping at 1.0 and a cosine schedule. The training
corpora and their sizes are in Table~\ref{tab:corpora}.

\paragraph{Corpus-size sweep.}
\label{app:corpussweep}

The main recipe uses 30K expansion prompts. To test the size, we train Qwen3-8B drafters of both architectures in Arm A on
nested corpora of 10K, 30K and 100K prompts. We train each at the shipped 1,876 steps (four epochs on 30K prompts) and at
6,252 steps (four epochs on 100K prompts), so that corpus size and step count are separate (Table~\ref{tab:corpus-sweep}).
At 1,876 steps, the median paired $\Delta\tau$ against the original B16 drafter is $\CorpusSweepTenKMedianDtau$,
$\CorpusSweepThirtyKMedianDtau$ and $\CorpusSweepHundredKMedianDtau$ for 10K, 30K and 100K prompts. At 6,252 steps it is
$\CorpusSweepTenKLongMedianDtau$, $\CorpusSweepThirtyKLongMedianDtau$ and $\CorpusSweepHundredKLongMedianDtau$, so more steps
help the larger corpora and hurt the smallest. Each cell is one training run. Three seeds of the same recipe agree within
\SeedMaxDev{} tokens.

\begin{table}[H]
\centering
\small
\caption{Corpus-size sweep (Qwen3-8B, Arm A). Each cell is the paired change in committed length ($\Delta\tau$) against the original B16 drafter, with its 95\% interval. B16 $\tau$ is the committed length of the original drafter. (ns) means that the interval includes zero.}
\label{tab:corpus-sweep}
\resizebox{\ifdim\width>\linewidth\linewidth\else\width\fi}{!}{\begin{tabular}{lrrrrrrrr}
\toprule
\multirow{2}{*}{drafter} & \multirow{2}{*}{bench} & \multirow{2}{*}{\makecell{B16 \\ $\tau$}} & \multicolumn{3}{c}{1,876 steps} & \multicolumn{3}{c}{6,252 steps} \\
\cmidrule(lr){4-6} \cmidrule(lr){7-9}
 &  &  & 10K & 30K & 100K & 10K & 30K & 100K \\
\midrule
Qwen3-8B DFlare & GSM8K & 7.86 & +0.66 [+0.61, +0.70] & +0.71 [+0.67, +0.76] & +0.70 [+0.66, +0.75] & +0.54 [+0.49, +0.58] & +0.85 [+0.80, +0.90] & +0.89 [+0.84, +0.94] \\
Qwen3-8B DFlare & MATH-500 & 9.17 & +0.97 [+0.87, +1.07] & +1.02 [+0.92, +1.13] & +1.01 [+0.90, +1.11] & +0.86 [+0.75, +0.97] & +1.08 [+0.98, +1.19] & +1.20 [+1.09, +1.31] \\
Qwen3-8B DFlare & MBPP & 6.77 & +0.28 [+0.19, +0.37] & +0.31 [+0.22, +0.40] & +0.30 [+0.20, +0.39] & +0.07 [-0.01, +0.16] (ns) & +0.26 [+0.17, +0.35] & +0.33 [+0.24, +0.42] \\
Qwen3-8B DFlare & HumanEval & 7.07 & +0.54 [+0.43, +0.66] & +0.60 [+0.49, +0.71] & +0.57 [+0.47, +0.68] & +0.34 [+0.23, +0.45] & +0.56 [+0.45, +0.67] & +0.59 [+0.48, +0.70] \\
Qwen3-8B DFlare & LCB & 7.84 & +0.68 [+0.61, +0.75] & +0.76 [+0.68, +0.83] & +0.76 [+0.68, +0.83] & +0.50 [+0.43, +0.57] & +0.74 [+0.67, +0.82] & +0.79 [+0.72, +0.87] \\
Qwen3-8B DFlare & MT-Bench & 5.00 & +0.23 [+0.12, +0.35] & +0.26 [+0.15, +0.38] & +0.29 [+0.18, +0.41] & +0.13 [+0.02, +0.25] & +0.28 [+0.16, +0.41] & +0.26 [+0.14, +0.38] \\
Qwen3-8B DFlare & AIME21-26 & 8.63 & +0.96 [+0.68, +1.27] & +0.98 [+0.69, +1.28] & +0.98 [+0.70, +1.29] & +0.86 [+0.58, +1.17] & +0.97 [+0.69, +1.28] & +1.00 [+0.71, +1.31] \\
Qwen3-8B DFlash & GSM8K & 6.46 & +0.51 [+0.47, +0.54] & +0.55 [+0.51, +0.58] & +0.57 [+0.53, +0.60] & +0.53 [+0.50, +0.57] & +0.78 [+0.74, +0.82] & +0.83 [+0.79, +0.87] \\
Qwen3-8B DFlash & MATH-500 & 8.03 & +0.97 [+0.88, +1.06] & +1.01 [+0.92, +1.09] & +0.97 [+0.88, +1.06] & +0.94 [+0.84, +1.03] & +1.20 [+1.10, +1.29] & +1.23 [+1.13, +1.32] \\
Qwen3-8B DFlash & MBPP & 5.91 & +0.07 [-0.00, +0.14] (ns) & +0.10 [+0.02, +0.17] & +0.11 [+0.03, +0.18] & -0.03 [-0.11, +0.04] (ns) & +0.10 [+0.02, +0.17] & +0.13 [+0.05, +0.20] \\
Qwen3-8B DFlash & HumanEval & 6.49 & +0.39 [+0.29, +0.49] & +0.43 [+0.32, +0.54] & +0.43 [+0.32, +0.54] & +0.28 [+0.18, +0.38] & +0.42 [+0.32, +0.53] & +0.45 [+0.34, +0.56] \\
Qwen3-8B DFlash & LCB & 6.96 & +0.34 [+0.28, +0.40] & +0.38 [+0.32, +0.44] & +0.39 [+0.33, +0.45] & +0.19 [+0.13, +0.24] & +0.38 [+0.32, +0.44] & +0.45 [+0.39, +0.51] \\
Qwen3-8B DFlash & MT-Bench & 4.22 & +0.19 [+0.08, +0.31] & +0.21 [+0.10, +0.32] & +0.22 [+0.12, +0.34] & +0.15 [+0.04, +0.26] & +0.23 [+0.11, +0.36] & +0.23 [+0.12, +0.36] \\
Qwen3-8B DFlash & AIME21-26 & 7.68 & +0.87 [+0.63, +1.11] & +0.88 [+0.65, +1.12] & +0.90 [+0.67, +1.15] & +0.86 [+0.62, +1.11] & +0.97 [+0.73, +1.22] & +0.99 [+0.76, +1.24] \\
\bottomrule
\end{tabular}
}
\end{table}
 
\section{Loss-weighting ablation, per benchmark}
\label{app:lossablation}

Table~\ref{tab:lossablation} presents the full six-arm loss-weighting ablation
for Qwen3-8B at B24 with the balanced 60K corpus of Table~\ref{tab:corpora}, thinking mode disabled.
That corpus predates the choice in Section~\ref{sec:setup} to keep only high-ceiling domains. In particular, it
includes a chat slice and short-form code, which the 30K expansion corpus drops because those
benchmarks have little ceiling-bin mass to convert. We
rerun the weighting comparison on the expansion corpus below rather than assume it transfers.
The metric is $E[n]{+}1$, the number of tokens committed per target forward pass, averaged over
cycles per benchmark. The last two columns report the mean across the five benchmarks and the measured
speed-up. Each arm changes
only the per-position weighting, holding the corpus and all other knobs fixed. All six arms fall in
the same narrow band on every benchmark, so the weighting does not change committed length at this
corpus scale.

\begin{table}[H]
\centering
\small
\caption{Loss-weighting ablation on the 60K-prompt corpus (Qwen3-8B, B24). Cells are $E[n]{+}1$ per benchmark. The last two columns are the mean $E[n]{+}1$ and the mean measured speed-up over the five benchmarks. The first row is our recipe, the flat $3\times$ boost. Table~\ref{tab:lossrerun} repeats the ablation on the 30K expansion corpus.}
\label{tab:lossablation}
\resizebox{\ifdim\width>\linewidth\linewidth\else\width\fi}{!}{\begin{tabular}{lrrrrrrr}
\toprule
\makecell{Per-position \\ weighting} & GSM8K & MATH-500 & HumanEval & MBPP & MT-Bench & mean & \makecell{mean \\ sp} \\
\midrule
Flat 3$\times$ boost (ours) & 7.92 & 9.01 & 6.92 & 6.20 & 3.89 & 6.79 & 4.91$\times$ \\
Exp. decay, unit peak & 7.93 & 8.90 & 6.91 & 6.21 & 3.82 & 6.75 & 4.88$\times$ \\
Exp. decay + 3$\times$ tail boost & 7.95 & 8.93 & 6.89 & 6.18 & 3.84 & 6.76 & 4.89$\times$ \\
Flatter exp. decay & 7.93 & 8.93 & 6.90 & 6.20 & 3.82 & 6.76 & 4.89$\times$ \\
Frontier (sensitivity) & 7.93 & 8.97 & 6.90 & 6.20 & 3.85 & 6.77 & 4.90$\times$ \\
Frontier (mid-block bump) & 7.92 & 8.98 & 6.89 & 6.22 & 3.85 & 6.77 & 4.90$\times$ \\
\bottomrule
\end{tabular}
}
\end{table}
 
\paragraph{Rerun on the expansion corpus.}
\label{app:lossablation30k}

The 60K corpus of Table~\ref{tab:lossablation} is not the 30K expansion corpus of the main recipe, so we retrain the
weightings on the 30K corpus (Table~\ref{tab:lossrerun}), with the drafter, optimizer and schedule fixed. The rerun adds
\emph{Flat $1\times$ (no boost)}, which differs from \emph{Flat $3\times$ boost (ours)}, the main recipe, only in the
multiplier on the new tail positions. On every benchmark the \spell{\LossRerunConditions} weightings are within
$\LossRerunWidestSpread$ tokens of each other. In the lower panel, \spell{\LossRerunPairsExcluding} of the
\LossRerunPairs{} paired differences from Flat $3\times$ boost (ours) exclude zero, the largest is $\LossRerunPairsLargest$
tokens, and they point in both directions. Removing the boost altogether changes no benchmark by more than
$\LossRerunUniformMaxDiff$ tokens, and none of its \spell{\LossRerunUniformBenches} differences excludes zero. So no
weighting is clearly better, and we keep Flat $3\times$ boost (ours) because every drafter in this paper uses it. The rerun
also changes the code data, the learning rate, the position where the tail boost starts
(Section~\ref{sec:curriculum}) and the training path, but all \spell{\LossRerunConditions} conditions share them.

\begin{figure}[H]
\centering
\includegraphics[width=\linewidth]{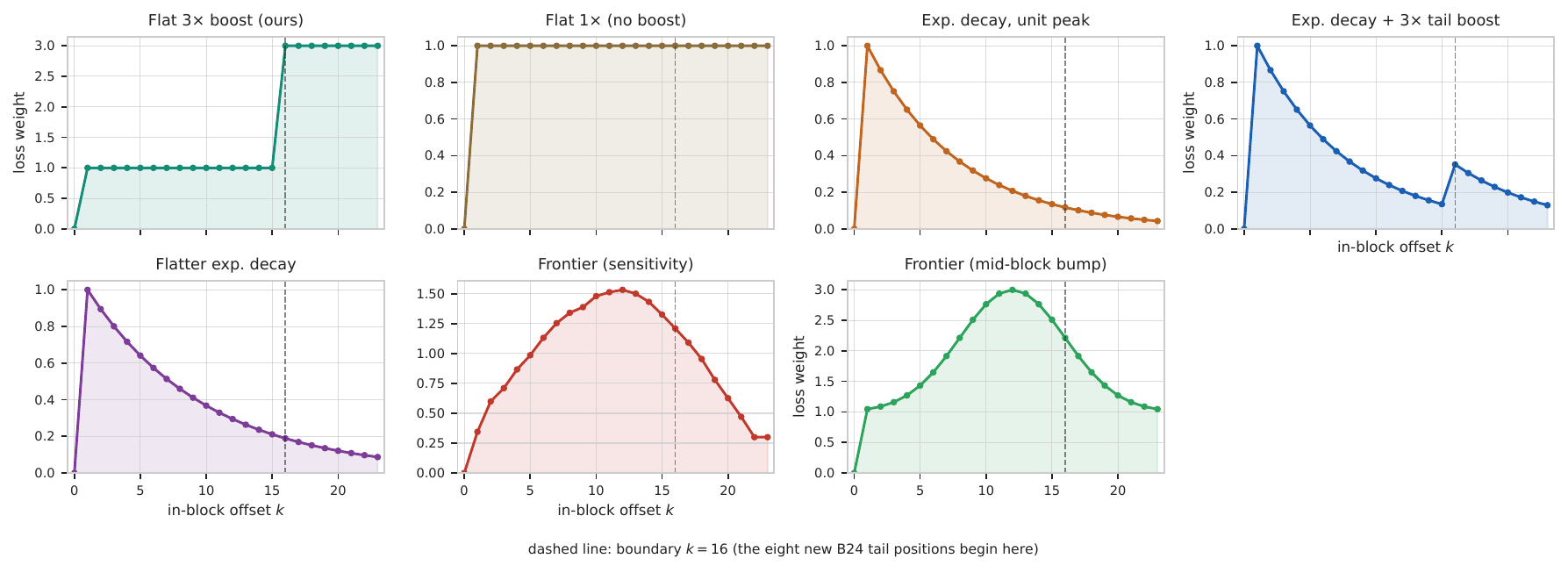}
\caption{Weight versus in-block offset $k$ for the \spell{\LossRerunConditions} loss-weighting profiles of
Tables~\ref{tab:lossablation} and~\ref{tab:lossrerun}.}
\label{fig:lossprofiles}
\end{figure}

Figure~\ref{fig:lossprofiles} shows the \spell{\LossRerunConditions} weight profiles. The x-axis is the in-block offset $k$
(offset $0$ is the anchor, which carries no loss), and the dashed line marks the boundary $k=16$ where
the eight new B24 tail positions begin. The arms differ only in how they shape the weight across the block.

\begin{table}[H]
\centering
\small
\caption{Loss-weighting rerun on the 30K expansion corpus. Top: the prompt-mean committed length $\tau$ over EOS-terminated responses. Bottom: the paired change in $\tau$ from Flat 3$\times$ boost (ours).}
\label{tab:lossrerun}
\resizebox{\ifdim\width>\linewidth\linewidth\else\width\fi}{!}{\begin{tabular}{lrrrrrrrr}
\toprule
\makecell{Per-position \\ weighting} & GSM8K & MATH-500 & MBPP & HumanEval & LCB & MT-Bench & AIME21-26 & mean \\
\midrule
Flat 3$\times$ boost (ours) & 8.57 & 10.18 & 7.08 & 7.65 & 8.55 & 5.29 & 9.40 & 8.10 \\
Flat 1$\times$ (no boost) & 8.58 & 10.16 & 7.08 & 7.64 & 8.57 & 5.29 & 9.40 & 8.10 \\
Exp. decay, unit peak & 8.59 & 10.13 & 7.08 & 7.64 & 8.58 & 5.29 & 9.39 & 8.10 \\
Exp. decay + 3$\times$ tail boost & 8.60 & 10.19 & 7.07 & 7.64 & 8.58 & 5.29 & 9.40 & 8.11 \\
Flatter exp. decay & 8.58 & 10.14 & 7.08 & 7.62 & 8.58 & 5.29 & 9.38 & 8.10 \\
Frontier (sensitivity) & 8.57 & 10.16 & 7.06 & 7.62 & 8.57 & 5.28 & 9.41 & 8.10 \\
Frontier (mid-block bump) & 8.57 & 10.16 & 7.07 & 7.62 & 8.56 & 5.27 & 9.40 & 8.10 \\
\midrule
\multicolumn{9}{l}{\itshape Difference from Flat 3$\times$ boost (ours). A star marks a 95\% paired interval that excludes zero.}\\
\midrule
Flat 1$\times$ (no boost) & +0.009 & -0.020 & -0.005 & -0.012 & +0.019 & -0.003 & +0.006 & -0.001 \\
Exp. decay, unit peak & +0.019 & -0.052* & -0.005 & -0.004 & +0.033* & -0.000 & -0.005 & -0.002 \\
Exp. decay + 3$\times$ tail boost & +0.028* & +0.003 & -0.006 & -0.011 & +0.037* & +0.001 & +0.009 & +0.009 \\
Flatter exp. decay & +0.017 & -0.040* & +0.004 & -0.028 & +0.034* & -0.003 & -0.013 & -0.004 \\
Frontier (sensitivity) & +0.004 & -0.022 & -0.017 & -0.029 & +0.021 & -0.012 & +0.011 & -0.007 \\
Frontier (mid-block bump) & +0.005 & -0.024 & -0.007 & -0.024 & +0.015 & -0.017 & +0.009 & -0.006 \\
\bottomrule
\end{tabular}
}
\end{table}
 
\section{Data-versus-expansion control}
\label{app:control}

To confirm that the Arm-A gain comes from the block expansion rather than from post-training on 30K more
prompts, we run a control that post-trains the original B16 DFlare-Qwen3-8B drafter on the same 30K
corpus but keeps the block size at 16, with no expansion. Table~\ref{tab:control} reports each paired change with a 95\% interval over the shared EOS-terminated prompts. The
data-only effect is positive but smaller than the expansion on every benchmark, so the extra data alone do not account for
the improvement.

\begin{table}[H]
\centering
\small
\caption{Data-versus-expansion control (Qwen3-8B, DFlare). Committed length $\tau$ of the original B16, the control and the Arm-A B24 expansion, and the paired change of the control and of the expansion from the original B16.}
\label{tab:control}
\resizebox{\ifdim\width>\linewidth\linewidth\else\width\fi}{!}{\begin{tabular}{lrrrrr}
\toprule
bench & \makecell{B16 \\ $\tau$} & \makecell{B16+30K \\ $\tau$} & \makecell{Arm-A \\ B24 \\ $\tau$} & \makecell{$\Delta\tau$ \\ data-only} & \makecell{$\Delta\tau$ \\ expansion} \\
\midrule
GSM8K & 7.86 & 8.07 & 8.57 & +0.21 [+0.18, +0.24] & +0.71 [+0.67, +0.76] \\
MATH-500 & 9.17 & 9.29 & 10.19 & +0.12 [+0.08, +0.16] & +1.02 [+0.92, +1.13] \\
AIME21-26 & 8.63 & 8.68 & 9.60 & +0.05 [+0.01, +0.08] & +0.98 [+0.69, +1.28] \\
HumanEval & 7.07 & 7.16 & 7.66 & +0.10 [+0.05, +0.15] & +0.60 [+0.49, +0.71] \\
MBPP & 6.77 & 6.86 & 7.08 & +0.08 [+0.03, +0.14] & +0.31 [+0.22, +0.40] \\
LCB & 7.84 & 8.01 & 8.60 & +0.17 [+0.14, +0.21] & +0.76 [+0.68, +0.83] \\
MT-Bench & 5.00 & 5.08 & 5.26 & +0.08 [+0.03, +0.13] & +0.26 [+0.15, +0.38] \\
\bottomrule
\end{tabular}
}
\end{table}
 
\section{Seed robustness}
\label{app:seed}

We train the Qwen3-8B DFlare Arm-A B24 drafter from the DBloom recipe under \SeedCount\ different
training seeds (feeding the seed to weight initialization, dropout, and data order) and re-evaluate all
seven benchmarks. Table~\ref{tab:seed} reports it. The \SeedCount\ seeds agree to within a few hundredths of a
token on every benchmark, far inside the evaluation confidence interval, so the acceptance gains are
stable to the training seed. This is a Qwen3-8B DFlare spot-check.

\begin{table}[H]
\centering
\small
\caption{Seed robustness of the Arm-A expansion (Qwen3-8B, DFlare). Committed length $\tau$ with a 95\% interval under three training seeds.}
\label{tab:seed}
\begin{tabular}{lrrrr}
\toprule
bench & \makecell{seed \\ 42 \\ $\tau$} & \makecell{seed \\ 123 \\ $\tau$} & \makecell{seed \\ 789 \\ $\tau$} & \makecell{max \\ |$\Delta\tau$|} \\
\midrule
GSM8K & 8.57 $\pm$ 0.09 & 8.56 $\pm$ 0.09 & 8.56 $\pm$ 0.09 & 0.016 \\
MATH-500 & 10.18 $\pm$ 0.22 & 10.18 $\pm$ 0.22 & 10.17 $\pm$ 0.21 & 0.008 \\
AIME21-26 & 9.42 $\pm$ 0.41 & 9.39 $\pm$ 0.41 & 9.39 $\pm$ 0.41 & 0.029 \\
HumanEval & 7.66 $\pm$ 0.20 & 7.67 $\pm$ 0.20 & 7.65 $\pm$ 0.20 & 0.016 \\
MBPP & 7.08 $\pm$ 0.20 & 7.07 $\pm$ 0.19 & 7.08 $\pm$ 0.19 & 0.011 \\
LCB & 8.55 $\pm$ 0.15 & 8.55 $\pm$ 0.15 & 8.54 $\pm$ 0.15 & 0.011 \\
MT-Bench & 5.27 $\pm$ 0.45 & 5.28 $\pm$ 0.46 & 5.27 $\pm$ 0.45 & 0.012 \\
\bottomrule
\end{tabular}
\end{table}
 
\section{Why \texorpdfstring{$\tau$}{tau} and \texorpdfstring{$E[n]{+}1$}{E[n]+1} are committed-length statistics, how they relate to acceptance length, and when they differ}
\label{app:tau}

In each decode cycle, the target verifies a proposed block, accepts a prefix of $n$ draft
tokens ($0 \le n \le B{-}1$), then commits one guaranteed bonus token, producing
$n{+}1$ tokens. The two summaries of a run in Section~\ref{sec:metrics} both include that bonus
token but weight cycles differently, since $\tau$ weights each prompt equally while $E[n]{+}1$ weights
each cycle equally. Calling either one ``the acceptance length'' obscures both the bonus token and
the choice of weighting, so we name each by what it counts, the tokens committed per cycle, with $\tau$
prompt-weighted and $E[n]{+}1$ cycle-weighted. Both are committed lengths.

\paragraph{The identity.} With $c_p$ the cycle count of prompt $p$ and $\bar{n}_p$ its
per-prompt mean accepted count, $E[n]{+}1$ is a cycle-count-weighted average of the same
per-prompt means that $\tau$ averages evenly. The gap is the weighted-minus-unweighted-mean
identity,
\begin{equation}
E[n]{+}1 - \tau = \frac{\mathrm{Cov}(c_p,\ \bar{n}_p)}{\overline{c}},
\end{equation}
where $\overline{c} = \frac{1}{P}\sum_p c_p$ is the mean cycle count over the $P$ prompts and
$\mathrm{Cov}(c_p, \bar{n}_p)$ is the covariance between a prompt's cycle count and its per-prompt mean
accepted count. The denominator is positive, so the sign depends entirely on whether a prompt's cycle
count is correlated with its acceptance. When long generations draft poorly (negative covariance), the many low-acceptance
cycles dominate the cycle-weighted mean and pull $E[n]{+}1$ below $\tau$. When long generations draft well (positive
covariance), $E[n]{+}1$ rises above $\tau$.
The two summaries are equal only when generation length and acceptance are uncorrelated.

\paragraph{A minimal illustration.} Two prompts with ten cycles between them make the mechanism concrete. In Table~\ref{tab:toy}, both cases share the same per-prompt behavior, so $\tau = \ToyTau$ in each, yet the throughput $E[n]{+}1$ ranges from $\ToyEnpLow$ to $\ToyEnpHigh$ depending on whether the many-cycle prompt is the one the drafter handles well.

\begin{table}[h]
\centering
\small
\caption{Identical $\tau$, opposite $E[n]{+}1$. Each prompt contributes $\bar{n}_p{+}1$. $\tau$ is the
unweighted mean of those contributions and $E[n]{+}1$ the cycle-weighted mean. Both cases share the same
per-prompt behavior, so $\tau = \ToyTau$ in each, but $E[n]{+}1$ follows the many-cycle prompt: below
$\tau$ when it drafts poorly (Case 1) and above $\tau$ when it drafts well (Case 2).}
\label{tab:toy}
\begin{tabular}{llrl r rr}
\toprule
Case & $p$ & Cycles & Accepted $n$ each cycle & $\bar{n}_p{+}1$ & $\tau$ & $E[n]{+}1$ \\
\midrule
\multirow{2}{*}{1}
 & A (short, drafts well)   & 2 & 7, 7            & 8 & \multirow{2}{*}{$\ToyTau$} &   \multirow{2}{*}{$\ToyEnpLow$} \\
 & B (long, drafts poorly)  & 8 & 1 (eight times) & 2 & \\
\midrule
\multirow{2}{*}{2}
 & A (short, drafts poorly) & 2 & 1, 1            & 2 & \multirow{2}{*}{$\ToyTau$} &  \multirow{2}{*}{$\ToyEnpHigh$} \\
 & B (long, drafts well)    & 8 & 7 (eight times) & 8 & \\
\bottomrule
\end{tabular}
\end{table}

\paragraph{Real data, both directions.} Both cases occur across our runs (greedy $T=0$, full test
sets, B24, EOS-terminated responses only), though the divergence is modest once the runaway generations
are filtered out. The negative direction is larger. On MATH-500 with the DBloom-DFlash Qwen3-4B
Arm-B B24 drafter, the long, low-acceptance chains dominate the cycle-weighted mean, so $E[n]{+}1 =
\TauEnpDflashFourBMathEnp$ sits $\TauEnpDflashFourBMathDiffMag$ tokens below $\tau =
\TauEnpDflashFourBMathTau$, and reporting only $\tau$ would overstate throughput. The positive direction
appears on AIME21-26 with the DBloom-DFlare Qwen3-8B Arm-A B24 drafter, where the longer generations
draft slightly better than the short ones, so $E[n]{+}1 = \TauEnpDflareEightBAimeEnp$ rises
$\TauEnpDflareEightBAimeDiffMag$ tokens above $\tau = \TauEnpDflareEightBAimeTau$ and reporting only
$\tau$ would understate it. The gap is a few tenths of a token in either direction here, but it can
grow when a workload's generation length and acceptance are strongly correlated. The two summaries answer
different questions. $\tau$ is the right per-prompt statistic for a confidence interval and
$E[n]{+}1$ is the right throughput statistic, so we report both summaries and keep them distinct.

\paragraph{Committed length tracks the measured speed-up.} That $E[n]{+}1$ is the throughput
statistic is more than definitional. Across every published target-drafter pair, it tracks the measured
decode-only speed-up closely (Table~\ref{tab:speedup-corr}). Within each target-drafter pair,
we fit one line to its two B24 arms across benchmarks, yielding one point per benchmark and arm. This is an empirical
association. A fixed model and block size do not guarantee equal cycle time across workloads.
We measure every run on the prompts that end with an end-of-sequence token across the stages of its
target-drafter group, the same
prompts as the per-cell tables, since the greedy runaways that never emit a stop token inflate length without a
matching throughput increase. A run with no such group uses its own EOS-terminated responses. $E[n]{+}1$ reaches Pearson $r \ge \SpeedupCorrMinR{}$
in every pair, with a bootstrap lower bound above that of $\tau$ throughout, confirming the
cycle-weighted summary as the speed-up proxy. Within a fixed target-drafter implementation,
cycle-mean committed length closely tracks the measured decode-only speed-up in these runs.
As a token count it stays comparable across stacks, though realized
latency depends on the implementation and hardware.

\begin{table}[t]
\centering
\small
\caption{Committed length against measured decode-only speed-up, one row per target-drafter pair. The correlation is over its two arms and all benchmarks ($m$ points, one per run). Intervals are 95\% paired percentile bootstrap intervals.}
\label{tab:speedup-corr}
\begin{tabular}{llr rr rr}
\toprule
Target & Drafter & $m$ & $r(E[n]{+}1)$ & 95\% CI & $r(\tau)$ & 95\% CI \\
\midrule
Qwen3-4B & DFlare & 14 & $+0.988$ & $[+0.97, +1.00]$ & $+0.963$ & $[+0.90, +0.99]$ \\
Qwen3-4B & DFlash & 14 & $+0.988$ & $[+0.97, +1.00]$ & $+0.969$ & $[+0.91, +0.99]$ \\
Qwen3-8B & DFlare & 14 & $+0.997$ & $[+0.99, +1.00]$ & $+0.969$ & $[+0.90, +0.99]$ \\
Qwen3-8B & DFlash & 14 & $+0.999$ & $[+1.00, +1.00]$ & $+0.984$ & $[+0.96, +1.00]$ \\
Gemma-4-12B-IT & DFlare & 14 & $+0.978$ & $[+0.92, +1.00]$ & $+0.889$ & $[+0.66, +0.97]$ \\
\bottomrule
\end{tabular}
\end{table}
 
\section{Why naive block expansion lowers per-position acceptance, and by how much the accepted length falls}
\label{app:blocksize}

This appendix analyzes the claim in Section~\ref{sec:motivation} in two steps. First, running a
drafter trained at block size $B_{\text{tr}}$ at a larger block $B$ changes the proposal distribution
at every position, including the early ones. We explain why this shift happens, and show that at temperature zero
it can preserve or break a match the trained block already made but never create one,
so it cannot raise an already-accepted position.
We do not claim the shift always increases the distance to the target at every position under sampled
decoding, and we treat the empirical fact that per-position acceptance falls in every measured cell as
just that, an empirical finding. Second, the accepted length sums acceptance over positions and stops
at the first rejection, so a drop in per-position rates need not lower it by the same amount. The extra
positions a wider block reaches can still be accepted, which recovers some of the length lost at the
front. Together, these explain why some cells lose almost no accepted length even though per-position
acceptance falls across the block.

\paragraph{Acceptance as a distance.} Write $q_i^{(B)}$ for the drafter's distribution at
masked position $i$ of a $B$-wide block, and $p_i$ for the target's true conditional at that
position. The target does not depend on $B$. Under speculative sampling, the verifier accepts a
proposed token $x_i \sim q_i^{(B)}$ with probability $\min(1, p_i(x_i)/q_i^{(B)}(x_i))$, so the
per-position acceptance rate, \emph{conditional} on the first $i-1$ positions having been accepted
(the state in which position $i$ is verified), is
\begin{equation}
\alpha_i^{(B)} = \sum_x \min\!\big(p_i(x),\, q_i^{(B)}(x)\big) = 1 - \mathrm{TV}\!\big(p_i,\, q_i^{(B)}\big),
\label{eq:accept-tv}
\end{equation}
which equals one minus the total variation distance between the drafter's proposal and the target's, where
$\mathrm{TV}(p, q) = \tfrac{1}{2}\sum_x |p(x) - q(x)|$ is the largest gap in probability the two
distributions assign to any event. Greedy
decoding is the temperature-zero special case, where both sides concentrate on their argmax and
$\alpha_i^{(B)} \in \{0,1\}$ recovers exact-match acceptance. The verifier stops at the first
rejection, so with $\alpha_i^{(B)}$ read as this conditional rate the accepted length satisfies
$\Pr[n \ge k] = \prod_{i=1}^{k} \alpha_i^{(B)}$ and
$E[n] = \sum_{k=1}^{B-1} \prod_{i=1}^{k} \alpha_i^{(B)}$.

\paragraph{The proposal depends on the block size.} A block-diffusion drafter fills its block with
bidirectional attention, so every masked position, including an early one, attends to the whole block.
Widening the block adds positions that even an early query now attends to, so its attention mixture, and
with it the token it proposes, moves away from what the drafter learned at its trained size. The new positions
add a second dependence, because the drafter now sees offsets and a position range it never trained on.
Neither effect depends on the head-sharing scheme (MHA, GQA, MQA, MLA) or the positional design (rotary,
learned, ALiBi). Both follow from attention being bidirectional over the block. An autoregressive drafter
behaves differently. It computes each position from only the causal prefix, so asking for more positions
leaves its earlier proposals untouched, and extending the horizon cannot lower their per-position
acceptance. A block-diffusion drafter has no matching guarantee, so widening the block can move its
proposal at every position, the early ones included. We do not claim the shift always moves the proposal
farther from the target, either at a single position or once the residual stream and later layers
recombine the heads. We treat the resulting drop in per-position acceptance as an empirical fact, present
in every measured cell.

\begin{figure}[tbp]
\centering
\includegraphics[width=0.82\linewidth]{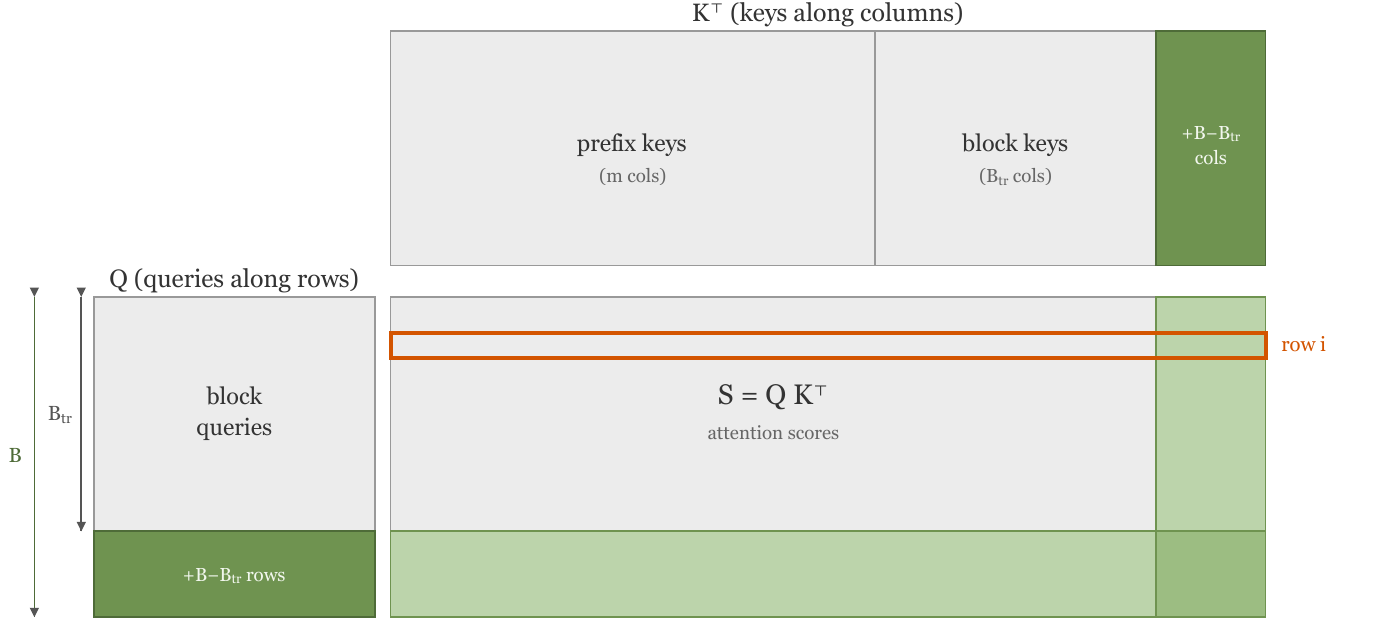}
\caption{Attention scores for one block. Expanding the block from the trained size $B_{\text{tr}}$ to a
larger $B$ appends key columns and query rows (green). Even an early query row $i \ll B_{\text{tr}}$
(orange) gains new score entries, so its attention weights, and thus the token it proposes, shift away
from what the drafter learned at its trained size. This drift can reduce per-position acceptance, and it
does so across the evaluated cells.}
\label{fig:gemm}
\end{figure}

\paragraph{Per-position consequence of greedy decoding.} Greedy decoding reduces each acceptance decision
to an exact match. At a position the trained block already accepted, the distribution shift can preserve or
break that match but cannot improve it further. At a position the trained block rejected, the shift can
leave the rejection or correct it. Neither direction is guaranteed at every position, so no consistent
pattern applies. Empirically, the effect is one-directional. The per-position match rate on the old
positions falls in every measured cell, which drives the front erosion analyzed below. Under sampled
decoding, the same argument does not apply, yet the ceiling columns of Table~\ref{tab:naive-massshift}
show the block-ceiling fraction falling to near zero in every measured cell regardless.

\begin{figure}[tbp]
\centering
\includegraphics[width=\linewidth]{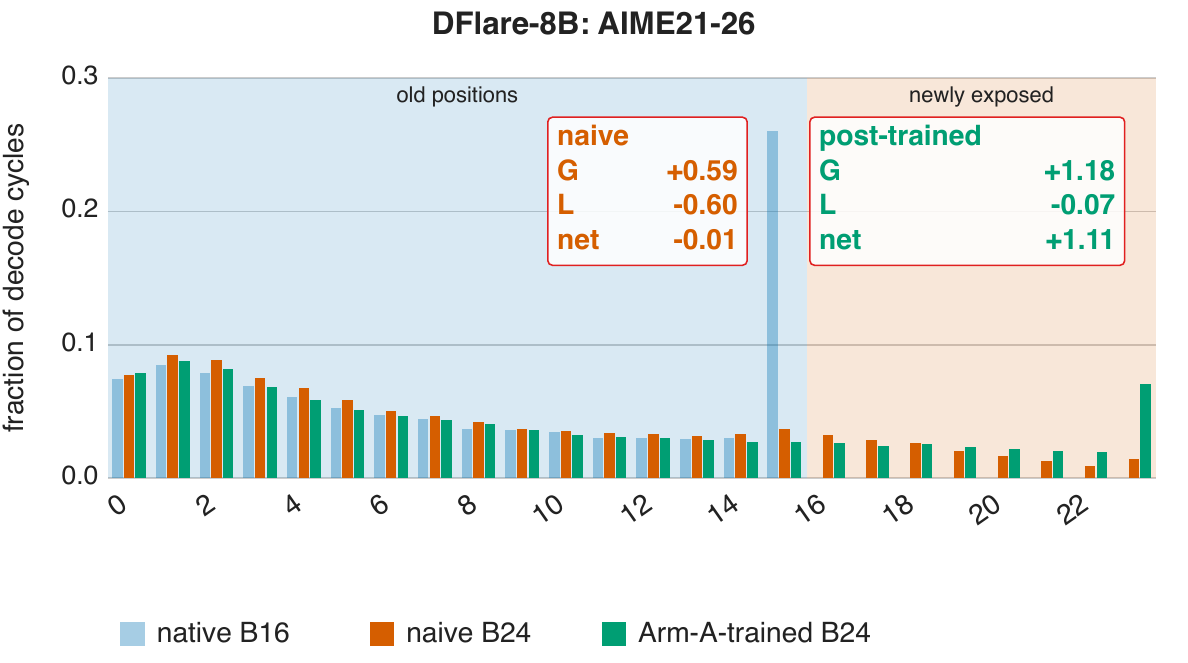}\\[6pt]
\includegraphics[width=\linewidth]{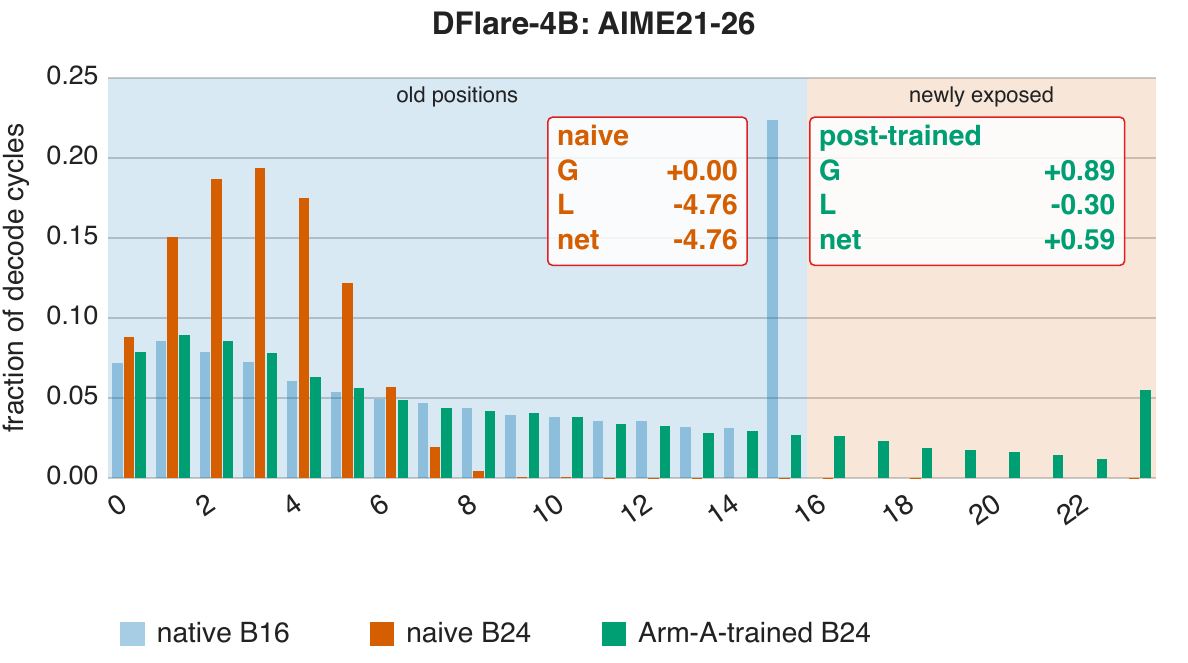}
\caption{Per-cycle accepted-length histograms for DFlare on AIME21-26 under naive expansion, on the
shared $0$ to $23$ axis: native B16 (dimmed), naive B24 (original drafter run at B24, no training), and
Arm-A-trained B24. Top Qwen3-8B, bottom Qwen3-4B. Shaded bands mark the old positions $1$--$15$ and the
new positions $16$--$23$. The $G/L/\text{net}$ readouts give the Eq.~\ref{eq:massshift}
decomposition for the naive pair and the post-trained pair. Table~\ref{tab:naive-massshift} quantifies every cell.}
\label{fig:naivehist}
\end{figure}

\paragraph{The accepted length is a censored sum.} We now turn to the second part. The expected accepted length does not inherit the per-position monotonicity.
From Eq.~\ref{eq:accept-tv}, $E_B[n] = \sum_{k=1}^{B-1} \Pr_B[n \ge k]$ is a
\emph{sum whose number of terms grows with $B$}. Enlarging the block appends new nonnegative
terms even as each surviving term shrinks. Writing $P_B(k) = \Pr_B[n \ge k]$, the change from the
trained block to the larger one splits into a gain and a loss,
\begin{equation}
E_B[n] - E_{B_{\text{tr}}}[n]
= \underbrace{\sum_{k=B_{\text{tr}}}^{B-1} P_B(k)}_{G:\ \text{uncapped gain}}
- \underbrace{\sum_{k=1}^{B_{\text{tr}}-1} \big[P_{B_{\text{tr}}}(k) - P_B(k)\big]}_{L:\ \text{front erosion}},
\label{eq:massshift}
\end{equation}
where $G$ is the survival gained at the newly reachable positions and $L$ is the survival lost
on the old positions, nonnegative whenever old-position survival does not increase ($P_B(k) \le
P_{B_{\text{tr}}}(k)$ for $k < B_{\text{tr}}$).
The accepted length rises if and only if $G > L$. An upper bound on the uncapped
gain is the amount of the block that survives to the new positions,
\begin{equation}
G \le (B - B_{\text{tr}})\, P_B(B_{\text{tr}}),
\label{eq:gbound}
\end{equation}
because $P_B(k)$ is nonincreasing in $k$. This bound uses only that $P_B$ is a survival function, so it holds true
for any block-diffusion drafter.
Table~\ref{tab:naive-massshift}
reports $G$, $L$, and the survival $P_B(B_{\text{tr}})$ per cell.
Over EOS-terminated responses, naive
expansion raises the accepted length on no cell. Every $G-L$ is at most zero. The size of the fall tracks
how much of the block survives to the new positions. On the 8B drafters, a modest ceiling survives past the
boundary ($P_B(B_{\text{tr}})$ around 9 to 16\%), so $G$ nearly offsets $L$ and the accepted length stays
about flat, dipping slightly (DFlare Qwen3-8B AIME21-26 $G-L=\MassShiftNetDflareEightBAime$,
LiveCodeBench $\MassShiftNetDflareEightBLcb$, MATH-500 $\MassShiftNetDflareEightBMath$, and DFlash Qwen3-8B
AIME21-26 $\MassShiftNetDflashEightBAime$). On the 4B drafters, $P_B(B_{\text{tr}}) \approx 0$, so
$G \approx 0$ and the accepted length falls by essentially the full front erosion.
Figure~\ref{fig:naivehist} makes $G$ and $L$ concrete on two AIME21-26 cells: a near-flat 8B case
where the surviving ceiling almost balances the front loss, and a 4B case where almost no mass clears
the boundary so the distribution shifts toward the low bins.

\paragraph{When a rise happens (and why it is unlikely).} A simple model makes the condition explicit.
Suppose survival at the first newly reachable position is
$S = P_B(B_{\text{tr}})$ and decays by a factor $\beta$ per additional position,
so $P_B(B_{\text{tr}}+j) = S\,\beta^{\,j}$. Then the uncapped gain is a geometric sum,
\begin{equation}
G = S\, \frac{1 - \beta^{\,B - B_{\text{tr}}}}{1 - \beta},
\label{eq:gmodel}
\end{equation}
and $E_B[n] > E_{B_{\text{tr}}}[n]$ iff $G > L$. This model assumes the new positions are homogeneous and independent, so it remains a heuristic. It nonetheless pinpoints the requirement, since a rise needs both a large surviving ceiling $S$ and a capable drafter $\beta$. That combination is uncommon because the same block-size shift that exposes the new positions also erodes the front that feeds $S$, and when $S \approx 0$, as on every 4B cell, Eq.~\ref{eq:gbound} forces a fall. A speed-up gain is rarer still, since even a flat accepted length does not offset the wider block's higher per-cycle cost.

\begin{table}[tbp]
\centering
\small
\caption{Mass shift under naive expansion. The change in E[n] splits into the uncapped gain G and the front erosion L, with the ceiling fraction and the low-acceptance mass before and after.}
\label{tab:naive-massshift}
\resizebox{\ifdim\width>\linewidth\linewidth\else\width\fi}{!}{\begin{tabular}{llrrrrrrrrrr}
\toprule
model & size & bench & \makecell{native \\ E[n]} & \makecell{naive \\ E[n]} & \makecell{G \\ uncapped} & \makecell{L \\ front} & $\Pr(n \ge 16)$ & \makecell{ceiling \\ native} & \makecell{ceiling \\ naive} & \makecell{$\Pr(n \le 4)$ \\ native} & \makecell{$\Pr(n \le 4)$ \\ naive} \\
\midrule
\multirow{14}{*}{DFlare} & \multirow{7}{*}{8B} & GSM8K & 6.70 & 6.34 & 0.30 & 0.66 & 9\% & 16\% & 1\% & 44\% & 48\% \\
 &  & MATH-500 & 8.06 & 7.65 & 0.50 & 0.91 & 14\% & 26\% & 1\% & 35\% & 40\% \\
 &  & HumanEval & 5.96 & 5.87 & 0.28 & 0.38 & 8\% & 14\% & 1\% & 52\% & 53\% \\
 &  & MBPP & 5.38 & 5.07 & 0.13 & 0.44 & 4\% & 9\% & 0\% & 54\% & 57\% \\
 &  & MT-Bench & 3.03 & 2.88 & 0.05 & 0.20 & 1\% & 3\% & 0\% & 78\% & 80\% \\
 &  & AIME21-26 & 7.91 & 7.90 & 0.59 & 0.60 & 16\% & 26\% & 1\% & 37\% & 40\% \\
 &  & LCB & 6.24 & 6.16 & 0.32 & 0.41 & 9\% & 15\% & 1\% & 48\% & 51\% \\
\cmidrule(lr){2-12}
 & \multirow{7}{*}{4B} & GSM8K & 6.66 & 2.74 & 0.00 & 3.92 & 0\% & 15\% & 0\% & 44\% & 81\% \\
 &  & MATH-500 & 7.91 & 3.09 & 0.00 & 4.82 & 0\% & 24\% & 0\% & 36\% & 76\% \\
 &  & HumanEval & 5.90 & 2.28 & 0.00 & 3.62 & 0\% & 13\% & 0\% & 51\% & 90\% \\
 &  & MBPP & 5.45 & 2.14 & 0.00 & 3.31 & 0\% & 9\% & 0\% & 54\% & 93\% \\
 &  & MT-Bench & 3.17 & 1.52 & 0.00 & 1.65 & 0\% & 3\% & 0\% & 77\% & 97\% \\
 &  & AIME21-26 & 7.72 & 2.95 & 0.00 & 4.76 & 0\% & 22\% & 0\% & 37\% & 80\% \\
 &  & LCB & 6.14 & 2.09 & 0.00 & 4.05 & 0\% & 14\% & 0\% & 50\% & 94\% \\
\midrule
\multirow{14}{*}{DFlash} & \multirow{7}{*}{8B} & GSM8K & 5.34 & 4.96 & 0.15 & 0.53 & 5\% & 10\% & 0\% & 56\% & 60\% \\
 &  & MATH-500 & 6.97 & 6.61 & 0.37 & 0.72 & 11\% & 20\% & 1\% & 44\% & 49\% \\
 &  & HumanEval & 5.39 & 5.13 & 0.21 & 0.47 & 6\% & 12\% & 1\% & 57\% & 60\% \\
 &  & MBPP & 4.63 & 4.33 & 0.09 & 0.39 & 3\% & 6\% & 0\% & 61\% & 64\% \\
 &  & MT-Bench & 2.30 & 2.18 & 0.03 & 0.15 & 1\% & 2\% & 0\% & 86\% & 87\% \\
 &  & AIME21-26 & 6.91 & 6.84 & 0.37 & 0.44 & 11\% & 21\% & 1\% & 45\% & 47\% \\
 &  & LCB & 5.45 & 5.20 & 0.19 & 0.43 & 6\% & 11\% & 0\% & 55\% & 58\% \\
\cmidrule(lr){2-12}
 & \multirow{7}{*}{4B} & GSM8K & 5.24 & 3.39 & 0.01 & 1.86 & 0\% & 9\% & 0\% & 56\% & 71\% \\
 &  & MATH-500 & 6.83 & 4.49 & 0.05 & 2.39 & 2\% & 19\% & 0\% & 45\% & 58\% \\
 &  & HumanEval & 5.43 & 3.27 & 0.03 & 2.19 & 1\% & 13\% & 0\% & 57\% & 74\% \\
 &  & MBPP & 4.63 & 3.00 & 0.01 & 1.64 & 0\% & 7\% & 0\% & 62\% & 77\% \\
 &  & MT-Bench & 2.40 & 1.72 & 0.00 & 0.69 & 0\% & 2\% & 0\% & 85\% & 92\% \\
 &  & AIME21-26 & 6.76 & 4.20 & 0.04 & 2.60 & 1\% & 20\% & 0\% & 46\% & 61\% \\
 &  & LCB & 5.30 & 3.12 & 0.03 & 2.21 & 1\% & 10\% & 0\% & 56\% & 76\% \\
\bottomrule
\end{tabular}
}
\end{table}
 
\paragraph{The distributions.} Figure~\ref{fig:naivehist} contrasts the 8B and 4B DFlare
distributions under naive expansion, with the Eq.~\ref{eq:massshift} decomposition overlaid.

\section{EOS termination and its effect on committed length}
\label{app:eos}

Section~\ref{sec:metrics} computes acceptance over EOS-terminated responses. This appendix quantifies
why. Table~\ref{tab:eos-retention} reports, for each original B16 drafter and benchmark, how many
responses terminate normally and the prompt-mean committed length with and without the capped runaways.
Termination is near-complete on every benchmark except competition math, where it drops to about two
thirds and the runaways inflate the committed length the most, as the final column of
Table~\ref{tab:eos-retention} shows. The size of the inflation grows with the generation-length cap
(here 16{,}384 tokens), since a longer cap lets each runaway accumulate more cycles before it stops. For
that reason the cycle-mean $E[n]{+}1$ inflates more than the prompt-mean $\tau$, because a runaway
contributes one prompt but many cycles. On the most-filtered cell, DFlare Qwen3-8B on AIME21-26,
including the runaways moves $\tau$ from $\EosInflDflareEightBAimeTauEos$ to
$\EosInflDflareEightBAimeTauAll$ ($\EosInflDflareEightBAimeTauDelta$) but $E[n]{+}1$ from
$\EosInflDflareEightBAimeEnEos$ to $\EosInflDflareEightBAimeEnAll$
($\EosInflDflareEightBAimeEnDelta$), so reporting either statistic over all responses would overstate it. In the table,
$\tau$ (EOS) uses each cell's own EOS-terminated responses, so it can differ by a few hundredths from the per-cell tables
of Appendix~\ref{app:tables}, which use the shared survivor set (8.68 here against 8.63 there for DFlare Qwen3-8B on
AIME21-26). $\tau$ (all) adds the runaways back, and $\tau$ inflation is the difference. A dash marks a cell where every
response terminates.

\begin{table}[H]
\centering
\small
\caption{EOS-terminated count and its effect on committed length, per original B16 drafter and benchmark.}
\label{tab:eos-retention}
\resizebox{\ifdim\width>\linewidth\linewidth\else\width\fi}{!}{\begin{tabular}{lllrrrrrr}
\toprule
\makecell{drafter \\ arch.} & target & size & bench & responses & EOS-terminated & \makecell{$\tau$ \\ (EOS)} & \makecell{$\tau$ \\ (all)} & \makecell{$\tau$ \\ inflation} \\
\midrule
\multirow{21}{*}{DFlare} & \multirow{14}{*}{Qwen3} & \multirow{7}{*}{8B} & GSM8K & 1319 & 1319 (100\%) & 7.86 & 7.86 & -- \\
 &  &  & MATH-500 & 500 & 480 (96\%) & 9.16 & 9.34 & +0.19 \\
 &  &  & HumanEval & 164 & 164 (100\%) & 7.07 & 7.07 & -- \\
 &  &  & MBPP & 257 & 257 (100\%) & 6.77 & 6.77 & -- \\
 &  &  & MT-Bench & 160 & 159 (99\%) & 5.00 & 5.05 & +0.05 \\
 &  &  & AIME21-26 & 179 & 115 (64\%) & 8.68 & 10.27 & +1.60 \\
 &  &  & LCB & 1055 & 1004 (95\%) & 7.81 & 8.13 & +0.32 \\
 &  & \multirow{7}{*}{4B} & GSM8K & 1319 & 1318 (100\%) & 7.95 & 7.96 & +0.01 \\
 &  &  & MATH-500 & 500 & 479 (96\%) & 9.14 & 9.31 & +0.17 \\
 &  &  & HumanEval & 164 & 161 (98\%) & 7.10 & 7.23 & +0.13 \\
 &  &  & MBPP & 257 & 255 (99\%) & 6.91 & 6.96 & +0.05 \\
 &  &  & MT-Bench & 160 & 160 (100\%) & 5.30 & 5.30 & -- \\
 &  &  & AIME21-26 & 179 & 114 (64\%) & 8.69 & 10.13 & +1.44 \\
 &  &  & LCB & 1055 & 978 (93\%) & 7.69 & 8.10 & +0.42 \\
\cmidrule(lr){2-9}
 & \multirow{7}{*}{Gemma4} & \multirow{7}{*}{12B} & GSM8K & 1319 & 1319 (100\%) & 7.42 & 7.42 & -- \\
 &  &  & MATH-500 & 500 & 492 (98\%) & 8.08 & 8.10 & +0.02 \\
 &  &  & HumanEval & 164 & 164 (100\%) & 8.78 & 8.78 & -- \\
 &  &  & MBPP & 257 & 256 (100\%) & 6.27 & 6.30 & +0.03 \\
 &  &  & MT-Bench & 160 & 160 (100\%) & 4.71 & 4.71 & -- \\
 &  &  & AIME21-26 & 179 & 141 (79\%) & 7.82 & 8.04 & +0.22 \\
 &  &  & LCB & 1055 & 983 (93\%) & 6.66 & 6.71 & +0.05 \\
\midrule
\multirow{14}{*}{DFlash} & \multirow{14}{*}{Qwen3} & \multirow{7}{*}{8B} & GSM8K & 1319 & 1319 (100\%) & 6.46 & 6.46 & -- \\
 &  &  & MATH-500 & 500 & 480 (96\%) & 8.02 & 8.17 & +0.15 \\
 &  &  & HumanEval & 164 & 164 (100\%) & 6.49 & 6.49 & -- \\
 &  &  & MBPP & 257 & 257 (100\%) & 5.91 & 5.91 & -- \\
 &  &  & MT-Bench & 160 & 159 (99\%) & 4.22 & 4.26 & +0.03 \\
 &  &  & AIME21-26 & 179 & 115 (64\%) & 7.73 & 8.67 & +0.95 \\
 &  &  & LCB & 1055 & 1004 (95\%) & 6.94 & 7.08 & +0.15 \\
 &  & \multirow{7}{*}{4B} & GSM8K & 1319 & 1318 (100\%) & 6.46 & 6.46 & +0.00 \\
 &  &  & MATH-500 & 500 & 479 (96\%) & 7.99 & 8.15 & +0.16 \\
 &  &  & HumanEval & 164 & 161 (98\%) & 6.64 & 6.74 & +0.10 \\
 &  &  & MBPP & 257 & 255 (99\%) & 5.95 & 5.99 & +0.04 \\
 &  &  & MT-Bench & 160 & 160 (100\%) & 4.38 & 4.38 & -- \\
 &  &  & AIME21-26 & 179 & 114 (64\%) & 7.72 & 8.84 & +1.12 \\
 &  &  & LCB & 1055 & 978 (93\%) & 6.72 & 7.02 & +0.30 \\
\bottomrule
\end{tabular}
}
\end{table}
 
A runaway need not be easy for the drafter, as the two excerpts below show. The first repeats one line
verbatim, so the drafter matches it almost perfectly and its accepted length is high. The second runs to
the cap by enumerating a fresh integer on each line, content the drafter cannot anticipate, so its
accepted length stays low. Including either kind distorts the estimand, and neither is a representative
completed response. Each excerpt lists the drafter's mean accepted length over the response.

\providecommand{\RunawayAccRepetitive}{14.9}
\paragraph{A high-acceptance runaway.}
A DFlare Qwen3-8B response on AIME21-26 that never emits a stop token and repeats one line verbatim to the generation-length cap. Because the repeated tokens are identical each cycle, the drafter predicts them almost perfectly, so the response accepts a mean of \RunawayAccRepetitive{} tokens, far above a typical completed response.
\begin{quote}\small\ttfamily\raggedright
\ldots \\
Try \$ a\_1 = 11, a\_2 = 1, a\_3 = 0 \$: Not valid \\
Try \$ a\_1 = 11, a\_2 = 1, a\_3 = 0 \$: Not valid \\
Try \$ a\_1 = 11, a\_2 = 1, a\_3 = 0 \$: Not valid
\end{quote}
\providecommand{\RunawayAccDigitString}{8.0}
\paragraph{A low-acceptance runaway.}
A DFlare Qwen3-8B response on AIME21-26 that also runs to the cap, enumerating a different large integer on each line. The lines never repeat exactly, so the drafter mispredicts the digits and the response accepts only \RunawayAccDigitString{} tokens. Runaways thus span both extremes of acceptance, and neither is a representative completed response.
\begin{quote}\small\ttfamily\raggedright
\ldots \\
Try \$ a\_1 = 10 \$, \$ a\_2 = 234 \$: 13803492693581127574869511724554050904902217944340773110325048447598592 \\
Try \$ a\_1 = 10 \$, \$ a\_2 = 235 \$: 27606985387162255149739023449108101809804435888681546220650096895197184
\end{quote} 
\section{Full per-cell result tables}
\label{app:tables}

This appendix provides the complete per-benchmark tables for all four cells and both arms, plus
the JetSpec comparison, from which the summary in Section~\ref{sec:results} is drawn. The block-ceiling
percentage is the fraction of decode cycles that accept the full block ($n = B{-}1$).

\subsection{Block-ceiling indicator, full detail.}
\label{app:block-ceiling-gate}
\begin{figure}[tbp]
\centering
\includegraphics[width=0.82\linewidth]{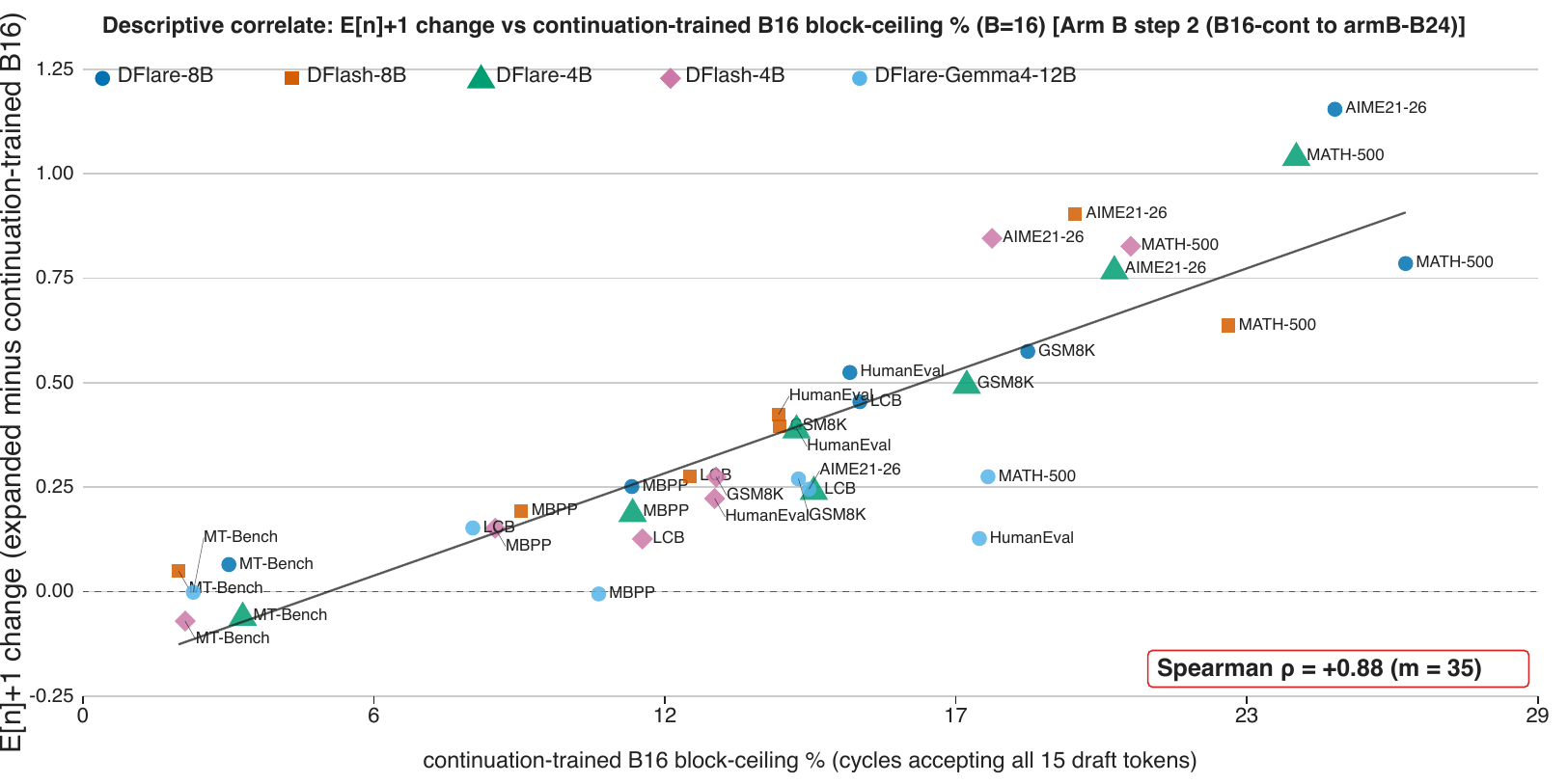}
\caption{The combined Arm-B indicator. The cycle-mean $E[n]{+}1$ improvement from the Arm-B B24 expansion (y)
versus the pre-expansion block-ceiling fraction of the continuation-trained B16 drafter (x), with the Spearman
correlation. This combined view is descriptive.}
\label{fig:gate-armb-en}
\end{figure}

\begin{figure}[tbp]
\centering
\includegraphics[width=\linewidth]{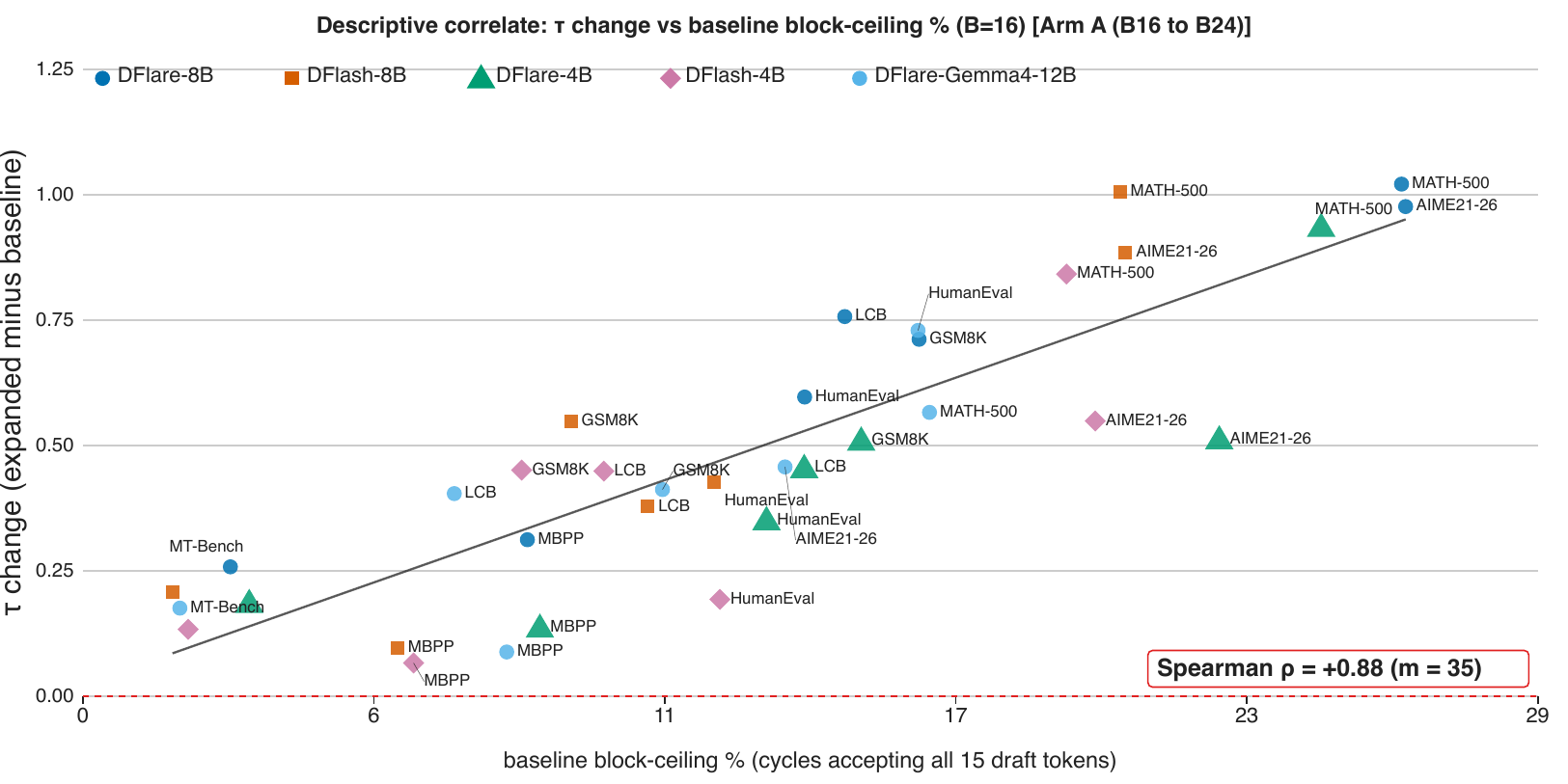}
\caption{The Arm-A one-step indicator. The $\tau$ gain of the Arm-A B24 expansion over the original B16 drafter (y)
versus the original-B16 block-ceiling fraction (x), one point per drafter-benchmark, annotated with
the Spearman rank correlation. This panel is the one-step counterpart of the
Arm-B indicator in Figure~\ref{fig:gate-armb-en}. Here, both the indicator ceiling and the expansion start from the
original B16 drafter. The sample is $\GateArmAN$ drafter-benchmark points.}
\label{fig:gate-arma}
\end{figure}

\begin{figure}[tbp]
\centering
\includegraphics[width=0.88\linewidth]{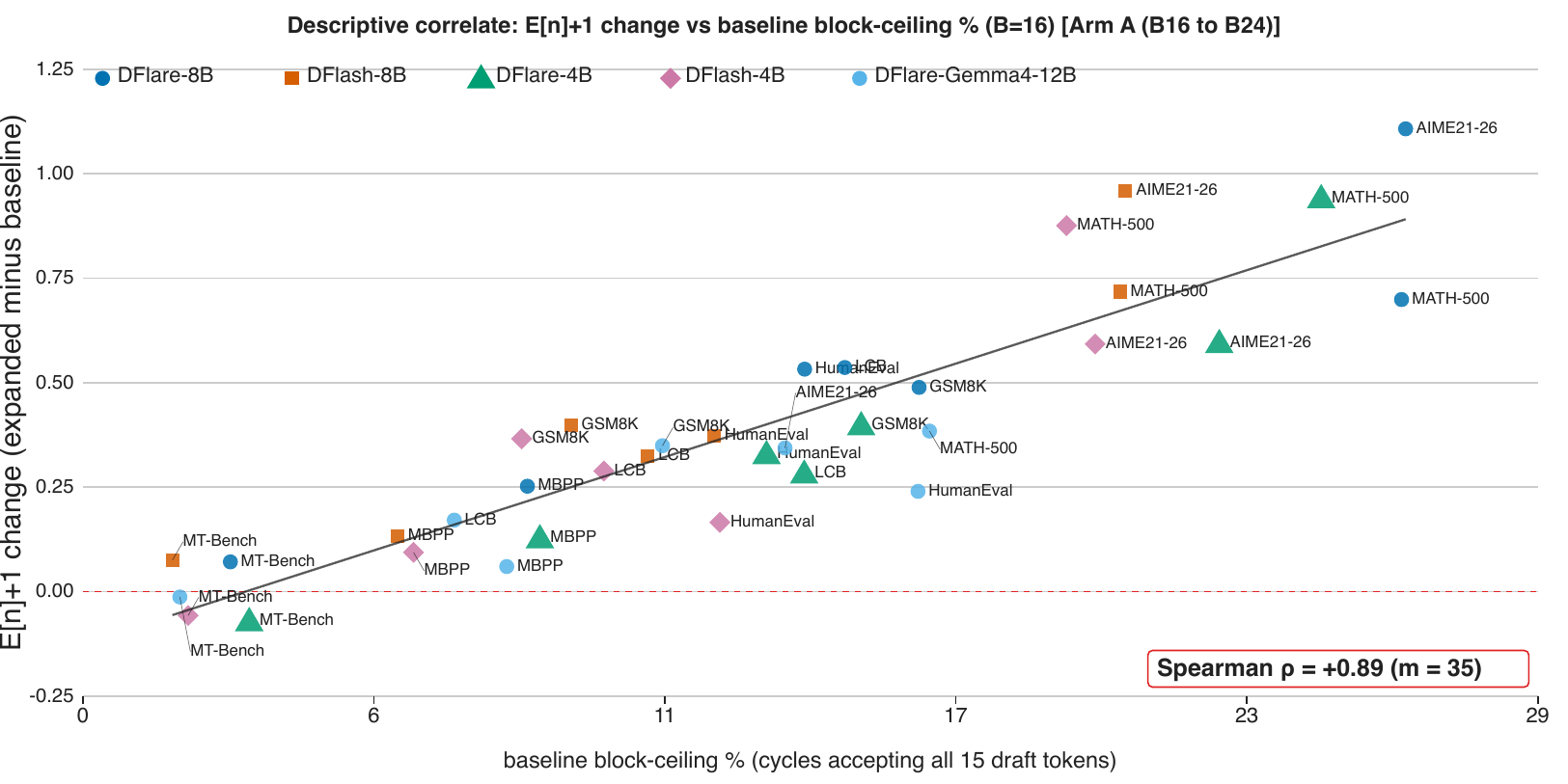}
\caption{The Arm-A indicator scored on the cycle-mean $E[n]{+}1$ gain rather than the prompt-mean $\tau$ gain.
This is the $E[n]{+}1$ companion of the Arm-A $\tau$ indicator in Figure~\ref{fig:gate-arma}, over the same
$\GateArmAN$ drafter-benchmark points and the same original-B16 ceiling on the x-axis, with only the outcome
changed. The association is close to the $\tau$ version ($\rho = \GateArmARhoEn$ against
$\GateArmARho$).}
\label{fig:gate-arma-en}
\end{figure}

\begin{figure}[tbp]
\centering
\includegraphics[width=0.88\linewidth]{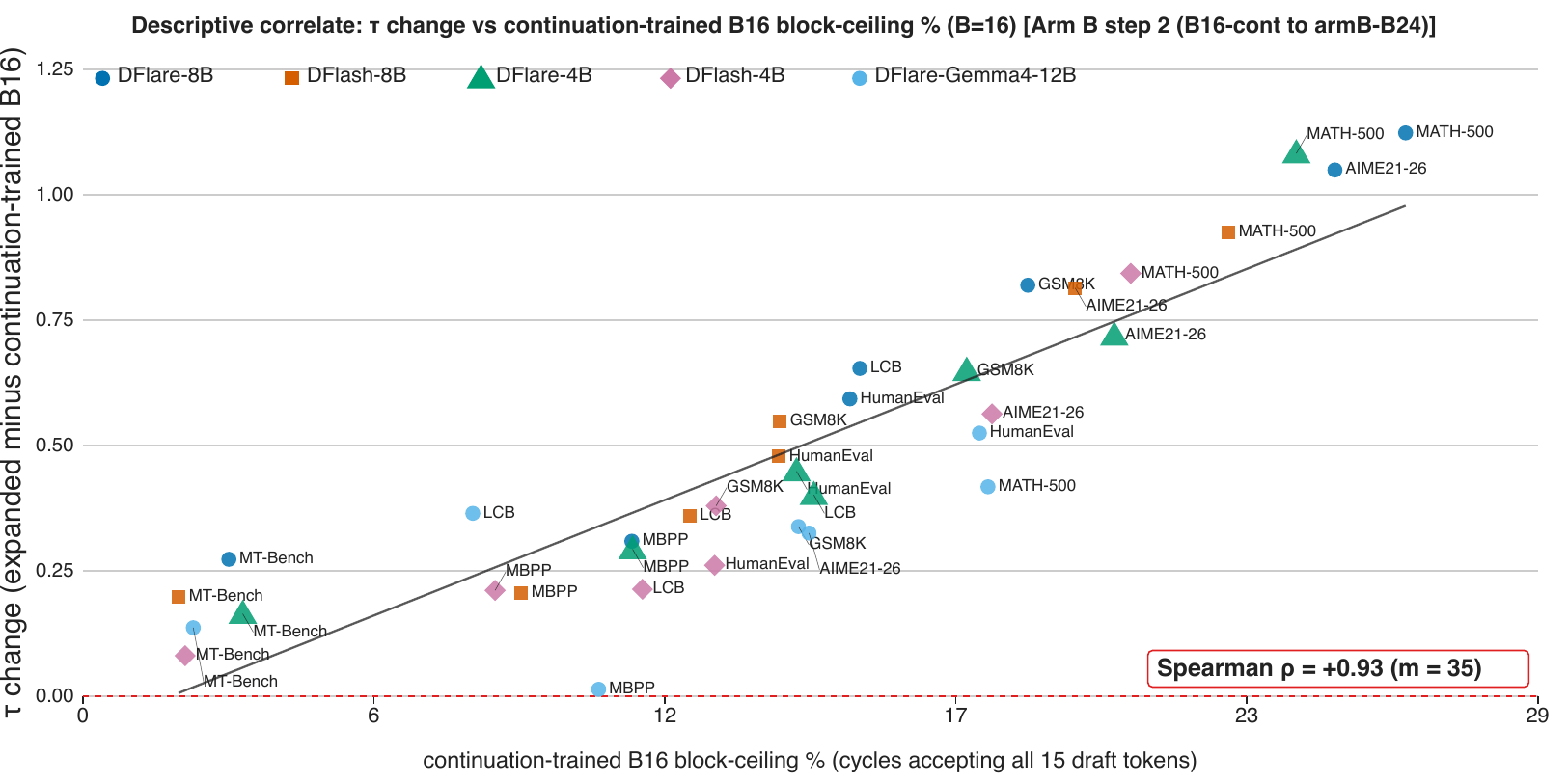}
\caption{The Arm-B indicator on the prompt-mean $\tau$ change, the $\tau$ companion of
Figure~\ref{fig:gate-armb-en}. The $\tau$ change from the Arm-B B24 expansion (y) versus the block-ceiling fraction
of the continuation-trained B16 drafter (x), with the Spearman correlation. The y-axis is the ``$\Delta\tau$ (over
Cont B16)'' column of Table~\ref{tab:summary}.}
\label{fig:gate-armb-tau}
\end{figure}

The scatter plots in this appendix combine the four Qwen3 drafters with \GemmaTarget{} for a descriptive summary.
The held-out correlations of Section~\ref{sec:res-gate} use the line fitted on Qwen3 alone.
Figures~\ref{fig:gate-armb-en}, \ref{fig:gate-arma}, \ref{fig:gate-arma-en}, and
\ref{fig:gate-armb-tau} show the four panels of the $(\tau, E[n]{+}1) \times
(\text{Arm A}, \text{Arm B})$ set, so each committed-length statistic appears for each arm.
Because these $\GateArmBContGemmaN$ points per arm come from only five drafter-target cells, the four Qwen
cells plus the \GemmaTarget{} cell, we treat the correlation as descriptive. Within each cell, we separately compute the
rank correlation across the seven benchmarks. The Arm-B $\tau$ improvement has the Spearman rank correlation
$\rho = \GateArmBContGemmaWithinDflareEightB$, $\GateArmBContGemmaWithinDflashEightB$,
$\GateArmBContGemmaWithinDflareFourB$, $\GateArmBContGemmaWithinDflashFourB$, and
$\GateArmBContGemmaWithinGemmaTwelveB$ for DFlare-8B, DFlash-8B,
DFlare-4B, DFlash-4B, and \GemmaTarget{}, at or above the Spearman two-tailed $5\%$ critical value for
\NBenchWithinWord{} benchmarks ($|\rho| \approx \SpearmanCritWithin$) in every Qwen cell, so the same
association is visible across cells.

\subsection{Family-level significance of the expansion gain}
\label{app:holm}

Table~\ref{tab:summary} reports our best configuration. To show typical behavior,
we combine all four architecture-target cells and both arms across the three high-ceiling benchmarks,
which yields \FamilySize{} comparisons. We report each comparison two ways: the incremental effect of the B24
expansion over the drafter it expands (Arm A over the original B16, Arm B over the continuation-trained
B16), and the end-to-end improvement over the original B16.
Every $\Delta\tau$ is a paired difference on the same prompts,
with a 95\% bootstrap confidence interval resampled at the prompt level (the per-cell intervals are in
Appendix~\ref{app:per-cell-results}). These \FamilySize{} comparisons are nested (cell $\times$ arm
$\times$ benchmark) and not independent, so we describe the distribution.
Measured over the drafter each B24 expands, the median $\Delta\tau$ is
$\FamilyIncrMedianDtau$ (range $\FamilyIncrMinDtau$ to $\FamilyIncrMaxDtau$). Measured end to end over the
original B16, it is $\FamilyMedianDtau$ (range $\FamilyMinDtau$ to $\FamilyMaxDtau$). All \FamilySize{} are
positive in both.

We bound the family-wise error with a Holm step-down over the family of per-comparison two-sided
p-values. Each comparison's p-value comes from the same paired prompt bootstrap as its interval
(the two-sided value is twice the smaller bootstrap tail probability that $\Delta\tau$ has the opposite
sign, clamped at one). Sorting the \FamilySize{} p-values ascending and testing the $j$-th against
$\alpha / (\FamilySize{} - j + 1)$ at $\alpha = 0.05$, all \FamilyPositiveAfterHolm{} remain
significant and positive, and none is significant and negative.
The method is a consistent gain on high-ceiling benchmarks across every cell and arm.
Extending the family to all seven benchmarks yields \FamilyAllSize{} comparisons (adding the
low-ceiling chat and short-form code benchmarks, where expansion has little room to gain). Every
$\Delta\tau$ stays positive, with a median of $\FamilyAllMedianDtau$ and a range from
$\FamilyAllMinDtau$ to $\FamilyAllMaxDtau$, and \FamilyAllPositiveAfterHolm{} of the
\FamilyAllSize{} remain significantly positive under the same Holm correction.
The three that do not clear the correction are
small positive gains on low-ceiling benchmarks for the Qwen3-4B direct-expansion (Arm A) cells, so
folding in the benchmarks where the method has least room still yields no negative anywhere.

\subsection{Reproduction check and the effect of the generation-length cap}
\label{app:repro-cap}

We include the DFlash and DFlare papers' published $\tau$ only as an external reference. We measure every number in this paper (all $\tau$, $E[n]{+}1$, and speed-up values, across both arms and all four cells) on one evaluation stack with a fixed protocol, so the comparisons are self-consistent. Rather than compare averages, we test whether each published value falls inside our 95\% bootstrap CI. Each cell reports the source paper's published $\tau$, our matched-protocol B16 $\tau$ with its CI, the percentage gap, and the verdict (yes / no, with the direction when no). The AIME21-26 set has no published counterpart, so its row is n/a and competition math is compared on the matched AIME25 row.

The tables below reproduce the published values closely on the short-generation benchmarks, where the evaluation protocols coincide, and read higher on long-generation math. That gap follows from one protocol choice external to our accept/verify path, the generation-length cap. We evaluate at a 16{,}384-token cap so long responses run to completion, whereas the source papers use a much shorter one (2048 tokens for DFlash). A shorter cap truncates the long math responses and lowers their measured committed length, so an untruncated stack reads a higher $\tau$ there and matches on the short benchmarks. The tables below apply the matched 2048-token cap, which is why they align with the published numbers. Our full-cap runs, reported in the main results, are what raise the long-generation $\tau$. Because the block-horizon-expansion conclusions rest solely on within-stack comparisons at a fixed decoding budget, they remain valid regardless of how our absolute numbers compare with the published ones.

\begin{table}[H]
\centering
\small
\caption{Qwen3-8B reproduction check: our B16 $\tau$ vs the source papers' published $\tau$.}
\label{tab:repro-qwen8b}
\resizebox{\ifdim\width>\linewidth\linewidth\else\width\fi}{!}{\begin{tabular}{lrrrrrrrr}
\toprule
\multirow{2}{*}{bench} & \multicolumn{4}{c}{DFlare} & \multicolumn{4}{c}{DFlash} \\
\cmidrule(lr){2-5} \cmidrule(lr){6-9}
 & \makecell{paper \\ $\tau$} & \makecell{our \\ B16 \\ $\tau$ \\ (cap \\ 2048)} & \makecell{\% \\ diff} & \makecell{paper $\tau$ \\ in our CI?} & \makecell{paper \\ $\tau$} & \makecell{our \\ B16 \\ $\tau$ \\ (cap \\ 2048)} & \makecell{\% \\ diff} & \makecell{paper $\tau$ \\ in our CI?} \\
\midrule
GSM8K & 7.88 & 7.86$\pm$0.07 & -0.3\% & yes & 6.54 & 6.46$\pm$0.06 & -1.3\% & no (ours<paper) \\
MATH-500 & 8.95 & 9.14$\pm$0.15 & +2.2\% & no (ours>paper) & 7.87 & 8.02$\pm$0.14 & +1.9\% & no (ours>paper) \\
AIME21-26 & n/a & 8.50$\pm$0.23 & n/a & n/a (diff. set) & n/a & 7.53$\pm$0.23 & n/a & n/a (diff. set) \\
HumanEval & 7.08 & 7.07$\pm$0.17 & -0.2\% & yes & 6.50 & 6.49$\pm$0.16 & -0.1\% & yes \\
MBPP & 6.86 & 6.77$\pm$0.16 & -1.3\% & yes & 5.95 & 5.91$\pm$0.14 & -0.6\% & yes \\
LCB & n/a & 7.96$\pm$0.12 & n/a & n/a & 7.27 & 7.07$\pm$0.11 & -2.7\% & no (ours<paper) \\
MT-Bench & 5.09 & 5.02$\pm$0.39 & -1.4\% & yes & 4.24 & 4.24$\pm$0.34 & -0.0\% & yes \\
AIME25 (matched) & 8.12 & 8.24$\pm$0.43 & +1.5\% & yes & 7.08 & 7.28$\pm$0.42 & +2.9\% & yes \\
\bottomrule
\end{tabular}
}
\end{table}

\begin{table}[H]
\centering
\small
\caption{Qwen3-4B reproduction check: our B16 $\tau$ vs the source papers' published $\tau$.}
\label{tab:repro-qwen4b}
\resizebox{\ifdim\width>\linewidth\linewidth\else\width\fi}{!}{\begin{tabular}{lrrrrrrrr}
\toprule
\multirow{2}{*}{bench} & \multicolumn{4}{c}{DFlare} & \multicolumn{4}{c}{DFlash} \\
\cmidrule(lr){2-5} \cmidrule(lr){6-9}
 & \makecell{paper \\ $\tau$} & \makecell{our \\ B16 \\ $\tau$ \\ (cap \\ 2048)} & \makecell{\% \\ diff} & \makecell{paper $\tau$ \\ in our CI?} & \makecell{paper \\ $\tau$} & \makecell{our \\ B16 \\ $\tau$ \\ (cap \\ 2048)} & \makecell{\% \\ diff} & \makecell{paper $\tau$ \\ in our CI?} \\
\midrule
GSM8K & 7.93 & 7.96$\pm$0.08 & +0.3\% & yes & 6.53 & 6.47$\pm$0.06 & -1.0\% & no (ours<paper) \\
MATH-500 & 8.97 & 9.16$\pm$0.14 & +2.1\% & no (ours>paper) & 7.84 & 8.02$\pm$0.14 & +2.3\% & no (ours>paper) \\
AIME21-26 & n/a & 8.49$\pm$0.22 & n/a & n/a (diff. set) & n/a & 7.46$\pm$0.21 & n/a & n/a (diff. set) \\
HumanEval & 7.17 & 7.17$\pm$0.20 & -0.0\% & yes & 6.64 & 6.70$\pm$0.20 & +0.9\% & yes \\
MBPP & 7.08 & 6.94$\pm$0.19 & -2.0\% & yes & 6.09 & 5.98$\pm$0.15 & -1.8\% & yes \\
LCB & n/a & 7.92$\pm$0.12 & n/a & n/a & 7.09 & 6.96$\pm$0.11 & -1.9\% & no (ours<paper) \\
MT-Bench & 5.30 & 5.30$\pm$0.41 & -0.0\% & yes & 4.35 & 4.38$\pm$0.36 & +0.7\% & yes \\
AIME25 (matched) & 8.35 & 8.11$\pm$0.52 & -2.9\% & yes & 7.27 & 7.12$\pm$0.51 & -2.0\% & yes \\
\bottomrule
\end{tabular}
}
\end{table}

The per-cell tables below share these conventions. Every $\tau$ and $E[n]{+}1$ cell carries a 95\% bootstrap CI over prompts. The final $\Delta\tau$ column is the \emph{paired} difference between the last and first stage on the same prompts, with its own 95\% bootstrap CI resampled at the prompt level, and a $\Delta\tau$ whose interval includes zero is tagged (ns), meaning the change is within noise rather than a directional gain. Original Checkpoint B16 numbers are our own evaluation rather than values copied from the source papers. On the short-form-code benchmarks (HumanEval, MBPP) a few $E[n]{+}1$ intervals are wide because those benchmarks have few, short generations, so deltas against them are less reliable than on the high-ceiling benchmarks. Competition math uses the AIME21-26 set. Each cell's table stacks two sub-panels under one caption: the top panel is Arm A (original B16 to Arm-A B24 in one step) and the bottom panel is Arm B (the B16 continuation and Arm-B B24). Both arms start from the same original B16 drafter, so it is shown once, in the top panel, and is the baseline for both the Arm-A $\Delta\tau$ (top, original B16 to Arm-A B24) and the Arm-B full-pipeline $\Delta\tau$ (bottom, original B16 to Arm-B B24). Placing the two arms adjacent lets the one-step Arm-A B24 and the two-step Arm-B B24 be read against their shared start and against each other. Each table caption below names only its drafter and target.

\subsection{Per-cell results, both arms}
\label{app:per-cell-results}

\begin{table}[H]
\centering
\small
\caption{Qwen3-8B DBloom-DFlare.}
\label{tab:armAB-dflare-qwen8b-B16ours}
\resizebox{\ifdim\width>\linewidth\linewidth\else\width\fi}{!}{\begin{tabular}{@{}c@{}}
{\itshape Arm A: direct expansion from the original B16 drafter.} \\[2pt]
\begin{tabular}{lrrrrrrrrr}
\toprule
\multirow{2}{*}{bench} & \multicolumn{4}{c}{Original Checkpoint B16} & \multicolumn{4}{c}{DBloom-DFlare Arm-A B24} & \multirow{2}{*}{\makecell{$\Delta\tau$ (Arm A) \\ $\pm$95\% CI}} \\
\cmidrule(lr){2-5} \cmidrule(lr){6-9}
 & $\tau$ & $E[n]{+}1$ & sp & ceiling\% & $\tau$ & $E[n]{+}1$ & sp & ceiling\% &  \\
\midrule
GSM8K & 7.86$\pm$0.07 & 7.70$\pm$0.24 & 5.53$\times$ & 16\% & 8.57$\pm$0.09 & 8.19$\pm$0.10 & 5.97$\times$ & 4\% & +0.71 [+0.67, +0.76] \\
MATH-500 & 9.17$\pm$0.16 & 9.06$\pm$0.21 & 6.69$\times$ & 26\% & 10.19$\pm$0.22 & 9.76$\pm$0.25 & 7.20$\times$ & 6\% & +1.02 [+0.92, +1.13] \\
AIME21-26 & 8.63$\pm$0.34 & 8.91$\pm$0.49 & 6.91$\times$ & 26\% & 9.60$\pm$0.45 & 10.02$\pm$0.75 & 7.73$\times$ & 7\% & +0.98 [+0.69, +1.28] \\
HumanEval & 7.07$\pm$0.17 & 6.96$\pm$0.18 & 5.10$\times$ & 14\% & 7.66$\pm$0.20 & 7.50$\pm$0.21 & 5.48$\times$ & 3\% & +0.60 [+0.49, +0.71] \\
MBPP & 6.77$\pm$0.16 & 6.38$\pm$0.16 & 4.65$\times$ & 9\% & 7.08$\pm$0.20 & 6.63$\pm$0.17 & 4.82$\times$ & 1\% & +0.31 [+0.22, +0.40] \\
LCB & 7.84$\pm$0.12 & 7.24$\pm$0.12 & 5.40$\times$ & 15\% & 8.60$\pm$0.15 & 7.78$\pm$0.15 & 5.79$\times$ & 3\% & +0.76 [+0.68, +0.83] \\
MT-Bench & 5.00$\pm$0.39 & 4.03$\pm$0.23 & 2.96$\times$ & 3\% & 5.26$\pm$0.46 & 4.10$\pm$0.25 & 2.93$\times$ & 0\% & +0.26 [+0.15, +0.38] \\
\bottomrule
\end{tabular}
\\[6pt]
{\itshape Arm B: continuation then expansion.} \\[2pt]
\begin{tabular}{lrrrrrrrrrr}
\toprule
\multirow{2}{*}{bench} & \multicolumn{4}{c}{DBloom-DFlare B16 continuation} & \multicolumn{4}{c}{DBloom-DFlare Arm-B B24} & \multirow{2}{*}{\makecell{$\Delta\tau$ (over Cont B16) \\ $\pm$95\% CI}} & \multirow{2}{*}{\makecell{$\Delta\tau$ (over Original B16) \\ $\pm$95\% CI}} \\
\cmidrule(lr){2-5} \cmidrule(lr){6-9}
 & $\tau$ & $E[n]{+}1$ & sp & ceiling\% & $\tau$ & $E[n]{+}1$ & sp & ceiling\% &  &  \\
\midrule
GSM8K & 8.41$\pm$0.08 & 8.18$\pm$0.23 & 5.87$\times$ & 19\% & 9.23$\pm$0.10 & 8.75$\pm$0.11 & 6.38$\times$ & 5\% & +0.82 [+0.77, +0.87] & +1.37 [+1.32, +1.43] \\
MATH-500 & 9.42$\pm$0.16 & 9.16$\pm$0.21 & 6.77$\times$ & 26\% & 10.54$\pm$0.23 & 9.95$\pm$0.26 & 7.36$\times$ & 7\% & +1.12 [+1.02, +1.23] & +1.37 [+1.26, +1.49] \\
AIME21-26 & 8.53$\pm$0.34 & 8.79$\pm$0.49 & 6.76$\times$ & 25\% & 9.58$\pm$0.45 & 9.95$\pm$0.70 & 7.67$\times$ & 7\% & +1.05 [+0.78, +1.35] & +0.95 [+0.69, +1.25] \\
HumanEval & 7.33$\pm$0.17 & 7.22$\pm$0.18 & 5.28$\times$ & 15\% & 7.93$\pm$0.20 & 7.74$\pm$0.21 & 5.66$\times$ & 4\% & +0.59 [+0.48, +0.72] & +0.86 [+0.73, +0.99] \\
MBPP & 7.41$\pm$0.19 & 6.86$\pm$0.19 & 5.00$\times$ & 11\% & 7.72$\pm$0.22 & 7.11$\pm$0.20 & 5.20$\times$ & 2\% & +0.31 [+0.22, +0.40] & +0.94 [+0.81, +1.08] \\
LCB & 8.05$\pm$0.12 & 7.41$\pm$0.12 & 5.52$\times$ & 15\% & 8.71$\pm$0.16 & 7.86$\pm$0.15 & 5.85$\times$ & 3\% & +0.65 [+0.58, +0.72] & +0.87 [+0.79, +0.94] \\
MT-Bench & 5.11$\pm$0.41 & 4.03$\pm$0.23 & 2.93$\times$ & 3\% & 5.38$\pm$0.50 & 4.10$\pm$0.26 & 2.97$\times$ & 1\% & +0.27 [+0.14, +0.42] & +0.38 [+0.23, +0.53] \\
\bottomrule
\end{tabular}
\end{tabular}
}
\end{table}

\begin{table}[H]
\centering
\small
\caption{Qwen3-8B DBloom-DFlash.}
\label{tab:armAB-dflash-qwen8b-B16ours}
\resizebox{\ifdim\width>\linewidth\linewidth\else\width\fi}{!}{\begin{tabular}{@{}c@{}}
{\itshape Arm A: direct expansion from the original B16 drafter.} \\[2pt]
\begin{tabular}{lrrrrrrrrr}
\toprule
\multirow{2}{*}{bench} & \multicolumn{4}{c}{Original Checkpoint B16} & \multicolumn{4}{c}{DBloom-DFlash Arm-A B24} & \multirow{2}{*}{\makecell{$\Delta\tau$ (Arm A) \\ $\pm$95\% CI}} \\
\cmidrule(lr){2-5} \cmidrule(lr){6-9}
 & $\tau$ & $E[n]{+}1$ & sp & ceiling\% & $\tau$ & $E[n]{+}1$ & sp & ceiling\% &  \\
\midrule
GSM8K & 6.46$\pm$0.06 & 6.34$\pm$0.19 & 4.77$\times$ & 10\% & 7.01$\pm$0.07 & 6.74$\pm$0.08 & 5.16$\times$ & 2\% & +0.55 [+0.51, +0.58] \\
MATH-500 & 8.03$\pm$0.15 & 7.97$\pm$0.21 & 6.18$\times$ & 20\% & 9.03$\pm$0.20 & 8.68$\pm$0.24 & 6.78$\times$ & 5\% & +1.01 [+0.92, +1.09] \\
AIME21-26 & 7.68$\pm$0.33 & 7.91$\pm$0.46 & 6.41$\times$ & 21\% & 8.57$\pm$0.42 & 8.87$\pm$0.61 & 7.06$\times$ & 5\% & +0.88 [+0.65, +1.12] \\
HumanEval & 6.49$\pm$0.16 & 6.39$\pm$0.17 & 4.91$\times$ & 12\% & 6.92$\pm$0.19 & 6.76$\pm$0.20 & 5.22$\times$ & 3\% & +0.43 [+0.32, +0.54] \\
MBPP & 5.91$\pm$0.14 & 5.63$\pm$0.14 & 4.31$\times$ & 6\% & 6.01$\pm$0.14 & 5.76$\pm$0.14 & 4.42$\times$ & 1\% & +0.10 [+0.02, +0.17] \\
LCB & 6.96$\pm$0.11 & 6.45$\pm$0.11 & 5.06$\times$ & 11\% & 7.34$\pm$0.13 & 6.78$\pm$0.14 & 5.30$\times$ & 2\% & +0.38 [+0.32, +0.44] \\
MT-Bench & 4.22$\pm$0.34 & 3.30$\pm$0.20 & 2.54$\times$ & 2\% & 4.43$\pm$0.41 & 3.38$\pm$0.22 & 2.57$\times$ & 0\% & +0.21 [+0.10, +0.32] \\
\bottomrule
\end{tabular}
\\[6pt]
{\itshape Arm B: continuation then expansion.} \\[2pt]
\begin{tabular}{lrrrrrrrrrr}
\toprule
\multirow{2}{*}{bench} & \multicolumn{4}{c}{DBloom-DFlash B16 continuation} & \multicolumn{4}{c}{DBloom-DFlash Arm-B B24} & \multirow{2}{*}{\makecell{$\Delta\tau$ (over Cont B16) \\ $\pm$95\% CI}} & \multirow{2}{*}{\makecell{$\Delta\tau$ (over Original B16) \\ $\pm$95\% CI}} \\
\cmidrule(lr){2-5} \cmidrule(lr){6-9}
 & $\tau$ & $E[n]{+}1$ & sp & ceiling\% & $\tau$ & $E[n]{+}1$ & sp & ceiling\% &  &  \\
\midrule
GSM8K & 7.60$\pm$0.07 & 7.32$\pm$0.16 & 5.51$\times$ & 14\% & 8.15$\pm$0.09 & 7.72$\pm$0.10 & 5.92$\times$ & 3\% & +0.55 [+0.51, +0.59] & +1.69 [+1.64, +1.74] \\
MATH-500 & 8.81$\pm$0.17 & 8.53$\pm$0.20 & 6.62$\times$ & 23\% & 9.73$\pm$0.22 & 9.17$\pm$0.25 & 7.15$\times$ & 5\% & +0.93 [+0.83, +1.02] & +1.71 [+1.59, +1.82] \\
AIME21-26 & 7.92$\pm$0.33 & 7.91$\pm$0.38 & 6.23$\times$ & 20\% & 8.73$\pm$0.41 & 8.82$\pm$0.48 & 6.85$\times$ & 4\% & +0.81 [+0.62, +1.03] & +1.05 [+0.82, +1.28] \\
HumanEval & 6.84$\pm$0.17 & 6.72$\pm$0.18 & 5.19$\times$ & 14\% & 7.32$\pm$0.19 & 7.14$\pm$0.20 & 5.48$\times$ & 3\% & +0.48 [+0.37, +0.59] & +0.83 [+0.71, +0.95] \\
MBPP & 6.72$\pm$0.17 & 6.26$\pm$0.17 & 4.79$\times$ & 9\% & 6.93$\pm$0.19 & 6.45$\pm$0.18 & 4.96$\times$ & 1\% & +0.21 [+0.12, +0.29] & +1.01 [+0.89, +1.14] \\
LCB & 7.29$\pm$0.11 & 6.72$\pm$0.11 & 5.30$\times$ & 12\% & 7.65$\pm$0.13 & 7.00$\pm$0.14 & 5.48$\times$ & 2\% & +0.36 [+0.30, +0.42] & +0.68 [+0.62, +0.75] \\
MT-Bench & 4.39$\pm$0.37 & 3.38$\pm$0.21 & 2.59$\times$ & 2\% & 4.59$\pm$0.44 & 3.43$\pm$0.23 & 2.63$\times$ & 0\% & +0.20 [+0.09, +0.31] & +0.36 [+0.23, +0.50] \\
\bottomrule
\end{tabular}
\end{tabular}
}
\end{table}

\begin{table}[H]
\centering
\small
\caption{Qwen3-4B DBloom-DFlare.}
\label{tab:armAB-dflare-qwen4b-B16ours}
\resizebox{\ifdim\width>\linewidth\linewidth\else\width\fi}{!}{\begin{tabular}{@{}c@{}}
{\itshape Arm A: direct expansion from the original B16 drafter.} \\[2pt]
\begin{tabular}{lrrrrrrrrr}
\toprule
\multirow{2}{*}{bench} & \multicolumn{4}{c}{Original Checkpoint B16} & \multicolumn{4}{c}{DBloom-DFlare Arm-A B24} & \multirow{2}{*}{\makecell{$\Delta\tau$ (Arm A) \\ $\pm$95\% CI}} \\
\cmidrule(lr){2-5} \cmidrule(lr){6-9}
 & $\tau$ & $E[n]{+}1$ & sp & ceiling\% & $\tau$ & $E[n]{+}1$ & sp & ceiling\% &  \\
\midrule
GSM8K & 7.96$\pm$0.08 & 7.66$\pm$0.09 & 5.57$\times$ & 15\% & 8.47$\pm$0.09 & 8.06$\pm$0.11 & 5.86$\times$ & 3\% & +0.51 [+0.47, +0.55] \\
MATH-500 & 9.18$\pm$0.16 & 8.91$\pm$0.19 & 6.56$\times$ & 24\% & 10.11$\pm$0.21 & 9.85$\pm$0.24 & 7.25$\times$ & 7\% & +0.94 [+0.84, +1.03] \\
AIME21-26 & 8.68$\pm$0.32 & 8.72$\pm$0.42 & 6.97$\times$ & 22\% & 9.19$\pm$0.40 & 9.31$\pm$0.44 & 7.49$\times$ & 5\% & +0.51 [+0.30, +0.72] \\
HumanEval & 7.10$\pm$0.19 & 6.90$\pm$0.20 & 5.04$\times$ & 13\% & 7.46$\pm$0.22 & 7.23$\pm$0.25 & 5.28$\times$ & 3\% & +0.35 [+0.26, +0.45] \\
MBPP & 6.91$\pm$0.18 & 6.45$\pm$0.17 & 4.70$\times$ & 9\% & 7.04$\pm$0.18 & 6.57$\pm$0.18 & 4.81$\times$ & 1\% & +0.14 [+0.05, +0.23] \\
LCB & 7.71$\pm$0.12 & 7.14$\pm$0.12 & 5.34$\times$ & 14\% & 8.17$\pm$0.14 & 7.42$\pm$0.14 & 5.54$\times$ & 2\% & +0.45 [+0.39, +0.52] \\
MT-Bench & 5.30$\pm$0.41 & 4.17$\pm$0.25 & 3.02$\times$ & 3\% & 5.48$\pm$0.48 & 4.10$\pm$0.24 & 2.99$\times$ & 0\% & +0.19 [+0.03, +0.34] \\
\bottomrule
\end{tabular}
\\[6pt]
{\itshape Arm B: continuation then expansion.} \\[2pt]
\begin{tabular}{lrrrrrrrrrr}
\toprule
\multirow{2}{*}{bench} & \multicolumn{4}{c}{DBloom-DFlare B16 continuation} & \multicolumn{4}{c}{DBloom-DFlare Arm-B B24} & \multirow{2}{*}{\makecell{$\Delta\tau$ (over Cont B16) \\ $\pm$95\% CI}} & \multirow{2}{*}{\makecell{$\Delta\tau$ (over Original B16) \\ $\pm$95\% CI}} \\
\cmidrule(lr){2-5} \cmidrule(lr){6-9}
 & $\tau$ & $E[n]{+}1$ & sp & ceiling\% & $\tau$ & $E[n]{+}1$ & sp & ceiling\% &  &  \\
\midrule
GSM8K & 8.45$\pm$0.08 & 8.08$\pm$0.10 & 5.88$\times$ & 18\% & 9.10$\pm$0.11 & 8.58$\pm$0.12 & 6.24$\times$ & 4\% & +0.65 [+0.60, +0.70] & +1.14 [+1.09, +1.20] \\
MATH-500 & 9.30$\pm$0.16 & 8.94$\pm$0.19 & 6.60$\times$ & 24\% & 10.39$\pm$0.21 & 9.98$\pm$0.25 & 7.32$\times$ & 6\% & +1.08 [+0.99, +1.18] & +1.21 [+1.11, +1.31] \\
AIME21-26 & 8.50$\pm$0.32 & 8.51$\pm$0.40 & 6.76$\times$ & 21\% & 9.22$\pm$0.40 & 9.29$\pm$0.43 & 7.46$\times$ & 5\% & +0.72 [+0.51, +0.92] & +0.55 [+0.33, +0.76] \\
HumanEval & 7.29$\pm$0.20 & 7.08$\pm$0.22 & 5.15$\times$ & 14\% & 7.74$\pm$0.24 & 7.47$\pm$0.27 & 5.44$\times$ & 3\% & +0.45 [+0.35, +0.55] & +0.63 [+0.51, +0.75] \\
MBPP & 7.49$\pm$0.21 & 6.89$\pm$0.21 & 5.02$\times$ & 11\% & 7.78$\pm$0.22 & 7.08$\pm$0.23 & 5.18$\times$ & 1\% & +0.29 [+0.21, +0.38] & +0.87 [+0.75, +1.00] \\
LCB & 7.89$\pm$0.12 & 7.27$\pm$0.12 & 5.39$\times$ & 15\% & 8.29$\pm$0.14 & 7.51$\pm$0.14 & 5.59$\times$ & 2\% & +0.40 [+0.34, +0.47] & +0.58 [+0.51, +0.65] \\
MT-Bench & 5.37$\pm$0.42 & 4.14$\pm$0.25 & 3.02$\times$ & 3\% & 5.53$\pm$0.50 & 4.08$\pm$0.25 & 2.96$\times$ & 0\% & +0.16 [+0.01, +0.31] & +0.23 [+0.06, +0.40] \\
\bottomrule
\end{tabular}
\end{tabular}
}
\end{table}

\begin{table}[H]
\centering
\small
\caption{Qwen3-4B DBloom-DFlash.}
\label{tab:armAB-dflash-qwen4b-B16ours}
\resizebox{\ifdim\width>\linewidth\linewidth\else\width\fi}{!}{\begin{tabular}{@{}c@{}}
{\itshape Arm A: direct expansion from the original B16 drafter.} \\[2pt]
\begin{tabular}{lrrrrrrrrr}
\toprule
\multirow{2}{*}{bench} & \multicolumn{4}{c}{Original Checkpoint B16} & \multicolumn{4}{c}{DBloom-DFlash Arm-A B24} & \multirow{2}{*}{\makecell{$\Delta\tau$ (Arm A) \\ $\pm$95\% CI}} \\
\cmidrule(lr){2-5} \cmidrule(lr){6-9}
 & $\tau$ & $E[n]{+}1$ & sp & ceiling\% & $\tau$ & $E[n]{+}1$ & sp & ceiling\% &  \\
\midrule
GSM8K & 6.47$\pm$0.06 & 6.24$\pm$0.07 & 4.77$\times$ & 9\% & 6.92$\pm$0.08 & 6.60$\pm$0.08 & 5.06$\times$ & 1\% & +0.45 [+0.42, +0.49] \\
MATH-500 & 8.02$\pm$0.15 & 7.83$\pm$0.18 & 6.10$\times$ & 19\% & 8.86$\pm$0.20 & 8.71$\pm$0.25 & 6.75$\times$ & 5\% & +0.84 [+0.76, +0.92] \\
AIME21-26 & 7.69$\pm$0.33 & 7.76$\pm$0.46 & 6.58$\times$ & 20\% & 8.24$\pm$0.40 & 8.35$\pm$0.46 & 7.15$\times$ & 4\% & +0.55 [+0.33, +0.76] \\
HumanEval & 6.64$\pm$0.20 & 6.43$\pm$0.20 & 4.92$\times$ & 13\% & 6.83$\pm$0.22 & 6.60$\pm$0.24 & 5.06$\times$ & 2\% & +0.19 [+0.10, +0.29] \\
MBPP & 5.95$\pm$0.14 & 5.63$\pm$0.14 & 4.34$\times$ & 7\% & 6.01$\pm$0.14 & 5.72$\pm$0.15 & 4.40$\times$ & 1\% & +0.07 [-0.01, +0.14] (ns) \\
LCB & 6.75$\pm$0.10 & 6.30$\pm$0.11 & 4.95$\times$ & 10\% & 7.20$\pm$0.13 & 6.59$\pm$0.12 & 5.18$\times$ & 1\% & +0.45 [+0.39, +0.51] \\
MT-Bench & 4.38$\pm$0.36 & 3.40$\pm$0.21 & 2.61$\times$ & 2\% & 4.52$\pm$0.40 & 3.35$\pm$0.20 & 2.56$\times$ & 0\% & +0.13 [+0.02, +0.26] \\
\bottomrule
\end{tabular}
\\[6pt]
{\itshape Arm B: continuation then expansion.} \\[2pt]
\begin{tabular}{lrrrrrrrrrr}
\toprule
\multirow{2}{*}{bench} & \multicolumn{4}{c}{DBloom-DFlash B16 continuation} & \multicolumn{4}{c}{DBloom-DFlash Arm-B B24} & \multirow{2}{*}{\makecell{$\Delta\tau$ (over Cont B16) \\ $\pm$95\% CI}} & \multirow{2}{*}{\makecell{$\Delta\tau$ (over Original B16) \\ $\pm$95\% CI}} \\
\cmidrule(lr){2-5} \cmidrule(lr){6-9}
 & $\tau$ & $E[n]{+}1$ & sp & ceiling\% & $\tau$ & $E[n]{+}1$ & sp & ceiling\% &  &  \\
\midrule
GSM8K & 7.50$\pm$0.08 & 7.16$\pm$0.09 & 5.47$\times$ & 13\% & 7.88$\pm$0.09 & 7.43$\pm$0.10 & 5.70$\times$ & 2\% & +0.38 [+0.34, +0.42] & +1.42 [+1.37, +1.47] \\
MATH-500 & 8.61$\pm$0.16 & 8.24$\pm$0.19 & 6.42$\times$ & 21\% & 9.45$\pm$0.21 & 9.07$\pm$0.24 & 7.00$\times$ & 5\% & +0.84 [+0.76, +0.93] & +1.44 [+1.34, +1.54] \\
AIME21-26 & 7.79$\pm$0.32 & 7.56$\pm$0.39 & 6.22$\times$ & 18\% & 8.36$\pm$0.39 & 8.40$\pm$0.41 & 7.11$\times$ & 4\% & +0.56 [+0.34, +0.79] & +0.66 [+0.45, +0.86] \\
HumanEval & 6.83$\pm$0.19 & 6.63$\pm$0.20 & 5.07$\times$ & 13\% & 7.10$\pm$0.21 & 6.86$\pm$0.25 & 5.27$\times$ & 2\% & +0.26 [+0.18, +0.34] & +0.46 [+0.36, +0.56] \\
MBPP & 6.75$\pm$0.18 & 6.23$\pm$0.18 & 4.78$\times$ & 8\% & 6.96$\pm$0.20 & 6.38$\pm$0.20 & 4.90$\times$ & 1\% & +0.21 [+0.13, +0.29] & +1.01 [+0.90, +1.13] \\
LCB & 7.20$\pm$0.11 & 6.61$\pm$0.12 & 5.20$\times$ & 11\% & 7.41$\pm$0.13 & 6.74$\pm$0.12 & 5.31$\times$ & 1\% & +0.21 [+0.16, +0.27] & +0.66 [+0.60, +0.72] \\
MT-Bench & 4.58$\pm$0.39 & 3.46$\pm$0.22 & 2.65$\times$ & 2\% & 4.66$\pm$0.43 & 3.39$\pm$0.21 & 2.60$\times$ & 0\% & +0.08 [-0.03, +0.21] (ns) & +0.28 [+0.15, +0.42] \\
\bottomrule
\end{tabular}
\end{tabular}
}
\end{table}

\begin{table}[H]
\centering
\small
\caption{Gemma-4-12B-IT DBloom-DFlare.}
\label{tab:gemma-b24}
\resizebox{\ifdim\width>\linewidth\linewidth\else\width\fi}{!}{\begin{tabular}{@{}c@{}}
{\itshape Arm A: direct expansion from the original B16 drafter.} \\[2pt]
\begin{tabular}{lrrrrrrrrr}
\toprule
\multirow{2}{*}{bench} & \multicolumn{4}{c}{Original Checkpoint B16} & \multicolumn{4}{c}{DBloom-DFlare Arm-A B24} & \multirow{2}{*}{\makecell{$\Delta\tau$ (Arm A) \\ $\pm$95\% CI}} \\
\cmidrule(lr){2-5} \cmidrule(lr){6-9}
 & $\tau$ & $E[n]{+}1$ & sp & ceiling\% & $\tau$ & $E[n]{+}1$ & sp & ceiling\% &  \\
\midrule
GSM8K & 7.42$\pm$0.07 & 7.16$\pm$0.09 & 6.30$\times$ & 11\% & 7.84$\pm$0.08 & 7.51$\pm$0.10 & 6.58$\times$ & 2\% & +0.41 [+0.38, +0.44] \\
MATH-500 & 8.08$\pm$0.12 & 7.81$\pm$0.16 & 6.91$\times$ & 17\% & 8.65$\pm$0.16 & 8.24$\pm$0.19 & 7.25$\times$ & 3\% & +0.58 [+0.52, +0.64] \\
AIME21-26 & 7.89$\pm$0.21 & 7.56$\pm$0.21 & 7.77$\times$ & 14\% & 8.36$\pm$0.27 & 7.91$\pm$0.26 & 8.00$\times$ & 2\% & +0.46 [+0.34, +0.59] \\
HumanEval & 8.78$\pm$0.34 & 7.41$\pm$0.40 & 6.49$\times$ & 16\% & 9.51$\pm$0.44 & 7.65$\pm$0.45 & 6.67$\times$ & 2\% & +0.73 [+0.56, +0.89] \\
MBPP & 6.27$\pm$0.12 & 6.07$\pm$0.12 & 5.34$\times$ & 8\% & 6.36$\pm$0.12 & 6.14$\pm$0.13 & 5.37$\times$ & 1\% & +0.09 [+0.03, +0.15] \\
LCB & 6.70$\pm$0.10 & 5.85$\pm$0.07 & 4.90$\times$ & 8\% & 7.11$\pm$0.13 & 6.02$\pm$0.09 & 5.13$\times$ & 1\% & +0.40 [+0.35, +0.46] \\
MT-Bench & 4.71$\pm$0.40 & 3.62$\pm$0.22 & 3.14$\times$ & 2\% & 4.89$\pm$0.45 & 3.61$\pm$0.23 & 3.12$\times$ & 0\% & +0.18 [+0.09, +0.27] \\
\bottomrule
\end{tabular}
\\[6pt]
{\itshape Arm B: continuation then expansion.} \\[2pt]
\begin{tabular}{lrrrrrrrrrr}
\toprule
\multirow{2}{*}{bench} & \multicolumn{4}{c}{DFlare-Gemma B16 continuation} & \multicolumn{4}{c}{DBloom-DFlare Arm-B B24} & \multirow{2}{*}{\makecell{$\Delta\tau$ (over Cont B16) \\ $\pm$95\% CI}} & \multirow{2}{*}{\makecell{$\Delta\tau$ (over Original B16) \\ $\pm$95\% CI}} \\
\cmidrule(lr){2-5} \cmidrule(lr){6-9}
 & $\tau$ & $E[n]{+}1$ & sp & ceiling\% & $\tau$ & $E[n]{+}1$ & sp & ceiling\% &  &  \\
\midrule
GSM8K & 8.07$\pm$0.07 & 7.73$\pm$0.10 & 6.79$\times$ & 14\% & 8.41$\pm$0.09 & 8.00$\pm$0.12 & 7.02$\times$ & 2\% & +0.34 [+0.30, +0.37] & +0.98 [+0.94, +1.02] \\
MATH-500 & 8.48$\pm$0.13 & 8.13$\pm$0.17 & 7.22$\times$ & 18\% & 8.91$\pm$0.16 & 8.46$\pm$0.20 & 7.42$\times$ & 2\% & +0.43 [+0.37, +0.49] & +0.83 [+0.76, +0.90] \\
AIME21-26 & 8.11$\pm$0.21 & 7.75$\pm$0.22 & 7.93$\times$ & 15\% & 8.43$\pm$0.27 & 7.98$\pm$0.26 & 8.14$\times$ & 2\% & +0.32 [+0.20, +0.44] & +0.53 [+0.41, +0.66] \\
HumanEval & 9.22$\pm$0.34 & 7.82$\pm$0.42 & 6.85$\times$ & 18\% & 9.74$\pm$0.42 & 7.95$\pm$0.47 & 6.94$\times$ & 2\% & +0.52 [+0.37, +0.68] & +0.96 [+0.80, +1.13] \\
MBPP & 6.89$\pm$0.15 & 6.62$\pm$0.15 & 5.80$\times$ & 10\% & 6.91$\pm$0.15 & 6.63$\pm$0.15 & 5.80$\times$ & 1\% & +0.02 [-0.04, +0.08] (ns) & +0.64 [+0.57, +0.71] \\
LCB & 6.96$\pm$0.11 & 6.01$\pm$0.08 & 5.11$\times$ & 8\% & 7.32$\pm$0.14 & 6.16$\pm$0.10 & 5.25$\times$ & 1\% & +0.36 [+0.31, +0.42] & +0.62 [+0.57, +0.68] \\
MT-Bench & 4.86$\pm$0.41 & 3.68$\pm$0.23 & 3.19$\times$ & 2\% & 5.00$\pm$0.46 & 3.68$\pm$0.24 & 3.19$\times$ & 0\% & +0.14 [+0.00, +0.29] & +0.29 [+0.16, +0.42] \\
\bottomrule
\end{tabular}
\end{tabular}
}
\end{table}

\begin{table}[H]
\centering
\small
\caption{GPT-OSS-20B DBloom-DFlash. A single Arm-A expansion of the original B8 drafter to B12, the same 1.5x step as Qwen3 B16 to B24. This cell checks acceptance on a third model family.}
\label{tab:gptoss-dflash-b24}
\resizebox{\ifdim\width>\linewidth\linewidth\else\width\fi}{!}{\begin{tabular}{@{}c@{}}
{\itshape Arm A: expansion to B12 from the original B8 drafter.} \\[2pt]
\begin{tabular}{lrrrrrrr}
\toprule
\multirow{2}{*}{bench} & \multicolumn{3}{c}{Z Lab DFlash B8} & \multicolumn{3}{c}{DBloom-DFlash Arm-A B12} & \multirow{2}{*}{\makecell{$\Delta\tau$ (Arm A) \\ $\pm$95\% CI}} \\
\cmidrule(lr){2-4} \cmidrule(lr){5-7}
 & $\tau$ & $E[n]{+}1$ & ceiling\% & $\tau$ & $E[n]{+}1$ & ceiling\% &  \\
\midrule
GSM8K & 4.61$\pm$0.03 & 4.53$\pm$0.08 & 26\% & 5.52$\pm$0.04 & 5.36$\pm$0.10 & 13\% & +0.91 [+0.88, +0.94] \\
MATH-500 & 4.98$\pm$0.05 & 4.70$\pm$0.15 & 29\% & 5.93$\pm$0.08 & 5.51$\pm$0.19 & 14\% & +0.95 [+0.90, +1.00] \\
AIME21-26 & 4.56$\pm$0.11 & 4.25$\pm$0.14 & 21\% & 5.15$\pm$0.15 & 4.69$\pm$0.16 & 8\% & +0.59 [+0.50, +0.69] \\
HumanEval & 4.26$\pm$0.07 & 4.27$\pm$0.08 & 22\% & 4.73$\pm$0.10 & 4.76$\pm$0.12 & 10\% & +0.47 [+0.41, +0.53] \\
MBPP & 4.16$\pm$0.08 & 4.48$\pm$0.19 & 27\% & 4.69$\pm$0.13 & 5.11$\pm$0.27 & 14\% & +0.53 [+0.40, +0.66] \\
LCB & 4.07$\pm$0.03 & 3.70$\pm$0.08 & 14\% & 4.40$\pm$0.04 & 3.87$\pm$0.09 & 4\% & +0.33 [+0.30, +0.36] \\
MT-Bench & 3.86$\pm$0.17 & 3.76$\pm$0.34 & 20\% & 4.39$\pm$0.25 & 4.10$\pm$0.43 & 9\% & +0.52 [+0.36, +0.69] \\
\bottomrule
\end{tabular}
\end{tabular}
}
\end{table}

\begin{table}[H]
\centering
\small
\caption{GPT-OSS-20B DBloom-DFlare. A single Arm-A expansion of the original B8 drafter to B12, the same 1.5x step as Qwen3 B16 to B24. This cell checks acceptance on a third model family.}
\label{tab:gptoss-b24}
\resizebox{\ifdim\width>\linewidth\linewidth\else\width\fi}{!}{\begin{tabular}{@{}c@{}}
{\itshape Arm A: expansion to B12 from the original B8 drafter.} \\[2pt]
\begin{tabular}{lrrrrrrr}
\toprule
\multirow{2}{*}{bench} & \multicolumn{3}{c}{AngelSlim DFlare B8} & \multicolumn{3}{c}{DBloom-DFlare Arm-A B12} & \multirow{2}{*}{\makecell{$\Delta\tau$ (Arm A) \\ $\pm$95\% CI}} \\
\cmidrule(lr){2-4} \cmidrule(lr){5-7}
 & $\tau$ & $E[n]{+}1$ & ceiling\% & $\tau$ & $E[n]{+}1$ & ceiling\% &  \\
\midrule
GSM8K & 5.02$\pm$0.03 & 4.44$\pm$0.35 & 26\% & 5.73$\pm$0.04 & 5.12$\pm$0.35 & 11\% & +0.72 [+0.69, +0.75] \\
MATH-500 & 5.16$\pm$0.08 & 3.51$\pm$0.39 & 18\% & 5.96$\pm$0.10 & 4.21$\pm$0.49 & 8\% & +0.80 [+0.73, +0.86] \\
AIME21-26 & 4.03$\pm$0.23 & 2.73$\pm$0.26 & 10\% & 4.42$\pm$0.29 & 2.81$\pm$0.30 & 3\% & +0.39 [+0.16, +0.61] \\
HumanEval & 4.54$\pm$0.06 & 4.49$\pm$0.07 & 24\% & 5.01$\pm$0.10 & 4.54$\pm$0.47 & 8\% & +0.47 [+0.40, +0.54] \\
MBPP & 4.39$\pm$0.06 & 4.04$\pm$0.24 & 22\% & 4.82$\pm$0.08 & 4.45$\pm$0.27 & 8\% & +0.43 [+0.35, +0.52] \\
LCB & 4.08$\pm$0.05 & 2.85$\pm$0.15 & 9\% & 4.41$\pm$0.06 & 2.98$\pm$0.16 & 2\% & +0.32 [+0.28, +0.37] \\
MT-Bench & 3.77$\pm$0.21 & 2.37$\pm$0.42 & 8\% & 4.21$\pm$0.26 & 3.04$\pm$0.37 & 4\% & +0.44 [+0.27, +0.61] \\
\bottomrule
\end{tabular}
\end{tabular}
}
\end{table}

\subsection{Gemma-4-12B-IT: original DFlash vs our DFlare at B16}

This table is a reference point rather than a matched-budget architecture comparison. The DFlash column is Z Lab's original Gemma-4-12B-IT drafter, while the DFlare column is our from-scratch DFlare drafter for the same target. They differ in training budget and data, so the gap reflects our DFlare training as much as the architecture. $\Delta\tau$ (last minus first) is our DFlare-ours B16 minus original DFlash at B16.

\begin{table}[H]
\centering
\small
\caption{Gemma-4-12B-IT at B16, original DFlash vs our DFlare.}
\label{tab:armB-dflare-gemma12b-B16}
\resizebox{\ifdim\width>\linewidth\linewidth\else\width\fi}{!}{\begin{tabular}{lrrrrrrrrr}
\toprule
\multirow{2}{*}{bench} & \multicolumn{4}{c}{DFlash Z Lab B16 (original)} & \multicolumn{4}{c}{DFlare-ours B16} & \multirow{2}{*}{\makecell{$\Delta\tau$ (last-first) \\ $\pm$95\% CI}} \\
\cmidrule(lr){2-5} \cmidrule(lr){6-9}
 & $\tau$ & $E[n]{+}1$ & sp & ceiling\% & $\tau$ & $E[n]{+}1$ & sp & ceiling\% &  \\
\midrule
GSM8K & 4.92$\pm$0.04 & 4.80$\pm$0.04 & 4.35$\times$ & 1\% & 7.42$\pm$0.07 & 7.16$\pm$0.09 & 6.30$\times$ & 11\% & +2.51 [+2.46, +2.55] \\
MATH-500 & 5.46$\pm$0.07 & 5.29$\pm$0.09 & 4.83$\times$ & 2\% & 8.08$\pm$0.12 & 7.81$\pm$0.16 & 6.91$\times$ & 17\% & +2.62 [+2.54, +2.70] \\
AIME21-26 & 5.10$\pm$0.10 & 4.88$\pm$0.12 & 4.82$\times$ & 1\% & 7.82$\pm$0.20 & 7.50$\pm$0.26 & 7.33$\times$ & 14\% & +2.72 [+2.59, +2.85] \\
HumanEval & 4.65$\pm$0.12 & 4.39$\pm$0.11 & 3.97$\times$ & 1\% & 8.78$\pm$0.34 & 7.41$\pm$0.40 & 6.49$\times$ & 16\% & +4.13 [+3.85, +4.42] \\
MBPP & 4.15$\pm$0.05 & 4.08$\pm$0.06 & 3.69$\times$ & 0\% & 6.27$\pm$0.12 & 6.07$\pm$0.12 & 5.34$\times$ & 8\% & +2.12 [+2.04, +2.21] \\
LCB & 4.27$\pm$0.05 & 3.66$\pm$0.06 & 2.93$\times$ & 0\% & 6.66$\pm$0.10 & 5.79$\pm$0.07 & 4.61$\times$ & 7\% & +2.39 [+2.33, +2.45] \\
MT-Bench & 3.19$\pm$0.17 & 2.81$\pm$0.13 & 2.53$\times$ & 0\% & 4.71$\pm$0.40 & 3.62$\pm$0.22 & 3.14$\times$ & 2\% & +1.52 [+1.26, +1.81] \\
\bottomrule
\end{tabular}
}
\end{table} 
\section{Matched-prompt JetSpec comparison}
\label{app:jetspec-paired}

Figure~\ref{fig:jetspec} plots the paired committed-length difference. This
appendix gives the full per-budget tables for both arms. For each JetSpec budget
and benchmark we keep only prompts that terminated under both our drafter and that budget, then compute
the per-prompt paired $\Delta\tau$ (ours minus JetSpec) with a 95\% paired-bootstrap confidence interval,
the same estimator the expansion tables use.
A positive change whose interval excludes zero is a significant win, and an interval that includes zero is inconclusive.
Arm B does not show significant loss on any
benchmark at any budget, and Arm A does not show any through budget \JetArmABeats. No Arm-B interval falls
entirely below zero, and the two benchmarks whose Arm-B intervals include zero at the 256-node budget,
AIME21-26 and MT-Bench, are inconclusive rather than demonstrated equal. The direct-expansion Arm A has
one significant loss, on MBPP at the 256-node budget.
Pairing removes the differential EOS filtering of the two methods. Tables~\ref{tab:jetspec-paired-armb} and
\ref{tab:jetspec-paired-arma} report both arms and fold two budgets into each row.

\begin{table}[H]
\centering
\small
\caption{Matched-prompt JetSpec comparison (Qwen3-8B, DFlare DBloom Arm-B B24). Paired change $\Delta\tau$ (ours minus JetSpec) with a 95\% interval, by tree-node budget and benchmark.}
\label{tab:jetspec-paired-armb}
\resizebox{\ifdim\width>\linewidth\linewidth\else\width\fi}{!}{\begin{tabular}{lrrrrrrrrrrrr}
\toprule
bench & budget & kept & \makecell{$\tau$ \\ ours} & \makecell{$\tau$ \\ JetSpec} & $\Delta\tau$ & \makecell{95\% \\ CI} & budget & kept & \makecell{$\tau$ \\ ours} & \makecell{$\tau$ \\ JetSpec} & $\Delta\tau$ & \makecell{95\% \\ CI} \\
\midrule
\multirow{3}{*}{GSM8K} & 16 & 1317 & 9.23 & 5.98 & +3.25 & [+3.19, +3.32] & 32 & 1315 & 9.24 & 7.05 & +2.19 & [+2.13, +2.26] \\
 & 64 & 1318 & 9.23 & 7.58 & +1.65 & [+1.59, +1.72] & 128 & 1317 & 9.23 & 7.92 & +1.31 & [+1.24, +1.38] \\
 & 256 & 1316 & 9.24 & 8.36 & +0.88 & [+0.81, +0.95] &  &  &  &  &  &  \\
\midrule
\multirow{3}{*}{MATH-500} & 16 & 470 & 10.56 & 7.69 & +2.86 & [+2.74, +2.99] & 32 & 475 & 10.54 & 8.80 & +1.74 & [+1.62, +1.86] \\
 & 64 & 466 & 10.56 & 9.38 & +1.18 & [+1.06, +1.31] & 128 & 465 & 10.56 & 9.63 & +0.92 & [+0.80, +1.04] \\
 & 256 & 467 & 10.54 & 10.18 & +0.36 & [+0.23, +0.49] &  &  &  &  &  &  \\
\midrule
\multirow{3}{*}{AIME21-26} & 16 & 74 & 9.45 & 7.54 & +1.91 & [+1.31, +2.51] & 32 & 77 & 9.44 & 8.72 & +0.72 & [+0.11, +1.33] \\
 & 64 & 73 & 9.51 & 9.17 & +0.34 & [-0.29, +0.98] & 128 & 74 & 9.37 & 9.26 & +0.12 & [-0.53, +0.77] \\
 & 256 & 76 & 9.48 & 10.10 & -0.62 & [-1.28, +0.05] &  &  &  &  &  &  \\
\midrule
\multirow{3}{*}{HumanEval} & 16 & 164 & 7.93 & 5.26 & +2.67 & [+2.53, +2.81] & 32 & 163 & 7.92 & 6.28 & +1.65 & [+1.51, +1.78] \\
 & 64 & 163 & 7.92 & 6.88 & +1.05 & [+0.90, +1.20] & 128 & 164 & 7.93 & 7.17 & +0.75 & [+0.62, +0.89] \\
 & 256 & 163 & 7.93 & 7.63 & +0.30 & [+0.16, +0.45] &  &  &  &  &  &  \\
\midrule
\multirow{3}{*}{MBPP} & 16 & 257 & 7.72 & 4.97 & +2.75 & [+2.59, +2.91] & 32 & 256 & 7.72 & 5.98 & +1.74 & [+1.59, +1.91] \\
 & 64 & 256 & 7.73 & 6.54 & +1.18 & [+1.02, +1.35] & 128 & 257 & 7.72 & 6.89 & +0.83 & [+0.67, +0.99] \\
 & 256 & 257 & 7.72 & 7.34 & +0.37 & [+0.22, +0.53] &  &  &  &  &  &  \\
\midrule
\multirow{3}{*}{LCB} & 16 & 986 & 8.69 & 6.00 & +2.69 & [+2.59, +2.78] & 32 & 989 & 8.69 & 7.04 & +1.65 & [+1.55, +1.75] \\
 & 64 & 986 & 8.70 & 7.60 & +1.10 & [+1.00, +1.20] & 128 & 988 & 8.70 & 7.89 & +0.81 & [+0.71, +0.91] \\
 & 256 & 977 & 8.72 & 8.32 & +0.40 & [+0.30, +0.50] &  &  &  &  &  &  \\
\midrule
\multirow{3}{*}{MT-Bench} & 16 & 160 & 5.39 & 3.97 & +1.42 & [+0.83, +2.03] & 32 & 160 & 5.39 & 4.76 & +0.63 & [+0.01, +1.24] \\
 & 64 & 160 & 5.39 & 5.12 & +0.27 & [-0.35, +0.90] & 128 & 160 & 5.39 & 5.45 & -0.06 & [-0.68, +0.57] \\
 & 256 & 159 & 5.36 & 5.68 & -0.32 & [-0.96, +0.32] &  &  &  &  &  &  \\
\bottomrule
\end{tabular}
}
\end{table}

\begin{table}[H]
\centering
\small
\caption{Matched-prompt JetSpec comparison (Qwen3-8B, DFlare DBloom Arm-A B24). Paired change $\Delta\tau$ (ours minus JetSpec) with a 95\% interval, by tree-node budget and benchmark.}
\label{tab:jetspec-paired-arma}
\resizebox{\ifdim\width>\linewidth\linewidth\else\width\fi}{!}{\begin{tabular}{lrrrrrrrrrrrr}
\toprule
bench & budget & kept & \makecell{$\tau$ \\ ours} & \makecell{$\tau$ \\ JetSpec} & $\Delta\tau$ & \makecell{95\% \\ CI} & budget & kept & \makecell{$\tau$ \\ ours} & \makecell{$\tau$ \\ JetSpec} & $\Delta\tau$ & \makecell{95\% \\ CI} \\
\midrule
\multirow{3}{*}{GSM8K} & 16 & 1317 & 8.57 & 5.98 & +2.59 & [+2.54, +2.65] & 32 & 1315 & 8.58 & 7.05 & +1.53 & [+1.48, +1.59] \\
 & 64 & 1318 & 8.57 & 7.58 & +0.99 & [+0.94, +1.05] & 128 & 1317 & 8.57 & 7.92 & +0.65 & [+0.60, +0.71] \\
 & 256 & 1316 & 8.58 & 8.36 & +0.22 & [+0.16, +0.27] &  &  &  &  &  &  \\
\midrule
\multirow{3}{*}{MATH-500} & 16 & 470 & 10.20 & 7.69 & +2.51 & [+2.40, +2.62] & 32 & 475 & 10.19 & 8.80 & +1.39 & [+1.28, +1.50] \\
 & 64 & 466 & 10.20 & 9.38 & +0.82 & [+0.71, +0.94] & 128 & 465 & 10.20 & 9.63 & +0.57 & [+0.46, +0.68] \\
 & 256 & 467 & 10.19 & 10.18 & +0.01 & [-0.11, +0.12] &  &  &  &  &  &  \\
\midrule
\multirow{3}{*}{AIME21-26} & 16 & 74 & 9.46 & 7.54 & +1.92 & [+1.32, +2.52] & 32 & 77 & 9.44 & 8.72 & +0.72 & [+0.11, +1.33] \\
 & 64 & 73 & 9.53 & 9.17 & +0.37 & [-0.27, +1.00] & 128 & 74 & 9.37 & 9.26 & +0.11 & [-0.54, +0.77] \\
 & 256 & 76 & 9.47 & 10.10 & -0.62 & [-1.28, +0.05] &  &  &  &  &  &  \\
\midrule
\multirow{3}{*}{HumanEval} & 16 & 164 & 7.66 & 5.26 & +2.40 & [+2.28, +2.53] & 32 & 163 & 7.66 & 6.28 & +1.38 & [+1.26, +1.51] \\
 & 64 & 163 & 7.66 & 6.88 & +0.79 & [+0.65, +0.92] & 128 & 164 & 7.66 & 7.17 & +0.49 & [+0.36, +0.62] \\
 & 256 & 163 & 7.67 & 7.63 & +0.04 & [-0.10, +0.17] &  &  &  &  &  &  \\
\midrule
\multirow{3}{*}{MBPP} & 16 & 257 & 7.08 & 4.97 & +2.12 & [+1.99, +2.25] & 32 & 256 & 7.09 & 5.98 & +1.11 & [+0.98, +1.24] \\
 & 64 & 256 & 7.09 & 6.54 & +0.55 & [+0.42, +0.68] & 128 & 257 & 7.08 & 6.89 & +0.20 & [+0.08, +0.33] \\
 & 256 & 257 & 7.08 & 7.34 & -0.26 & [-0.38, -0.13] &  &  &  &  &  &  \\
\midrule
\multirow{3}{*}{LCB} & 16 & 986 & 8.58 & 6.00 & +2.58 & [+2.49, +2.68] & 32 & 989 & 8.59 & 7.04 & +1.54 & [+1.45, +1.64] \\
 & 64 & 986 & 8.59 & 7.60 & +0.99 & [+0.90, +1.09] & 128 & 988 & 8.59 & 7.89 & +0.70 & [+0.60, +0.81] \\
 & 256 & 977 & 8.61 & 8.32 & +0.29 & [+0.19, +0.39] &  &  &  &  &  &  \\
\midrule
\multirow{3}{*}{MT-Bench} & 16 & 160 & 5.27 & 3.97 & +1.30 & [+0.74, +1.87] & 32 & 160 & 5.27 & 4.76 & +0.50 & [-0.08, +1.09] \\
 & 64 & 160 & 5.27 & 5.12 & +0.14 & [-0.44, +0.74] & 128 & 160 & 5.27 & 5.45 & -0.18 & [-0.78, +0.41] \\
 & 256 & 159 & 5.24 & 5.68 & -0.44 & [-1.03, +0.17] &  &  &  &  &  &  \\
\bottomrule
\end{tabular}
}
\end{table}
 
\section{Arm-A block sizes B20, B24 and B32 on Qwen3-8B}
\label{app:qwenladder}

To see where the improvement stops, we expand the original Qwen3-8B drafters, DFlare and DFlash, directly to B20, B24 and B32,
with the same 30K corpus and recipe as the main text. B24 is rerun under the same trainer as
B20 and B32, and is not the original Arm-A checkpoint used in the main results.
Each block size is a separate expansion of the original B\QwenLadderNative{} drafter.
We use the full test sets of all \spell{\QwenLadderBenches} benchmarks, and for each benchmark the prompts that end with an
end-of-sequence token under all four drafters (Table~\ref{tab:qwen-ladder}).

At its best block size the speed-up over the original drafter improves by \QwenLadderDFlareSpeedGainHigh\% for DFlare
(at B24) and \QwenLadderDFlashSpeedGainHigh\% for DFlash (at B32) on the high-ceiling benchmarks, against
\QwenLadderDFlareSpeedGainLow\% and \QwenLadderDFlashSpeedGainLow\% on the low-ceiling ones.
Table~\ref{tab:qwen-ladder-delta} lists the paired change in $\tau$ for each comparison to the next block size, with a 95\%
interval. We call a comparison clear when its interval excludes zero. B16 to B20 is clear on all
\spell{\QwenLadderDFlareBenchesStepOne} benchmarks for both drafters, and B20 to B24 is clear on
\spell{\QwenLadderDFlareResolvedStepTwo} of \spell{\QwenLadderDFlareBenchesStepTwo} for DFlare and
\spell{\QwenLadderDFlashResolvedStepTwo} of \spell{\QwenLadderDFlashBenchesStepTwo} for DFlash. B24 to B32 is
unclear in \spell{\QwenLadderLastStepInconclusive} of \spell{\QwenLadderLastStepCells} cases, and the ceiling bin at B32
is almost empty (\QwenLadderDFlareCeilingLast\% for DFlare and \QwenLadderDFlashCeilingLast\% for DFlash).
This suggests limited additional headroom under the tested recipe. It does not rule out higher committed lengths at B64.

\begin{table}
\centering
\small
\caption{Block sizes on Qwen3-8B. The original B16 rung is set in italics. $\tau$, $\mathbb{E}[n]+1$ and the ceiling bin are means over all \spell{7} benchmarks. Speed-up is split into the \spell{3} high-ceiling benchmarks (GSM8K, MATH-500, AIME21-26) and the \spell{4} low-ceiling ones, and has no interval, as in Table~\ref{tab:speedup-change}.}
\label{tab:qwen-ladder}
\begin{tabular}{lrrrrrrr}
\toprule
\multirow{2}{*}{drafter} & \multirow{2}{*}{block} & \multirow{2}{*}{$\times$ original} & \multirow{2}{*}{$\tau$} & \multirow{2}{*}{$\mathbb{E}[n]+1$} & \multirow{2}{*}{ceiling bin} & \multicolumn{2}{c}{speed-up} \\
\cmidrule(lr){7-8}
 & & & & & & high ceiling & low ceiling \\
\midrule
DFlare & \textit{B16} & 1.00$\times$ & 7.489 & 7.178 & 15.6\% & 6.424 & 4.536 \\
DFlare & B20 & 1.25$\times$ & 8.005 & 7.626 & 8.6\% & 6.741 & 4.722 \\
DFlare & B24 & 1.50$\times$ & \textbf{8.167} & \textbf{7.746} & 3.7\% & 6.899 & 4.752 \\
DFlare & B32 & 2.00$\times$ & 8.143 & 7.703 & 0.3\% & 6.868 & 4.728 \\
\midrule
DFlash & \textit{B16} & 1.00$\times$ & 6.551 & 6.282 & 11.7\% & 5.831 & 4.212 \\
DFlash & B20 & 1.25$\times$ & 6.990 & 6.675 & 6.3\% & 6.250 & 4.350 \\
DFlash & B24 & 1.50$\times$ & \textbf{7.076} & \textbf{6.743} & 2.5\% & 6.283 & 4.364 \\
DFlash & B32 & 2.00$\times$ & 7.029 & 6.683 & 0.1\% & 6.326 & 4.370 \\
\bottomrule
\end{tabular}
\end{table}
 \begin{table}
\centering
\small
\caption{Change in committed length $\Delta\tau$ between neighboring block sizes of Table~\ref{tab:qwen-ladder}, with a 95\% paired-bootstrap interval. (ns) means that the interval includes zero.}
\label{tab:qwen-ladder-delta}
\resizebox{\ifdim\width>\linewidth\linewidth\else\width\fi}{!}{\begin{tabular}{lrrrrrr}
\toprule
 & \multicolumn{3}{c}{DFlare} & \multicolumn{3}{c}{DFlash} \\
\cmidrule(lr){2-4} \cmidrule(lr){5-7}
bench & B16 to B20 & B20 to B24 & B24 to B32 & B16 to B20 & B20 to B24 & B24 to B32 \\
\midrule
GSM8K & $+0.546$ [$+0.506$, $+0.586$] & $+0.168$ [$+0.146$, $+0.190$] & $+0.036$ [$-0.001$, $+0.072$] (ns) & $+0.505$ [$+0.474$, $+0.536$] & $+0.050$ [$+0.035$, $+0.066$] & $-0.018$ [$-0.046$, $+0.009$] (ns) \\
MATH-500 & $+0.713$ [$+0.628$, $+0.799$] & $+0.318$ [$+0.273$, $+0.366$] & $+0.116$ [$+0.029$, $+0.201$] & $+0.803$ [$+0.726$, $+0.880$] & $+0.198$ [$+0.164$, $+0.235$] & $+0.029$ [$-0.043$, $+0.100$] (ns) \\
MBPP & $+0.267$ [$+0.172$, $+0.357$] & $+0.041$ [$+0.006$, $+0.075$] & $-0.049$ [$-0.131$, $+0.031$] (ns) & $+0.085$ [$+0.014$, $+0.155$] & $+0.014$ [$-0.014$, $+0.042$] (ns) & $-0.020$ [$-0.076$, $+0.035$] (ns) \\
HumanEval & $+0.470$ [$+0.370$, $+0.574$] & $+0.101$ [$+0.061$, $+0.142$] & $-0.109$ [$-0.223$, $+0.002$] (ns) & $+0.341$ [$+0.241$, $+0.440$] & $+0.072$ [$+0.037$, $+0.107$] & $-0.075$ [$-0.179$, $+0.031$] (ns) \\
LCB & $+0.589$ [$+0.528$, $+0.651$] & $+0.164$ [$+0.120$, $+0.210$] & $-0.005$ [$-0.058$, $+0.049$] (ns) & $+0.321$ [$+0.267$, $+0.375$] & $+0.064$ [$+0.036$, $+0.093$] & $-0.019$ [$-0.061$, $+0.023$] (ns) \\
MT-Bench & $+0.192$ [$+0.094$, $+0.291$] & $+0.090$ [$+0.038$, $+0.148$] & $+0.064$ [$-0.026$, $+0.154$] (ns) & $+0.207$ [$+0.104$, $+0.315$] & $+0.010$ [$-0.029$, $+0.050$] (ns) & $+0.007$ [$-0.062$, $+0.076$] (ns) \\
AIME21-26 & $+0.836$ [$+0.550$, $+1.161$] & $+0.254$ [$+0.189$, $+0.321$] & $-0.222$ [$-0.544$, $+0.075$] (ns) & $+0.816$ [$+0.571$, $+1.084$] & $+0.189$ [$+0.137$, $+0.244$] & $-0.234$ [$-0.489$, $+0.015$] (ns) \\
\bottomrule
\end{tabular}}
\end{table}
 
\section{Testing the ceiling-bin indicator on held-out drafters}
\label{app:gateloo}

\begin{figure}[t]
\centering
\includegraphics[width=0.92\linewidth]{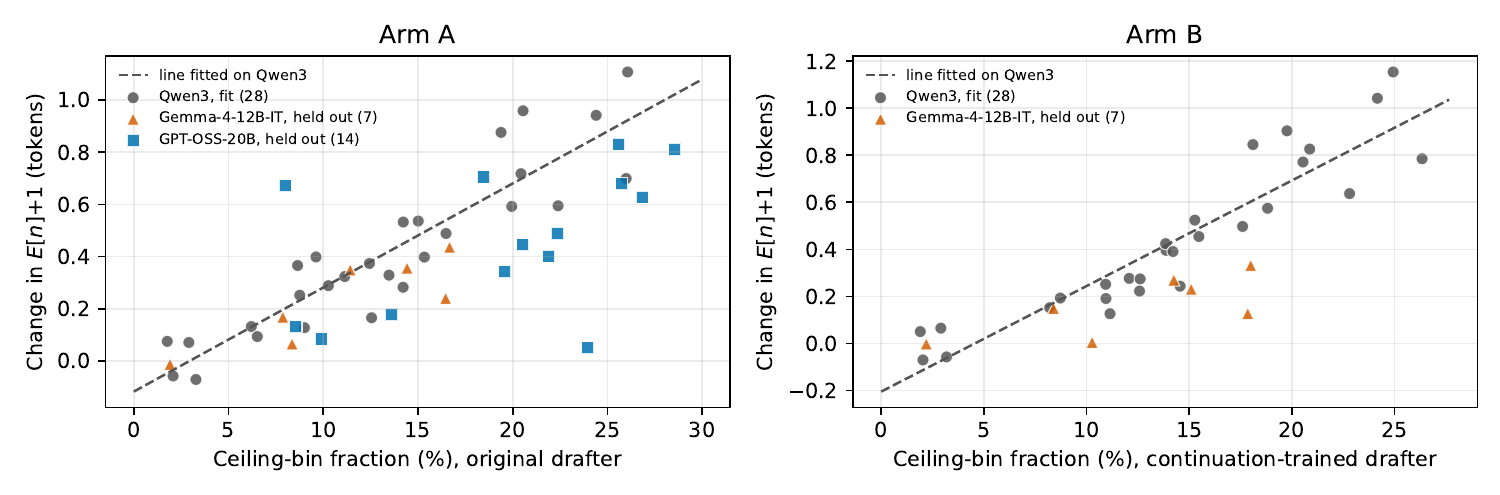}
\caption{Increase in $E[n]{+}1$ (y) versus the ceiling-bin fraction before expansion (x). Gray circles are the four Qwen3
drafters, which fit the dashed line. Orange triangles are \GemmaTarget{}, and blue squares are \GptOssTarget{}.}
\label{fig:gate}
\end{figure}

Section~\ref{sec:res-gate} fits the line on the four Qwen3 drafters, where we developed the recipe. Figure~\ref{fig:gate} and
Table~\ref{tab:gate-heldout} apply it unchanged to \GemmaTarget{} and \GptOssTarget{}. As a second check, we hold out each
Qwen3 drafter in turn, fit the line on the other three, and estimate the held-out drafter's increase. Across all
$\GateLooPredictions$ held-out estimates the mean absolute error is $\GateLooMeanAbs$~tokens, against measured increases from
$\GateLooGainMin$ to $\GateLooGainMax$~tokens. With \spell{\GateLooFolds} drafters we obtain only \spell{\GateLooFolds} held-out tests,
so both checks are small samples.

\begin{table}[H]
\centering
\small
\caption{Held-out test of the ceiling indicator, fitted on Qwen3. $\rho$ is the Spearman correlation inside the set. The error is the mean absolute difference between the Qwen3 line and the measured improvement, in tokens. GPT-OSS-20B uses the B12 drafter.}
\label{tab:gate-heldout}
\begin{tabular}{llrrrrr}
\toprule
arm & set & points & $\rho$ ($\mathbb{E}[n]{+}1$) & $\rho$ ($\tau$) & error ($\mathbb{E}[n]{+}1$) & error ($\tau$) \\
\midrule
Arm A & Qwen3 (fit) & 28 & +0.92 & +0.87 & fit & fit \\
 & Gemma-4-12B-IT & 7 & +0.86 & +0.86 & 0.10 & 0.10 \\
 & GPT-OSS-20B & 14 & +0.46 & +0.69 & 0.28 & 0.20 \\
\midrule
Arm B & Qwen3 (fit) & 28 & +0.95 & +0.97 & fit & fit \\
 & Gemma-4-12B-IT & 7 & +0.68 & +0.64 & 0.22 & 0.19 \\
\bottomrule
\end{tabular}
\end{table}
 
\section{Change in speed-up per benchmark}
\label{app:speedupchange}

Table~\ref{tab:speedup-change} shows the change in committed length and in speed-up of each arm against the original B16
drafter of the same cell, on the prompts that end with an end-of-sequence token under every cell of the row. Over its
\SpeedupChangeGapCount{} cells, the two changes differ by a median of \SpeedupChangeGapMedian{} percentage points.

\begin{table}[H]
\centering
\small
\caption{Change in committed length ($E[n]{+}1$) and in speed-up for each benchmark and arm, against the original B16 drafter of the same cell.}
\label{tab:speedup-change}
\resizebox{\ifdim\width>\linewidth\linewidth\else\width\fi}{!}{\begin{tabular}{lrrrrrrr}
\toprule
\multirow{2}{*}{drafter} & \multirow{2}{*}{bench} & \multicolumn{2}{c}{original B16} & \multicolumn{2}{c}{Arm A B24} & \multicolumn{2}{c}{Arm B B24} \\
\cmidrule(lr){3-4} \cmidrule(lr){5-6} \cmidrule(lr){7-8}
 &  & $E[n]{+}1$ & speed-up & \makecell{$E[n]{+}1$ \\ $\Delta$} & \makecell{speed-up \\ $\Delta$} & \makecell{$E[n]{+}1$ \\ $\Delta$} & \makecell{speed-up \\ $\Delta$} \\
\midrule
Qwen3-8B DFlare & GSM8K & 7.70 & 5.53x & +6.3\% & +7.9\% & +13.6\% & +15.2\% \\
Qwen3-8B DFlare & MATH-500 & 9.06 & 6.69x & +7.7\% & +7.6\% & +9.8\% & +10.1\% \\
Qwen3-8B DFlare & AIME21-26 & 8.91 & 6.91x & +12.4\% & +11.9\% & +11.7\% & +11.0\% \\
Qwen3-8B DFlare & HumanEval & 6.96 & 5.10x & +7.6\% & +7.3\% & +11.2\% & +10.9\% \\
Qwen3-8B DFlare & MBPP & 6.38 & 4.65x & +4.0\% & +3.6\% & +11.5\% & +11.8\% \\
Qwen3-8B DFlare & LCB & 7.24 & 5.40x & +7.4\% & +7.2\% & +8.6\% & +8.3\% \\
Qwen3-8B DFlare & MT-Bench & 4.03 & 2.96x & +1.8\% & -1.1\% & +1.7\% & +0.2\% \\
Qwen3-8B DFlash & GSM8K & 6.34 & 4.77x & +6.3\% & +8.3\% & +21.8\% & +24.2\% \\
Qwen3-8B DFlash & MATH-500 & 7.97 & 6.18x & +9.0\% & +9.8\% & +15.1\% & +15.7\% \\
Qwen3-8B DFlash & AIME21-26 & 7.91 & 6.41x & +12.1\% & +10.1\% & +11.5\% & +6.9\% \\
Qwen3-8B DFlash & HumanEval & 6.39 & 4.91x & +5.9\% & +6.3\% & +11.9\% & +11.6\% \\
Qwen3-8B DFlash & MBPP & 5.63 & 4.31x & +2.4\% & +2.5\% & +14.7\% & +15.0\% \\
Qwen3-8B DFlash & LCB & 6.45 & 5.06x & +5.0\% & +4.7\% & +8.5\% & +8.3\% \\
Qwen3-8B DFlash & MT-Bench & 3.30 & 2.54x & +2.3\% & +1.2\% & +4.0\% & +3.5\% \\
Qwen3-4B DFlare & GSM8K & 7.66 & 5.57x & +5.2\% & +5.3\% & +11.9\% & +12.1\% \\
Qwen3-4B DFlare & MATH-500 & 8.91 & 6.56x & +10.6\% & +10.5\% & +12.1\% & +11.7\% \\
Qwen3-4B DFlare & AIME21-26 & 8.72 & 6.97x & +6.8\% & +7.5\% & +6.5\% & +7.1\% \\
Qwen3-4B DFlare & HumanEval & 6.90 & 5.04x & +4.8\% & +4.7\% & +8.2\% & +7.9\% \\
Qwen3-4B DFlare & MBPP & 6.45 & 4.70x & +2.0\% & +2.4\% & +9.9\% & +10.4\% \\
Qwen3-4B DFlare & LCB & 7.14 & 5.34x & +4.0\% & +3.7\% & +5.2\% & +4.7\% \\
Qwen3-4B DFlare & MT-Bench & 4.17 & 3.02x & -1.7\% & -1.1\% & -2.1\% & -1.8\% \\
Qwen3-4B DFlash & GSM8K & 6.24 & 4.77x & +5.9\% & +6.1\% & +19.1\% & +19.7\% \\
Qwen3-4B DFlash & MATH-500 & 7.83 & 6.10x & +11.2\% & +10.8\% & +15.8\% & +14.9\% \\
Qwen3-4B DFlash & AIME21-26 & 7.76 & 6.58x & +7.6\% & +8.6\% & +8.3\% & +7.9\% \\
Qwen3-4B DFlash & HumanEval & 6.43 & 4.92x & +2.6\% & +2.8\% & +6.6\% & +7.1\% \\
Qwen3-4B DFlash & MBPP & 5.63 & 4.34x & +1.7\% & +1.2\% & +13.3\% & +12.8\% \\
Qwen3-4B DFlash & LCB & 6.30 & 4.95x & +4.6\% & +4.7\% & +6.9\% & +7.3\% \\
Qwen3-4B DFlash & MT-Bench & 3.40 & 2.61x & -1.7\% & -1.7\% & -0.4\% & -0.1\% \\
\bottomrule
\end{tabular}
}
\end{table}
 
\section{Temperature sampling}
\label{app:t1}

While all other results use greedy decoding, we test here whether the improvement survives temperature sampling.
We decode at temperature 1 using Qwen3-8B with the DFlash and DFlare drafters at the original B16, as well as the B24 drafters of Arm A and Arm B.
To maintain the output distribution of the target model, we apply the rejection-sampling verifier on GSM8K, MATH-500, AIME21-26 and
MBPP, with three generation seeds per cell, and average the seeds for each prompt. We pair the B24 drafter with the B16 drafter
on the same prompts, and bootstrap over prompts. We keep the prompts that end with an end-of-sequence token under every cell at
both temperatures: all of MBPP, 1{,}318 of 1{,}319 on GSM8K, 467 of 500 on MATH-500 and 89 of 179 on AIME21-26. Greedy
decoding returns runaway responses on 36\% to 38\% of AIME21-26 prompts, and sampling almost never does. The runaway prompts differ between cells, reducing the overlap of prompts that are kept across all cells at both temperatures.

\begin{table}[H]
\centering
\small
\caption{Change in committed length, B24 minus the original B16 drafter, on Qwen3-8B. Top: per prompt ($\tau$). Bottom: per cycle ($\mathbb{E}[n]+1$). The B16 column is an absolute value at T=1; every Arm A/B column is a change from B16 at that column's temperature. Only T=1 per-prompt changes carry 95\% paired intervals; (ns) means the interval includes zero.}
\label{tab:t1-sampling}
\begin{tabular}{llrrrrr}
\toprule
drafter & benchmark & B16 $\tau$ & Arm B, T=0 & Arm B, T=1 & Arm A, T=0 & Arm A, T=1 \\
\midrule
DFlare & GSM8K & 6.96 & +1.37 & +0.48 [+0.45, +0.52] & +0.71 & -0.03 [-0.07, +0.00] (ns) \\
DFlare & MATH-500 & 7.29 & +1.37 & +0.53 [+0.45, +0.61] & +1.02 & +0.23 [+0.17, +0.30] \\
DFlare & AIME21-26 & 5.68 & +0.95 & +0.26 [+0.13, +0.38] & +0.98 & +0.14 [+0.01, +0.26] \\
DFlare & MBPP & 5.81 & +0.94 & +0.31 [+0.23, +0.40] & +0.31 & -0.05 [-0.12, +0.02] (ns) \\
\midrule
DFlash & GSM8K & 5.45 & +1.69 & +1.16 [+1.12, +1.20] & +0.55 & +0.21 [+0.18, +0.23] \\
DFlash & MATH-500 & 6.34 & +1.71 & +0.97 [+0.89, +1.04] & +1.01 & +0.48 [+0.42, +0.54] \\
DFlash & AIME21-26 & 5.19 & +1.05 & +0.45 [+0.35, +0.55] & +0.88 & +0.38 [+0.28, +0.47] \\
DFlash & MBPP & 4.85 & +1.01 & +0.56 [+0.49, +0.63] & +0.10 & -0.01 [-0.06, +0.04] (ns) \\
\bottomrule
\end{tabular}
\\[6pt]
\begin{tabular}{llrrrrr}
\toprule
drafter & benchmark & B16 $\mathbb{E}[n]+1$ & Arm B, T=0 & Arm B, T=1 & Arm A, T=0 & Arm A, T=1 \\
\midrule
DFlare & GSM8K & 6.54 & +1.05 & +0.37 & +0.49 & -0.02 \\
DFlare & MATH-500 & 6.43 & +0.89 & +0.33 & +0.70 & +0.17 \\
DFlare & AIME21-26 & 5.30 & +1.04 & +0.13 & +1.11 & -0.00 \\
DFlare & MBPP & 5.38 & +0.73 & +0.16 & +0.25 & -0.08 \\
\midrule
DFlash & GSM8K & 5.19 & +1.38 & +0.97 & +0.40 & +0.19 \\
DFlash & MATH-500 & 5.81 & +1.20 & +0.60 & +0.72 & +0.35 \\
DFlash & AIME21-26 & 4.88 & +0.91 & +0.25 & +0.96 & +0.29 \\
DFlash & MBPP & 4.61 & +0.83 & +0.42 & +0.13 & +0.00 \\
\bottomrule
\end{tabular}
\end{table}
 
Arm B, the full recipe, increases committed length on all \spell{\TOnePairs} pairs of drafter and benchmark, with all
\spell{\TOneArmBClear} intervals excluding zero, and keeps \TOneShareMin\% to \TOneShareMax\% of its greedy increase. Arm A
improves on \spell{\TOneArmAClear} of the \spell{\TOnePairs} pairs, and its interval includes zero on the other
\spell{\TOneArmAHalf}. The per-cycle panel shows the same pattern: Arm B improves on \spell{\TOneCycleArmBPositive} pairs
and Arm A on \spell{\TOneCycleArmAPositive}. On every pair, accuracy stays inside the seed-to-seed spread. Arm B includes the continuation, so Arm A is the direct test of the expansion.
The ceiling-bin mass of the original drafter is lower at temperature 1 than at temperature 0 on every pair, so there is less mass for block expansion to spread to the new block positions. As a result, we expect a smaller improvement.


\begin{thebibliography}{17}
\providecommand{\natexlab}[1]{#1}
\providecommand{\url}[1]{\texttt{#1}}
\expandafter\ifx\csname urlstyle\endcsname\relax
  \providecommand{\doi}[1]{doi: #1}\else
  \providecommand{\doi}{doi: \begingroup \urlstyle{rm}\Url}\fi

\bibitem[Arriola et~al.(2025)Arriola, Gokaslan, Chiu, Yang, Qi, Han, Sahoo, and
  Kuleshov]{bd3lm2025}
Marianne Arriola, Aaron Gokaslan, Justin~T. Chiu, Zhihan Yang, Zhixuan Qi,
  Jiaqi Han, Subham~Sekhar Sahoo, and Volodymyr Kuleshov.
\newblock Block diffusion: Interpolating between autoregressive and diffusion
  language models.
\newblock In \emph{International Conference on Learning Representations
  (ICLR)}, 2025.
\newblock URL \url{https://openreview.net/forum?id=tyEyYT267x}.

\bibitem[Chen et~al.(2023)Chen, Borgeaud, Irving, Lespiau, Sifre, and
  Jumper]{chen2023}
Charlie Chen, Sebastian Borgeaud, Geoffrey Irving, Jean-Baptiste Lespiau,
  Laurent Sifre, and John Jumper.
\newblock Accelerating large language model decoding with speculative sampling.
\newblock \emph{arXiv preprint arXiv:2302.01318}, 2023.
\newblock URL \url{https://arxiv.org/abs/2302.01318}.

\bibitem[Chen et~al.(2026)Chen, Liang, and Liu]{dflash2026}
Jian Chen, Yesheng Liang, and Zhijian Liu.
\newblock {{DFlash}: Block Diffusion for Flash Speculative Decoding}.
\newblock In \emph{Proceedings of the International Conference on Machine
  Learning (ICML)}, 2026.
\newblock Released draft weights:
  \url{https://huggingface.co/z-lab/Qwen3-8B-DFlash-b16} and
  \url{https://huggingface.co/z-lab/Qwen3-4B-DFlash-b16}.

\bibitem[{Gemma Team}(2026)]{gemmateam2026gemma4technicalreport}
{Gemma Team}.
\newblock Gemma 4 technical report.
\newblock 2026.
\newblock URL \url{https://arxiv.org/abs/2607.02770}.

\bibitem[Hu et~al.(2026)Hu, Feng, Wu, Yuan, Zhao, Qian, Wang, Zhao, Jiang, Zhu,
  Rosing, and Zhang]{jetspec2026}
Lanxiang Hu, Zhaoxiang Feng, Yulun Wu, Haoran Yuan, Yujie Zhao, Yu-Yang Qian,
  Bojun Wang, Peng Zhao, Daxin Jiang, Yibo Zhu, Tajana Rosing, and Hao Zhang.
\newblock {JetSpec}: Breaking the scaling ceiling of speculative decoding with
  parallel tree drafting.
\newblock \emph{arXiv preprint arXiv:2606.18394}, 2026.
\newblock URL \url{https://arxiv.org/abs/2606.18394}.
\newblock Released draft weights:
  \url{https://huggingface.co/JetSpec/jetspec-qwen3-8b} (Qwen3-8B only).

\bibitem[Huang et~al.(2026)Huang, Zhang, Zhang, Lin, Xu, and Zhang]{domino2026}
Jianuo Huang, Yaojie Zhang, Qituan Zhang, Hao Lin, Hanlin Xu, and Linfeng
  Zhang.
\newblock Domino: Decoupling causal modeling from autoregressive drafting in
  speculative decoding.
\newblock \emph{arXiv preprint arXiv:2605.29707}, 2026.
\newblock URL \url{https://arxiv.org/abs/2605.29707}.

\bibitem[Leviathan et~al.(2023)Leviathan, Kalman, and Matias]{leviathan2023}
Yaniv Leviathan, Matan Kalman, and Yossi Matias.
\newblock Fast inference from transformers via speculative decoding.
\newblock In \emph{Proceedings of the 40th International Conference on Machine
  Learning (ICML)}, volume 202 of \emph{Proceedings of Machine Learning
  Research}, pp.\  19274--19286. PMLR, 2023.
\newblock URL \url{https://proceedings.mlr.press/v202/leviathan23a.html}.

\bibitem[Li et~al.(2024)Li, Wei, Zhang, and Zhang]{eagle2024}
Yuhui Li, Fangyun Wei, Chao Zhang, and Hongyang Zhang.
\newblock {EAGLE}: Speculative sampling requires rethinking feature
  uncertainty.
\newblock In \emph{Proceedings of the 41st International Conference on Machine
  Learning (ICML)}, volume 235 of \emph{Proceedings of Machine Learning
  Research}, pp.\  28935--28948. PMLR, 2024.
\newblock URL \url{https://proceedings.mlr.press/v235/li24bt.html}.

\bibitem[Li et~al.(2025)Li, Wei, Zhang, and Zhang]{eagle3_2025}
Yuhui Li, Fangyun Wei, Chao Zhang, and Hongyang Zhang.
\newblock {EAGLE-3}: Scaling up inference acceleration of large language models
  via training-time test.
\newblock In \emph{Advances in Neural Information Processing Systems 38
  (NeurIPS)}, 2025.
\newblock URL
  \url{http://papers.nips.cc/paper_files/paper/2025/hash/c7b5a35ea98b62512a869c19ea7b03cb-Abstract-Conference.html}.

\bibitem[Oh et~al.(2026)Oh, Cao, Kim, Jung, Ahmad, Bae, and Yun]{bastion2026}
Soowon Oh, Nam Cao, Yujin Kim, Hojung Jung, Huzama Ahmad, Sangmin Bae, and
  Se-Young Yun.
\newblock Bastion: Budget-aware speculative decoding with tree-structured block
  diffusion drafting.
\newblock \emph{arXiv preprint arXiv:2605.29727}, 2026.
\newblock URL \url{https://arxiv.org/abs/2605.29727}.

\bibitem[{OpenAI}(2025)]{gpt-oss}
{OpenAI}.
\newblock gpt-oss-120b \& gpt-oss-20b model card, 2025.
\newblock URL \url{https://arxiv.org/abs/2508.10925}.

\bibitem[{Qwen Team}(2025)]{DBLP:journals/corr/abs-2505-09388}
{Qwen Team}.
\newblock Qwen3 technical report.
\newblock \emph{arXiv preprint arXiv:2505.09388}, 2025.
\newblock URL \url{https://arxiv.org/abs/2505.09388}.

\bibitem[Ringel \& Romano(2026)Ringel and Romano]{ddtree2026}
Liran Ringel and Yaniv Romano.
\newblock Accelerating speculative decoding with block diffusion draft trees.
\newblock \emph{arXiv preprint arXiv:2604.12989}, 2026.
\newblock URL \url{https://arxiv.org/abs/2604.12989}.

\bibitem[Whalen et~al.(2026)Whalen, Ito, and Sakamoto]{whalen2026speculate}
Lexington Whalen, Yuki Ito, and Ryo Sakamoto.
\newblock Teaching diffusion to speculate left-to-right.
\newblock \emph{arXiv preprint arXiv:2606.11552}, 2026.
\newblock URL \url{https://arxiv.org/abs/2606.11552}.

\bibitem[Wu et~al.(2026)Wu, Yao, Qi, Zheng, Wang, Ma, Liao, Lakkaraju, Li, and
  Du]{dpace2026}
Tianyu Wu, Yu~Yao, Zhenting Qi, Han Zheng, Zhuohan Wang, Haoran Ma, Lawrence
  Liao, Himabindu Lakkaraju, Ju~Li, and Yilun Du.
\newblock D-pace: Dynamic position-aware cross-entropy for parallel speculative
  drafting.
\newblock \emph{arXiv preprint arXiv:2605.18810}, 2026.
\newblock URL \url{https://arxiv.org/abs/2605.18810}.

\bibitem[Zhang et~al.(2026{\natexlab{a}})Zhang, Yu, Liu, Yu, Li, Zhu, Duo,
  Xiong, Song, Yu, Zhu, and Li]{dflare2026}
Jiebin Zhang, Zhenghan Yu, Song Liu, Eugene~J. Yu, Zheng Li, Dawei Zhu,
  Jiangshan Duo, Weimin Xiong, Yifan Song, Guanghua Yu, Jianchen Zhu, and
  Sujian Li.
\newblock {DFlare}: Scaling up draft capacity for block diffusion speculative
  decoding.
\newblock \emph{arXiv preprint arXiv:2606.02091}, 2026{\natexlab{a}}.
\newblock URL \url{https://arxiv.org/abs/2606.02091}.
\newblock Released draft weights:
  \url{https://huggingface.co/AngelSlim/Qwen3-8b-dflare} and
  \url{https://huggingface.co/AngelSlim/Qwen3-4b-dflare}.

\bibitem[Zhang et~al.(2026{\natexlab{b}})Zhang, Qiu, He, and Dai]{caddtree2026}
Shuai Zhang, Huachuan Qiu, Hongliang He, and Yong Dai.
\newblock Cost-aware diffusion draft trees for speculative decoding.
\newblock \emph{arXiv preprint arXiv:2606.01813}, 2026{\natexlab{b}}.
\newblock URL \url{https://arxiv.org/abs/2606.01813}.

\end{thebibliography}
\end{document}